\documentclass[journal]{IEEEtran}
\usepackage{amsmath,amsfonts, amsthm, amssymb}
\usepackage{algorithmic}
\usepackage{array}
\usepackage[caption=false,font=normalsize,labelfont=sf,textfont=sf]{subfig}
\usepackage{textcomp}
\usepackage{stfloats}
\usepackage{url}
\usepackage{verbatim}
\usepackage{graphicx}
\usepackage{adjustbox}
\usepackage{booktabs}
\usepackage{multirow}
\def\BibTeX{{\rm B\kern-.05em{\sc i\kern-.025em b}\kern-.08em
    T\kern-.1667em\lower.7ex\hbox{E}\kern-.125emX}}
\usepackage{balance}
\usepackage[hidelinks]{hyperref}
\theoremstyle{definition}

\newtheorem{definition}{Definition}
\newtheorem{assumption}{Assumption}
\newtheorem{lemma}{Lemma}
\theoremstyle{remark}
\newtheorem*{interpretation}{Interpretation}

\begin{document}
\title{Understanding Imbalanced Forgetting in Rehearsal-Based Class-Incremental Learning}
\author{Alberto Tamajo, Srinandan Dasmahapatra, Rahman Attar \IEEEmembership{Senior Member, IEEE}
\thanks{The authors are with the School of Electronics and Computer Science, University of Southampton, SO17 1BJ Southampton, United Kingdom (e-mail: \href{mailto:at2n19@soton.ac.uk}{at2n19@soton.ac.uk}; \href{mailto:sd@ecs.soton.ac.uk}{sd@ecs.soton.ac.uk}; \href{mailto:r.attar@southampton.ac.uk}{r.attar@southampton.ac.uk}).}}


\maketitle

\begin{abstract}
Neural networks suffer from \emph{catastrophic forgetting} in class-incremental learning (CIL) settings. Rehearsal---replaying a subset of past samples---is a well-established mitigation strategy. However, recent results suggest that, despite balanced rehearsal allocation, some classes are forgotten substantially more than others. Despite its relevance, this \emph{imbalanced forgetting} phenomenon remains underexplored. This work shows that imbalanced forgetting arises systematically and severely in rehearsal-based CIL and investigates it extensively. Specifically, we construct, from a principled analysis, three last-layer coefficients that capture different gradient-level sources of interference affecting each past class \emph{during} an incremental step. We then demonstrate that, together, they reliably predict how past classes will rank in terms of forgetting \emph{at the end} of that step. While predictive performance alone does not establish causality, these results support the interpretation of the coefficients as a plausible mechanistic account linking last-layer gradient-level interactions during training to class-level forgetting outcomes. Notably, one coefficient---capturing self-induced interference---emerges as the strongest predictor, with controlled experiments providing evidence consistent with this coefficient being influenced by the new-class interference coefficient. Overall, our findings provide valuable insights and suggest promising directions for mitigating imbalanced forgetting by reducing class-wise disparities in the identified sources of interference.
\end{abstract}

\begin{IEEEkeywords}
Imbalanced forgetting, Catastrophic forgetting, Rehearsal, Experience Replay, Class-Incremental Learning, Continual Learning, Incremental Learning.
\end{IEEEkeywords}

\section{Introduction}\label{sec:introduction}
\IEEEPARstart{I}{n} a class-incremental learning (CIL) scenario \cite{RN113}, a deep neural network is trained over a sequence of incremental steps, each introducing a new set of classes to be learned. However, deep neural networks are prone to \emph{catastrophic forgetting} \cite{RN77, RN150, RN19}, i.e., they forget knowledge about past classes when exposed to such settings. \emph{Rehearsal} \cite{RN150, RN158} has emerged as one of the most established methods to mitigate this issue. Specifically, at each incremental step, it reduces forgetting by jointly training on new-class data and a small subset of past samples, known as the rehearsal set. Notably, this approach is also supported by findings in neuroscience \cite{RN66, RN151, RN67}. However, despite its widespread use and apparent effectiveness, recent experiments \cite{RN152} suggest that even when an equal proportion of rehearsal samples is allocated to each previously learned class, some classes experience substantially greater forgetting than others. If this phenomenon, termed \emph{imbalanced forgetting}, proves to be systematic, it would pose a significant challenge in real-world settings, especially in critical applications where all classes are of equal importance and none can be disproportionately forgotten. Despite its potential practical importance, imbalanced forgetting has received surprisingly limited attention in the literature thus far. Consequently, this work aims to address this gap by systematically investigating imbalanced forgetting in CIL scenarios with rehearsal.

While prior work \cite{RN152} observes imbalanced forgetting under rehearsal, it does not evaluate the phenomenon extensively, leaving open questions about its severity and systematic nature. To address this gap, we conduct comprehensive benchmarks spanning a wide range of CIL scenarios. We observe that, in every incremental step across these scenarios, the most-forgotten and least-forgotten halves of classes display noticeably different degrees of forgetting, and this disparity becomes even more pronounced when comparing the single most-forgotten class with the single least-forgotten one.\footnote{Notably, in 12\% of the incremental steps, the most affected class is heavily forgotten, whereas the least affected class is not only fully preserved but even attains slightly higher performance than it achieved when first introduced, despite having been trained on its full training dataset at that time.} Together, these results suggest that imbalanced forgetting is not an isolated artifact of specific experimental choices, but rather a systematic and robust phenomenon arising in rehearsal-based CIL, even under equal per-class allocation of rehearsal samples.

These findings motivate us to investigate \emph{why} certain past classes are forgotten more than others. To this end, we ground our investigation in the following principle, which we term the \emph{Preservation Reference Principle}: \textbf{for each past class, optimizing the loss computed on its original training dataset represents the natural empirical reference for preserving performance on that class}. This principle is warranted because, under the standard assumption that the original training and test data of a past class are drawn from the same distribution, this loss serves as a proxy for class performance. Accordingly, less effective optimization of this loss throughout an incremental step is  expected to result in lower test accuracy for that past class at the end of that step, and hence in more severe forgetting.\footnote{This perspective is consistent with episodic-memory approaches in continual learning \cite{RN119,RN121}, where retaining past knowledge is formulated in terms of constraining updates so as not to increase the loss on previous tasks.} However, rehearsal does not directly optimize over the original training data of a past class. Instead, it operates on a surrogate, a small set of rehearsal samples from that class, while jointly optimizing over the rehearsal samples from all the other past classes together with the training dataset of new classes. Consequently, class-wise differences in how well these original-data losses are optimized under rehearsal may provide an explanation for the observed imbalanced forgetting phenomenon.

Since gradient updates are the underlying mechanism driving training throughout an incremental step, we derive an upper bound, the \emph{Class-Wise R-SGD lemma}, to identify the gradient-level factors that lead the losses of different past classes, computed on their
original training datasets, to be optimized differently under rehearsal at each optimization step. From the terms appearing in the bound, we then construct three coefficients, each capturing different gradient-level sources of interference affecting each past class across the entire training trajectory of an incremental step. Although the bound involves gradients with respect to all network parameters, our coefficients restrict attention to the last layer. This choice is driven by computational considerations: computing full-network gradients throughout an incremental step is prohibitively expensive, whereas last-layer gradients are tractable. The constructed coefficients can therefore be interpreted as tractable approximations of the theoretical factors derived from the full-network analysis,\footnote{Prior work \cite{RN155, RN162, RN173, RN186, RN185} suggests that last layers disproportionately contribute to forgetting, thereby making our approximation empirically supported.} and offer a plausible mechanistic account linking last-layer gradient-level interactions during training to class-level forgetting outcomes. In more detail, we collectively refer to the resulting class-wise coefficients as the \emph{Last-Layer Imbalanced Forgetting Coefficients} and refer to them individually as follows: (i) Self-Induced Bias Interference Coefficient (SIC), (ii) Cross-Class Bias Interference Coefficient (CIC), and (iii) New-Dataset Interference Coefficient (NIC).
Notably, SIC quantifies the interference on each past class due to the bias between that class's rehearsal and original last-layer gradients. CIC quantifies the interference on each past class due to the bias between the rehearsal and original last-layer gradients of all the \emph{other} past classes. Finally, NIC quantifies the interference on each past class due to the gradients of new classes, where these gradients are restricted to the last-layer parameters associated with that past class, i.e., those parameters in the last-layer that compute the logits for that past class.

If the proposed coefficients capture relevant last-layer mechanisms contributing to imbalanced forgetting, then their values during training should carry information about the eventual class-wise pattern of forgetting. Consistent with this expectation, our comprehensive benchmarks show that, taken together, the three coefficients, which are measured \emph{during} an incremental step, reliably predict how past classes will rank in terms of forgetting \emph{at the end} of that step. While predictive performance alone does not establish causality, these results support the interpretation of the coefficients as a plausible last-layer mechanistic account of imbalanced forgetting. Specifically, we first show that SIC is the strongest predictor among the three coefficients, suggesting that it captures the most prominent last-layer gradient-level source of interference identified by our analysis. In more detail, we observe that past classes with higher SIC values during an incremental step exhibit a strong tendency to experience greater forgetting at the end of that step. Second, we show that jointly using all three coefficients provides a better prediction of how past classes will rank in terms of forgetting than SIC alone. This indicates that, despite their lower individual predictive power, CIC and NIC capture relevant complementary sources of interference beyond that reflected by SIC. Finally, in investigating why certain past classes exhibit higher SIC values than others, we find that SIC is closely and consistently associated with NIC. That is, past classes subject to greater interference from the last-layer gradients of new classes during an incremental step---as quantified by NIC---also tend to experience greater interference from the bias between their own rehearsal and original last-layer gradients. Controlled experiments provide evidence consistent with a directional influence of NIC on SIC, suggesting that the gradient-level source of interference captured by NIC may influence that of SIC.

Overall, our findings have several implications for the research community. First, they suggest that future work on rehearsal-based approaches should explicitly address imbalanced forgetting.  Second, they point to promising mitigation directions based on reducing class-wise disparities in the identified last-layer gradient-level sources of interference. To summarize, our contributions are as follows:
\begin{itemize}
    \item We demonstrate that imbalanced forgetting manifests consistently and severely across comprehensive rehearsal-based CIL benchmarks, providing evidence that this phenomenon arises systematically in this setting despite equal per-class rehearsal allocation.
    \item We construct, from a principled analysis, three training-time per-class coefficients that reliably predict the ranking of past classes in terms of forgetting, enabling prediction of the most-forgotten classes at the end of a step and offering a plausible last-layer mechanistic account of imbalanced forgetting.
    \item We show that SIC, which characterizes self-induced interference arising from the bias between a class's rehearsal and original last-layer gradients, is the strongest predictor among the three coefficients, suggesting that it captures the most prominent last-layer gradient-level source of interference identified by our analysis.
    \item Finally, we find that SIC is closely and consistently associated with NIC---the coefficient capturing per-class interference from the last-layer gradients of new classes---with controlled experiments providing evidence consistent with a directional influence of NIC on SIC.
\end{itemize}

The remainder of this paper is organized as follows. In Section \ref{sec:Related_Work}, we review relevant prior work, while Section \ref{sec:Foundations} introduces the foundations of CIL scenarios and rehearsal. In Section \ref{sec:class_wise_R-SGD Lemma}, we introduce the \emph{Last-Layer Imbalanced Forgetting Coefficients}. In Section \ref{sec:empirical_investigation}, we analyze the severity and systematic nature of imbalanced forgetting in rehearsal-based CIL, and evaluate the predictive strength of our coefficients. Finally, Section \ref{sec:Conclusion} concludes the paper, offering actionable insights for the CIL community and informing future research directions.

\section{Related Work}\label{sec:Related_Work}

\subsection{Imbalanced Forgetting in Rehearsal-Based Class-Incremental Learning}\label{sec:imbalanced_forgetting_related_work}
\noindent A recent study \cite{RN152} suggests that classes experience significantly different degrees of forgetting in rehearsal-based CIL even under equal per-class rehearsal allocation, referring to this phenomenon as \emph{imbalanced forgetting}. Although this observation represents an important first step, several aspects of this phenomenon remain unclear. First, the study does not introduce a quantitative metric to measure imbalanced forgetting, making it difficult to assess its magnitude. Second, the empirical analysis relies on a single class ordering from the CIFAR-100 dataset \cite{RN100} and a single rehearsal set size. Since both class ordering and rehearsal set size strongly influence optimization dynamics in CIL \cite{RN156, RN205}, this limited experimental configuration is insufficient to establish whether imbalanced forgetting is severe and systematic. Demonstrating that this phenomenon exhibits these properties, despite equal per-class rehearsal sampling, would be highly significant, as it would underscore a critical challenge in rehearsal-based CIL.

The authors of \cite{RN152} further hypothesize that imbalanced forgetting arises from the initial similarity between new and previously learned classes, with past classes that are initially more similar to the new ones experiencing greater forgetting. To quantify this similarity, they compute, for each past class, the average logit assigned to that class over all samples from the new classes prior to starting training on the corresponding incremental step. While this metric exhibits moderate correlation with class-wise forgetting across the second incremental steps in our benchmarks, we observe that its effectiveness degrades substantially at subsequent steps. In particular, it achieves an average Spearman’s correlation of only 0.25 across the third incremental steps in our TinyImageNet-based \cite{RN201} benchmark, indicating that initial similarity offers limited explanatory power for imbalanced forgetting.\footnote{Further details on the predictive performance of the metric in \cite{RN152} and its comparison with our coefficients are provided in Appendix \ref{sec:comparison_logits_with_our_coefficients}. The results show that our coefficients substantially outperform this metric in predicting the forgetting-based ranking of past classes at the end of a step.} Moreover, as this metric is computed prior to training, it does not capture the gradient-driven dynamics that govern how model parameters evolve throughout an incremental step. Consequently, it provides limited insight into the mechanisms through which imbalanced forgetting emerges. In addition, this metric fails to account for the role of rehearsal samples themselves in shaping imbalanced forgetting. In contrast, the coefficients introduced in our work are (i) gradient-based, (ii) defined over the entire training trajectory of an incremental step, and (iii) incorporate the influence of both rehearsal samples and new-class data.

Finally, \cite{RN152} proposes a regularization strategy that encourages feature representations of new classes to remain separated from those of similar past classes. While this approach leads to marginal gains over common rehearsal-based approaches in terms of \emph{average forgetting across past classes}, it is unclear whether these gains stem from mitigating imbalanced forgetting itself.

\subsection{Understanding Catastrophic Forgetting in Rehearsal-Based Class-Incremental Learning}
\noindent Several works provide insights into the factors contributing to catastrophic forgetting \cite{RN77, RN150, RN19} in rehearsal-based CIL. In more detail, the observations in \cite{RN178} suggest that, although rehearsal prevents optimization from leaving the initially identified low-loss region, the rehearsal set is susceptible to overfitting, leading to forgetting because rehearsal samples only coarsely approximate past data distributions. The results in \cite{RN180} indicate that the all-layer margins \cite{RN181} of past classes shrink significantly during the incremental learning process, suggesting that the rehearsal set is insufficient to maintain stable margins of past classes. Moreover, the empirical analysis in \cite{RN176} shows that the initial feature representations of newly added classes significantly overlap with those of previous classes, resulting in highly disruptive parameter updates that cannot be effectively mitigated by the limited number of rehearsal samples. Consistent with all of this, the work in \cite{RN177} shows that both feature representations and gradients computed from rehearsal sets tend to collapse into a few dominant directions, thereby capturing substantially less information than the original data distributions.
Collectively, the previous findings suggest that, although rehearsal is effective in practice, its reliance on a limited number of rehearsal samples fundamentally restricts its ability to preserve past knowledge. Complementing the negative effects of retaining only a few past-class samples, the nature of new classes also contributes to forgetting. Specifically, recent studies \cite{RN155, RN153, RN154, RN156, RN157} suggest that the degree of similarity between old and new data distributions influences the overall extent of forgetting, with intermediate levels of similarity resulting in the most severe forgetting. Additionally, not all layers in a deep neural network contribute equally to forgetting. In more detail, the findings in \cite{RN155} suggest that deeper layers (i.e., those closer to the output) disproportionately contribute to forgetting, with the final layer being particularly biased toward newly learned classes \cite{RN162, RN173, RN186}.

While the aforementioned studies provide valuable insights, they offer only a coarse-grained characterization of catastrophic forgetting in rehearsal-based CIL, overlooking the issue of imbalanced forgetting. In this work, we demonstrate that this imbalance arises systematically and severely, even with equal per-class rehearsal allocation, and accordingly conduct a class-level analysis to uncover the factors driving unequal forgetting across classes.

\section{Foundations}\label{sec:Foundations}
\noindent This section formalizes the CIL setting and introduces the mathematical formulation of rehearsal along with its optimization procedure, providing the ingredients needed to derive the \emph{Class-Wise R-SGD lemma} in the next section.

\subsection{Class-Incremental Learning}\label{sec:CIL_scenarios}
\noindent A \emph{CIL scenario} \cite{RN113} models a setting in which a deep neural network $f_\theta$ is trained over a sequence of incremental steps $S$:
\begin{equation}
    S = \left[ s_m\right]_{m \in I}\,,
\end{equation}
where $I=\{1,2,\dots,M\}$ indexes the ordered sequence of incremental steps. Each incremental step $s_m$, with $m \in I$, introduces a new training dataset $D^m$:
\begin{equation}
    D^m=\left\{\left(x^m_n,y^m_n\right)\right\}_{n=1}^{n_m}\,,
\end{equation}
which consists of $n_m$ training examples $x^m_n \in \mathbb{R}^q$ with associated targets $y^m_n \in \mathcal{Y}^m$. Note that $q$ is the dimensionality of the training examples, which remains constant throughout the sequence of incremental steps $S$, and $\mathcal{Y}^m$ is the label space for the classes introduced at step $s_m$. Importantly, there are no overlapping classes between incremental steps, that is:
\begin{equation}
    \mathcal{Y}^{m_1} \cap \mathcal{Y}^{m_2} = \emptyset, \quad \forall m_1,m_2 \in I \;\;\;\text{s.t.}\;\;\; m_1 \neq m_2\,.
\end{equation}
The ultimate goal in a CIL scenario is to continually train the underlying network $f_\theta$ so that it can recognize newly introduced classes without forgetting previously learned ones. Importantly, this must be accomplished under limited computational and memory resources, implying that training data from past incremental steps cannot be stored or revisited in full. More formally, let the empirical average loss over a dataset $D$, and evaluated at parameters $\theta$, be defined as:
\begin{equation}
    \mathcal{L}(\theta;D) := \frac{1}{n}\sum_{i=1}^nl(f_{\theta}(x_i),y_i)\,,
\end{equation}
where $l(\cdot, \cdot)$ is a sample-wise loss function, commonly the cross-entropy loss, and $n$ is the size of the dataset $D$. Then, at each incremental step $s_m$, with $m \in I$, the ideal objective is to find parameters $\theta^* \in \Theta^{m}$ that minimize the empirical average loss over all training data observed up to that step:
\begin{equation}\label{eq:CIL_objective}
    \theta^* = \underset{\theta \in \Theta^{m}}{\arg \min}\;\mathcal{L}(\theta; D^1 \cup \cdots \cup D^{m-1} \cup D^m)\,,
\end{equation}
while not having access to the actual past datasets $\{D^1, \dots, D^{m-1}\}$. Note that $\Theta^m$ denotes the parameter space of the network $f_\theta$ at step $s_m$. In particular, $\Theta^m \subseteq \mathbb{R}^{d(m)}$, where $d(m)$ denotes the number of parameters of $f_\theta$ at step $s_m$.

\subsection{Measuring Catastrophic Forgetting in Class-Incremental Learning}\label{sec:catastrophic_forgetting_CIL}
\noindent From a practical standpoint, the degree of forgetting of a past class $c$ at the end of step $s_m$, denoted by $\text{FG}_c^m$, is commonly quantified as the \emph{normalized difference in test accuracy} of $c$ between the incremental step in which it was first introduced and step $s_m$:
\begin{equation}\label{eq:forgetting}
    \text{FG}_c^m := \frac{A_c^{\text{init}} - A_c^m}{A_c^{\text{init}}}\,,
\end{equation}
where $A_c^{\text{init}}$ and $A_c^m$ denote the test accuracies of class $c$---on a $[0,1]$ scale---at the end of the step in which $c$ was first introduced and at the end of step $s_m$, respectively. For any fixed $A^{\text{init}}_c > 0$, note that $\text{FG}^m_c \in \left[1-\frac{1}{A^{\text{init}}_c}, 1\right]$, where positive values indicate forgetting, while negative values correspond to improved performance relative to the initial test accuracy. The upper bound of $1$ corresponds to complete forgetting.

\subsection{Rehearsal in Class-Incremental Learning}\label{sec:rehearsal_description}
\noindent Rehearsal \cite{RN150, RN158} is among the most established methods for pursuing the ideal objective in CIL scenarios. Specifically, it addresses constraints on computational and memory resources by storing and replaying only a subset of past training samples while learning new classes, thereby approximating past training datasets in (\ref{eq:CIL_objective}) using only a few exemplars.

Formally, at each incremental step $s_{m}$, with $m \in I$, rehearsal seeks to optimize the ideal objective in (\ref{eq:CIL_objective}) by optimizing the following surrogate objective instead:
\begin{equation}\label{eq:rehearsal_objective}
    \underset{\theta \in \Theta^m}{\min}\;\;\alpha^m\mathcal{L}(\theta; \mathcal{R}^{m-1}) + (1-\alpha^m) \mathcal{L}(\theta; D^{m})\,, 
\end{equation}
where $\mathcal{R}^{m-1}$ (the rehearsal set) is a subset of all past training data encountered prior to step $s_m$, and $\alpha^m \in (0,1)$ is a weighting coefficient that modulates the relative contribution of the two loss terms in the objective function. Typically, $\alpha^m$ is set to $1/2$ to give equal weight to both at each incremental step, although other schedules and weightings are also used.

Note that the rehearsal set is constructed iteratively by an underlying rehearsal policy $p$. Specifically, at the end of each step $s_{m}$, the policy $p$---subject to a specified memory budget---determines which samples to retain from both the previously constructed rehearsal set $\mathcal{R}^{m-1}$ and the new training dataset $D^{m}$, producing the updated rehearsal set $\mathcal{R}^{m}$:
\begin{equation}\label{eq:rehearsal_policy}
    \mathcal{R}^{m} := p(\mathcal{R}^{m-1}, D^{m})\,.
\end{equation}
It is worth emphasizing that, by construction, it follows that $\mathcal{R}^{m} \subseteq (\mathcal{R}^{m-1} \cup D^{m})$. Numerous rehearsal policies have been proposed in the literature, among which the most prominent are discussed in Section \ref{sec:rehearsal_policies}.

Finally, observe that the left term of the rehearsal training objective in (\ref{eq:rehearsal_objective}) can be decomposed as a sum of class-specific average loss functions over all previously encountered classes, with each term weighted by the empirical proportion of that past class in the rehearsal set $\mathcal{R}^{m-1}$. Consequently, the rehearsal training objective at step $s_m$ can be equivalently expressed as follows:
\begin{equation}\label{eq:rehearsal_equivalent_objective}
    \resizebox{\linewidth}{!}{$ \displaystyle\underset{\theta \in \Theta^m}{\min}\;\; \alpha^m \sum_{c \in \mathcal{Y}^{1:m-1}}\frac{|\mathcal{R}_c^{m-1}|}{|\mathcal{R}^{m-1}|}\mathcal{L}(\theta;\mathcal{R}^{m-1}_c)
    +  (1-\alpha^m) \mathcal{L}(\theta;D^{m})\,.$}
\end{equation}
Here, $\mathcal{Y}^{1:m-1}$ denotes the label space for all classes encountered before step $s_{m}$:
\begin{equation}
    \mathcal{Y}^{1:m-1}:= \bigcup_{r=1}^{m-1} \mathcal{Y}^r\,,
\end{equation}
and $\mathcal{R}^{m-1}_c$ is the subset of rehearsal samples belonging to past class $c$ at step $s_m$:
\begin{equation}
    \mathcal{R}^{m-1}_c := \{(x,y) \in \mathcal{R}^{m-1}\mid y=c\}, \quad \forall c \in \mathcal{Y}^{1:m-1}\,.
\end{equation}
By construction, it follows that, for each past class $c$, $\mathcal{R}^{m-1}_c$ is a subset of the original training dataset of $c$:
\begin{equation}\label{eq:class_specific_rehearsal_subset_class_specific_dataset}
    \mathcal{R}^{m-1}_c \subseteq D_c, \quad \forall c \in \mathcal{Y}^{1:m-1}\,,
\end{equation}
with $D_c$ denoting the original training dataset of class $c$.

Although (\ref{eq:rehearsal_objective}) and (\ref{eq:rehearsal_equivalent_objective}) are mathematically equivalent, the latter formulation makes explicit how each past class contributes to the overall loss. 

Lastly, we emphasize that the approach described so far corresponds to the standard rehearsal formulation in CIL scenarios. In this paper, we focus exclusively on this formulation, and do not explicitly consider alternative rehearsal-based approaches proposed in the literature. Since many of these approaches can be viewed as variants or extensions of the standard formulation (e.g., through auxiliary losses, additional regularization terms, or dynamic network expansion), focusing on the standard approach in isolation allows us to study imbalanced forgetting without introducing method-specific confounders.\footnote{For completeness, variants of rehearsal are reviewed in Appendix \ref{sec:extensions_rehearsal}, but fall outside the scope of the present work.}

\subsection{Rehearsal Policies in Class-Incremental Learning}\label{sec:rehearsal_policies}
\noindent Class-balanced uniform sampling is a widely adopted rehearsal policy in CIL. As the name suggests, at the end of each incremental step $s_m$, it constructs the rehearsal set $\mathcal{R}^{m}$ by uniformly sampling (without replacement) from $\mathcal{R}^{m-1} \cup D^m$, the union of the previously constructed rehearsal set and the new training dataset. The sampling procedure ensures that the classes present in $\mathcal{R}^{m-1} \cup D^m$ are equally represented in the new rehearsal set, i.e., each of those classes contributes the same number of samples to $\mathcal{R}^{m}$. Another popular rehearsal policy in CIL is herding \cite{RN113}. The latter also ensures class balance in $\mathcal{R}^m$ but it differs from class-balanced uniform sampling in that, for each class in $\mathcal{R}^{m-1} \cup D^m$, it selects the samples that lie closest to the respective class feature mean. Despite being simpler, class-balanced uniform sampling performs comparably to herding, as demonstrated in prior studies \cite{RN161, RN162, RN11}. Together, its simplicity, widespread adoption, and competitive performance make class-balanced uniform sampling a natural baseline for studying imbalanced forgetting;\footnote{Additionally, since each past class receives equal rehearsal allocation, any class-wise forgetting disparities cannot be attributed to rehearsal imbalance.} accordingly, we adopt it in our empirical investigation (Section \ref{sec:empirical_investigation}).

Although other rehearsal policies have been proposed in the CIL literature, they have not gained the same popularity as the two approaches mentioned above. This is largely because they add algorithmic complexity and computational cost while providing only modest gains in performance. We provide an overview of these methods in Appendix \ref{sec:adv_rehearal_policies}.

\subsection{Optimizing Rehearsal Objective in Class-Incremental Learning}\label{sec:variant_SGD}
\noindent At each incremental step $s_m$, with $m \in I$, the standard practice is to optimize the rehearsal training objective in (\ref{eq:rehearsal_equivalent_objective}) using a variant of stochastic gradient descent (SGD) tailored for rehearsal. For the purpose of this paper, we refer to this variant as R-SGD. The latter differs from SGD in that, at each iteration $t$ during step $s_m$, a total of $K^m_t$ samples are drawn from both the current rehearsal set $\mathcal{R}^{m-1}$ and the new training dataset $D^m$ according to a specific sampling procedure.

Specifically, $(1-\alpha^m)K_t^m$ samples are drawn independently and identically distributed (i.i.d.) from $D^m$, yielding the random mini-batch $\xi_t^{D^m}$.\footnote{Note that the coefficient 
$\alpha^m$ is identical to that used in the rehearsal training objective in (\ref{eq:rehearsal_equivalent_objective}).} The remaining $\alpha^m K_t^m$ samples are instead drawn from the rehearsal set $\mathcal{R}^{m-1}$ using the following two-stage hierarchical formulation. First, a multinomial distribution determines how many samples to draw from each past class, ensuring the total number of samples equals $\alpha^m K_t^m$:
\begin{equation}\label{eq:multinomial}
    k_t^m \sim \text{Multinomial}(\alpha^m K_t^m, p^m)\,,
\end{equation}
with $p^m \in \mathbb{R}^{|\mathcal{Y}^{1:m-1}|}$ representing the class sampling probabilities, one for each past class, and $k_t^m \in \mathbb{N}^{|\mathcal{Y}^{1:m-1}|}$ giving the resulting per-class mini-batch sizes at iteration $t$. In more detail, for any past class $c$, we denote its mini-batch size at iteration $t$ as $k_{t,c}^m$ and its sampling probability as $p_c^m$. The latter probability is simply equal to the proportion of rehearsal samples belonging to a past class $c$ within $\mathcal{R}^{m-1}$:
\begin{equation}
p_c^m := \frac{|\mathcal{R}^{m-1}_c|}{|\mathcal{R}^{m-1}|}, \quad \forall c \in \mathcal{Y}^{1:m-1}\,.
\end{equation}
Then, for each past class $c$,  $k_{t,c}^m$ samples are drawn i.i.d. from $\mathcal{R}^{m-1}_c$, yielding the random mini-batch $\xi_t^{\mathcal{R}^{m-1}_c}$. For convenience, we denote by $\xi_t^{\mathcal{R}^{m-1}}$ the set of all random mini-batches drawn from the rehearsal sets of past classes.

Regarding the update rule of R-SGD, it follows the spirit of the classic SGD formulation. Accordingly, at each iteration $t$ during step $s_m$, the update is expressed as:
\begin{equation}\label{eq:update_rule}
    \theta_{t+1} = \theta_t - \lambda_t^m\, \bar{g}(\theta_t; k_t^m, \xi_t^{\mathcal{R}^{m-1}}, \xi_t^{D^m}) \, ,
\end{equation}
with $\lambda_t^m > 0$ denoting the learning rate at iteration $t$ during step $s_m$, and the term it multiplies representing the overall stochastic gradient derived from the rehearsal training objective (\ref{eq:rehearsal_equivalent_objective}):\footnote{If any $k^m_{t,c} = 0$, the corresponding random mini-batch is empty, making the overall stochastic gradient undefined. To resolve this, we adopt the convention that, when $k^m_{t,c} = 0$, a single i.i.d.\ sample is nevertheless drawn from $\mathcal{R}^{m-1}_c$ while still treating $k^m_{t,c}$ as zero. This yields a well-defined overall stochastic gradient without introducing any contribution from the corresponding past class.}
\begin{equation}\label{eq:overall_stochastic_gradient}
    \resizebox{\linewidth}{!}{$ \displaystyle
    \bar{g}(\theta_t; k_t^m,\xi_t^{\mathcal{R}^{m-1}}, \xi_t^{D^m}) := \sum_{c \in \mathcal{Y}^{1:m-1}}\frac{k_{t,c}^m}{K_t^m}g(\theta_t;\xi_t^{\mathcal{R}^{m-1}_c})  +  (1-\alpha^m)g(\theta_t;\xi_t^{D^{m}})\,.
    $}
\end{equation}
In the equation above, for any random mini-batch $\xi^D_t \in \xi_t^{\mathcal{R}^{m-1}} \cup \{\xi_t^{D^m}\}$, $g(\theta_t; \xi_t^D)$ is the stochastic gradient evaluated on $\xi_t^D$ at iterate $\theta_t$:
\begin{equation}\label{eq:class_specific_stochastic_gradient}
g(\theta_t; \xi_t^D) := \nabla_\theta \mathcal{L}(\theta_t; D) + n(\theta_t, \xi_t^D)\,,
\end{equation}
where $\nabla_\theta \mathcal{L}(\theta_t; D)$ is the full-batch gradient at $\theta_t$ computed over the entire dataset $D$,\footnote{Throughout this paper, we denote by $\nabla_\theta$ the gradient with respect to the full parameter vector $\theta$ of a deep neural network $f_\theta$.} and $n(\theta_t, \xi_t^D)$ is a stochastic noise term denoting the deviation between the mini-batch gradient and the full-batch gradient, i.e.,:
\begin{equation}\label{eq:noise_term}
    n(\theta_t, \xi_t^D) := \nabla_\theta \mathcal{L}(\theta_t; \xi_t^D) - \nabla_\theta \mathcal{L}(\theta_t; D)\,.
\end{equation}

Importantly, the gradient of a past class $c$ over its rehearsal set $\mathcal{R}^{m-1}_c$ can be decomposed into the gradient of $c$ over its original training dataset $D_c$ plus an additional term:
\begin{equation}
    \nabla_\theta \mathcal{L}(\theta_t; \mathcal{R}^{m-1}_c) = \nabla_\theta \mathcal{L}(\theta_t; D_c) + b_c(\theta_t), \quad \forall c \in \mathcal{Y}^{1:m-1}\,,
\end{equation}
with $b_c(\theta_t)$, which we refer to as the \emph{bias vector}, capturing the deviation between the gradient over the rehearsal set of $c$ and the gradient over the original training dataset of $c$, both evaluated at $\theta_t$:
\begin{equation}\label{eq:deterministic_bias}
    b_c(\theta_t) := \nabla_\theta \mathcal{L}(\theta_t; \mathcal{R}^{m-1}_c) - \nabla_\theta \mathcal{L}(\theta_t; D_c), \quad \forall c \in \mathcal{Y}^{1:m-1}\,.
\end{equation} 

Building on the above, we can rewrite the overall stochastic gradient at iteration $t$ as follows:
\begin{align}\label{eq:equivalent_overall_stochastic_gradient}
    \bar{g}(\theta_t; k_t^m,\xi_t^{\mathcal{R}^{m-1}}, \xi_t^{D^m}) &= \resizebox{0.57\linewidth}{!}{$ \displaystyle \sum_{c \in \mathcal{Y}^{1:m-1}}\frac{k_{t,c}^m}{K_t^m}\left[\nabla_\theta \mathcal{L}(\theta_t; D_c) + B(\theta_t; \xi_t^{\mathcal{R}_c^{m-1}})\right]$} \nonumber \\  
    &\quad +  (1-\alpha^m)g(\theta_t;\xi_t^{D^{m}})\,,
\end{align}
where 
\begin{equation}\label{eq:random_bias}
    B(\theta_t; \xi_t^{\mathcal{R}^{m-1}_c}) := b_c(\theta_t) + n(\theta_t, \xi_t^{\mathcal{R}^{m-1}_c}), \quad \forall c \in \mathcal{Y}^{1:m-1}\,.
\end{equation}
The main advantage of the equation above over the mathematically equivalent expression in (\ref{eq:overall_stochastic_gradient}) is that it makes explicit the contribution of past classes' bias terms to the overall stochastic gradient.

\section{Construction of the Last-Layer Imbalanced Forgetting Coefficients}\label{sec:class_wise_R-SGD Lemma}
\noindent In this section, we first introduce the \emph{interference operator} and then present the \emph{Class-Wise R-SGD lemma}, which forms the basis for constructing the \emph{Last-Layer Imbalanced Forgetting Coefficients}. The notation follows that introduced in Section \ref{sec:Foundations}.

\subsection{Interference Operator}
\noindent The derivation of the \emph{Class-Wise R-SGD lemma} in the next subsection gives rise to an operator, which we term the \emph{interference operator}. At incremental step $s_m$, for arbitrary parameters $\theta \in \Theta^m$, any index set $S \subseteq \{1,..., d(m)\}$\footnote{As in Section \ref{sec:Foundations}, $d(m)$ denotes the number of parameters of a deep neural network $f_\theta$ at incremental step $s_m$.}, any vector $v \in \mathbb{R}^{|S|}$ and any past class $c \in \mathcal{Y}^{1:m-1}$, this operator is defined as the negative scalar projection of $v$ onto $\nabla_{\theta_S} \mathcal{L}(\theta; D_c)$:
\begin{equation}
    \text{Interf}_c^{(S)}(v; \theta) := -\frac{\langle v, \nabla_{\theta_S} \mathcal{L}(\theta; D_c)\rangle}{||\nabla_{\theta_S}\mathcal{L}(\theta; D_c)||_2}\,,
\end{equation}
where $\langle \cdot,\cdot\rangle$ denotes the inner product, $||\cdot||_2$ the $L_2$-norm, and $\nabla_{\theta_S} \mathcal{L}(\theta;D_c) \in \mathbb{R}^{|S|}$ the gradient of the loss---with respect to the subvector $\theta_S := (\theta_i)_{i \in S}$---evaluated at $\theta$ over the original training dataset of $c$. Simply put, $\nabla_{\theta_S}$ denotes the gradient with respect to the network parameters indexed by $S$. 

Using the standard geometric identity for the inner product, this operator can be equivalently expressed as:
\begin{equation}
    \text{Interf}_c^{(S)}(v; \theta) = -|| v ||_2\cos(v, \nabla_{\theta_S} \mathcal{L}(\theta; D_c))\,,
\end{equation}
where $\cos(\cdot,\cdot)$ denotes the cosine similarity between two vectors. Hence, this operator has 
a clear geometric interpretation: it measures the alignment between $v$ and $\nabla_{\theta_S} \mathcal{L}(\theta; D_c)$, scaled by the $L_2$-norm of $v$, followed by negation.\footnote{Note that the negative sign in the \emph{interference operator} is introduced to ensure that positive values align with the notion of interference.} Notably, in contrast to \cite{RN206,RN119}, which quantify interference only via cosine similarity, the \emph{interference operator} additionally accounts for the magnitude of $v$.

Since $\nabla_{\theta_S} \mathcal{L}(\theta; D_c)$ constitutes the natural first-order supervisory signal for a past class $c$ at parameters $\theta$ (restricted to the network parameters indexed by $S$), the operator $\text{Interf}_c^{(S)}(v; \theta)$ can be viewed as quantifying the degree of interference exerted by a vector $v$ on $c$ at $\theta$. In particular, positive values correspond to negative alignment between $v$ and that gradient, indicating destructive interference, whereas negative values reflect positive alignment and thus beneficial reinforcement.

\subsection{Formulation of the Class-Wise R-SGD Lemma}
\noindent Guided by the \emph{Preservation Reference Principle} (Section \ref{sec:introduction}), we here take an initial step toward understanding why certain past classes are forgotten more than others. Specifically, we aim to identify the gradient-level factors arising from the rehearsal set and the new-class training data that lead the losses of different past classes, computed on their original training datasets, to be optimized differently under the same single R-SGD update step (\ref{eq:update_rule}).\footnote{Since this paper investigates imbalanced forgetting in rehearsal-based CIL, we focus on the factors that directly arise from these components.}
To this end, inspired by the classical SGD lemma \cite{RN10}---which bounds the expected change of a loss
function after a single update step---we derive a class-wise analogue tailored to the rehearsal setting. We refer to the latter as the \emph{Class-Wise R-SGD lemma}. Similarly to SGD analyses, our result relies on a smoothness assumption on the class-specific loss of each past class:
\begin{assumption}[$L_c^m$-Smoothness]\label{ass:L_smootness}
    At incremental step $s_m$, for each past class $c \in \mathcal{Y}^{1:m-1}$, the loss of $c$ over its original training dataset, $\mathcal{L}(\cdot; D_c)$, is $L^m_c$-smooth over $\Theta^m$. That is, it is continuously differentiable on $\Theta^m$ and there exists a constant $L_c^m > 0$ such that for all $\theta_1, \theta_2 \in \Theta^m$:
    \begin{equation}
        \resizebox{\linewidth}{!}{$\displaystyle
        \mathcal{L}(\theta_2;D_c) \leq \mathcal{L}(\theta_1;D_c) + \langle\nabla_\theta \mathcal{L}(\theta_1;D_c),(\theta_2-\theta_1)\rangle + \frac{L_c^m}{2}||\theta_2-\theta_1||^2_2 \, .$}
    \end{equation}
\end{assumption}
Observe that the smoothness constant is allowed to vary across past classes, thereby accommodating heterogeneous curvature in the class-specific loss functions. Under the assumption above, the following bound holds, with the full derivation deferred to Appendix \ref{sec:proof_class_wise_SGD_lemma}:

\vspace{0.2em} 

\begin{lemma}[Class-Wise R-SGD Lemma]\label{lemma:per_iteration_class_wise_GD_inequality}
    Suppose that a single R-SGD update step (\ref{eq:update_rule}) is performed at iterate $\theta_t$ during incremental step $s_m$, producing the next iterate $\theta_{t+1}$. Then, for each past class $c \in \mathcal{Y}^{1:m-1}$, and under the $L_c^m$-smoothness assumption (Assumption \ref{ass:L_smootness}), the expected change in the loss of class $c$---computed on its original training dataset---is upper-bounded by the sum of three terms, each evaluated at $\theta_t$:
    \begin{align}
        \mathbb{E}_{z_t}\left[\Delta_t^c\right] &\leq - \lambda_t^m \alpha^m p_c^m \,|| \nabla_\theta \mathcal{L}(\theta_t;D_c)||_2^2 \nonumber \\
        & \quad +\lambda_t^m \, || \nabla_\theta \mathcal{L}(\theta_t;D_c)||_2 \,I_c^m(\theta_t) \nonumber \\
        & \quad + \frac{(\lambda_t^m)^2L_c^m}{2}\,\mathbb{E}_{z_t}\left[\left|\left|\bar{g}(\theta_t; k_t^m, \xi_t^{\mathcal{R}^{m-1}},\,\xi_t^{D^m} )\right|\right|^2_2\right] \,,
    \end{align}
    where $\mathbb{E}_{z_t}[\cdot]$ denotes the joint expectation over $k_t^m$, $\xi_t^{\mathcal{R}^{m-1}}$ and $\xi_t^{D^m}$, $\Delta_t^c := \mathcal{L}(\theta_{t+1}; D_c) -\mathcal{L}(\theta_t; D_c)$, and $I_c^m(\theta_t)$ is defined as follows:
    \begin{align}\label{eq:overall_interference_term}
        I^m_c(\theta_t) &:= \left(\alpha^m p_c^m\right)\,\text{Interf}_c^{(F)}(b_c(\theta_t);\,\theta_t) \nonumber \\
        & \quad + \sum_{y \in \mathcal{Y}^{1:m-1}_{-c}}\left(\alpha^m p_y^m\right)\,\text{Interf}_c^{(F)}(b_y(\theta_t);\,\theta_t) \nonumber \\
        & \quad + \left(1-\alpha^m \right)\,\text{Interf}_c^{(F)}(\nabla_\theta \mathcal{L}(\theta_t; D^{m});\,\theta_t)  \nonumber \\
        & \quad + \sum_{y \in \mathcal{Y}^{1:m-1}_{-c}}\left(\alpha^m p_y^m\right)\,\text{Interf}_c^{(F)}(\nabla_\theta \mathcal{L}(\theta_t; D_y);\,\theta_t)\,,
    \end{align}
    where $\mathcal{Y}^{1:m-1}_{-c}$ denotes the set of all past classes encountered prior to step $s_m$ excluding $c$, i.e., $\mathcal{Y}^{1:m-1}_{-c} := \mathcal{Y}^{1:m-1} \setminus \{c\}$, and $F$ indexes all parameters of the underlying neural network $f_\theta$ at step $s_m$, i.e., $F:= \{1,\dots,d(m)\}$.
\end{lemma}

According to the lemma, at iterate $\theta_t$ during step $s_m$, three terms upper-bound the expected one-step change in the loss of a past class $c$ on its original training dataset. Importantly, neither the first nor the third term capture differences in how past classes are optimized induced by the rehearsal set and the new-class training data.\footnote{We do not claim that these terms are irrelevant to class-wise optimization dynamics; rather, they do not isolate the rehearsal- and new-class-induced gradient-level factors that are the focus of our analysis.}  Specifically, the first term is proportional to the squared $L_2$-norm of the natural first-order signal for $c$ at $\theta_t$, which, conditioned on the current iterate, is independent of both the rehearsal set and the newly introduced training data. The only factor in this term related to rehearsal is $p^m_c$, i.e., the proportion of stored rehearsal samples belonging to class $c$ at step $s_m$. Since we study imbalanced forgetting under balanced rehearsal allocation, $p^m_c$ is identical across past classes, and thus it cannot explain class-wise optimization disparities. The third term corresponds to the expected squared $L_2$-norm of the overall stochastic gradient (\ref{eq:equivalent_overall_stochastic_gradient}), scaled by the quadratic curvature factor induced by the $L^m_c$-smoothness assumption. Its only class-specific component is the smoothness constant, which is intrinsic to each past class
and thus decoupled from both the rehearsal set and the new-class training data. Regarding the second term, it is proportional to $I^m_c(\theta_t)$ (\ref{eq:overall_interference_term}), which captures the overall degree to which multiple vector components interfere specifically with class $c$ at $\theta_t$. Importantly, the first and second terms of 
$I^m_c(\theta_t)$ depend on the rehearsal set, whereas the third term depends on the training data of newly introduced classes.\footnote{The last term in $I^m_c(\theta_t)$ (\ref{eq:overall_interference_term}) depends only on natural first-order supervisory signals; thus, it does not account for rehearsal- or CIL-induced differences in how past classes are optimized.} Therefore, the second term identifies the interference pathways through which the rehearsal set and the new-class training data induce disparities across past classes.
 
In summary, the \emph{Class-Wise R-SGD lemma} highlights that three gradient-level factors---two arising from the rehearsal set and one from the training data of new classes---lead distinct past classes to be optimized differently, in expectation, under the same single R-SGD update step. These factors correspond to the first, second, and third interference terms in (\ref{eq:overall_interference_term}). Although other terms in the bound may also influence class-wise loss changes, our empirical results in the next section show that the coefficients derived from these interference factors reliably predict the forgetting-based ranking of past classes.

\subsection{Last-Layer Imbalanced Forgetting Coefficients}\label{sec:imbalanced_forgetting_coefficients}
\noindent The gradient-level factors identified in the previous lemma capture differences in how past classes are optimized at a \emph{single} R-SGD update iteration, and therefore fail to characterize optimization discrepancies over the entire training process of an incremental step. To address this, we aggregate each factor along the full R-SGD training trajectory of a step, thereby yielding three per-class coefficients that capture distinct gradient-level sources of interference across the full training process.
However, while the factors in the lemma involve gradients with respect to all parameters of a deep neural network, our coefficients restrict attention to the last-layer parameters only. This modification is motivated by computational considerations because computing gradients with respect to all parameters along the training trajectory of an incremental step is computationally prohibitive, whereas restricting attention to the last-layer parameters is tractable. The constructed coefficients can
therefore be interpreted as tractable approximations of the theoretical factors derived from the \emph{Class-Wise R-SGD lemma}. Prior studies \cite{RN155, RN162, RN173, RN186, RN185} have shown that last layers contribute disproportionately to forgetting, thereby supporting our focus on these parameters.

Before formally introducing these coefficients, we provide some auxiliary definitions. Let $L$ be a set that indexes only the last-layer parameters of a deep neural network. Also, let $L_c$ be a set that indexes only the last-layer parameters of a deep neural network associated with a specific past class $c$, i.e., the parameters in the last layer that compute the logits for that class. Finally, in direct correspondence with the bias vector (\ref{eq:deterministic_bias}), we refer to $b_c^L(\theta_t)$ as the \emph{last-layer bias vector}. At incremental step $s_m$, this vector captures the deviation between the gradient over the rehearsal set of past class $c$ and the gradient over the original training dataset of $c$, both evaluated at $\theta_t$ and taken with respect to the last-layer parameters of the underlying network:
\begin{equation}
    b_c^L(\theta_t) := \nabla_{\theta_L}\mathcal{L}(\theta_t; \mathcal{R}^{m-1}_c) - \nabla_{\theta_L}\mathcal{L}(\theta_t; D_c), \quad \forall c \in \mathcal{Y}^{1:m-1}\,.
\end{equation}

Our coefficients, which we refer to as the \emph{Last-Layer Imbalanced Forgetting Coefficients}, are formally defined in the following:\\

\begin{definition}[Self-Induced Bias Interference Coefficient (SIC)] \label{def:SBC}
Let $\{\theta_t\}_{t=0}^{T-1}$ denote the entire sequence of R-SGD iterates during incremental step $s_m$. For each past class $c \in \mathcal{Y}^{1:m-1}$, $\text{SIC}_c^m$ is defined as follows:
\begin{equation}
    \text{SIC}_c^m := \left(\alpha^m p_c^m\right)\sum_{t=0}^{T-1} \text{Interf}^{(L)}_c(b_c^L(\theta_t);\, \theta_t).
\end{equation}
\end{definition}
\begin{interpretation}
    For each past class $c$, $\text{SIC}^m_c$ captures the interference on $c$ throughout step $s_m$ arising from the bias between $c$'s rehearsal and original last-layer gradients. This is then weighted by $\alpha^m$ and $p^m_c$, the proportion of past-class samples drawn at each R-SGD iteration and the proportion of stored rehearsal samples belonging to $c$, respectively, during step $s_m$.
\end{interpretation}

\vspace{0.5em} 

\begin{definition}[Cross-Class Bias Interference Coefficient (CIC)]
Let $\{\theta_t\}_{t=0}^{T-1}$ denote the entire sequence of R-SGD iterates during incremental step $s_m$. For each past class $c \in \mathcal{Y}^{1:m-1}$, $\text{CIC}_c^m$ is defined as follows:
\begin{equation}
    \text{CIC}_c^m := \sum_{y \in \mathcal{Y}^{1:m-1}_{-c}}\left(\alpha^m p_y^m\right)\,
    \sum_{t=0}^{T-1}\text{Interf}^{(L)}_c(b_y^L(\theta_t);\, \theta_t).
\end{equation}
\end{definition}
\begin{interpretation}
    For each past class $c$, $\text{CIC}^m_c$ captures the interference on $c$ throughout step $s_m$ arising from the bias between the rehearsal and original last-layer gradients of all other past classes $y \neq c$.  For each past class $y \neq c$, the corresponding interference contribution is weighted by $\alpha^m p^m_y$ and then summed over all such classes.
\end{interpretation}

\vspace{0.5em} 

\begin{definition}[New-Dataset Interference Coefficient (NIC)] \label{def:NIC}
Let $\{\theta_t\}_{t=0}^{T-1}$ denote the entire sequence of R-SGD iterates during incremental step $s_m$. For each past class $c \in \mathcal{Y}^{1:m-1}$, $\text{NIC}_c^m$ is defined as follows:
\begin{equation}
    \text{NIC}_c^m :=  \left(1-\alpha^m \right)\sum_{t=0}^{T-1} \text{Interf}^{(L_c)}_c(\nabla_{\theta_{L_c}} \mathcal{L}(\theta_t; D^{m});\, \theta_t)\,.
\end{equation}
\end{definition}
\begin{interpretation}
    For each past class $c$, $\text{NIC}^m_c$ captures the interference on $c$ throughout step $s_m$ arising from the $c$-specific last-layer gradients of new classes. This is then weighted by $(1-\alpha^m)$, the proportion of new-class samples drawn at each R-SGD iteration during step $s_m$. The $c$-specific last-layer gradients are taken with respect to the parameters in the last-layer associated with $c$, i.e, the parameters in the last-layer that compute the logits for class $c$.
\end{interpretation}

Finally, since gradient updates drive training throughout an incremental step, our constructed coefficients can be viewed as offering a plausible mechanistic account linking last-layer gradient-level interactions during training to class-level forgetting outcomes, and hence to imbalanced forgetting. Although they do not constitute definitive proof, the results in the next section are consistent with this interpretation.
Also, note that, unlike SIC and CIC, NIC is defined using gradients with respect to the last-layer parameters associated with a specific past class. This distinction arises from empirical observations indicating that restricting the gradients to class-specific last-layer parameters, rather than considering all last-layer parameters, yields stronger performance for NIC, i.e., better prediction of how past classes will rank in terms of forgetting and closer association with SIC. Further details are deferred to Appendix \ref{sec:empirical_comp_NIC_ALL_NIC}.

\section{Empirical Investigation}\label{sec:empirical_investigation}
\noindent This section investigates the generality and severity of imbalanced forgetting in CIL with rehearsal, and evaluates the predictive power of the \emph{Last-Layer Imbalanced Forgetting Coefficients}.\footnote{Code available at \href{https://github.com/albertotamajo/Understanding-Imbalanced-Forgetting-in-Rehearsal-based-Class-Incremental-Learning}{https://github.com/albertotamajo/Understanding-Imbalanced-Forgetting-in-Rehearsal-based-Class-Incremental-Learning}}

\subsection{Construction of the Benchmarks}\label{sec:benchmarks}
\noindent Optimization dynamics in rehearsal-based CIL depend on multiple factors, including the number of classes introduced per step, class ordering, rehearsal set size and composition, and R-SGD stochasticity. To systematically assess the generality and severity of imbalanced forgetting, and to evaluate the predictive strength of our coefficients, we explicitly control for these factors through comprehensive empirical evaluation. Specifically, we construct two full-factorial benchmarks, $\Omega_{\mathrm{C100}}$ and $\Omega_{\mathrm{TIN}}$, each defined as the Cartesian product of three experimental factor sets:\footnote{Typical CIL benchmarks \cite{RN183} use a single class ordering from a given dataset and a fixed rehearsal set size, and are therefore unsuitable for a systematic empirical investigation.}
\begin{align}
\resizebox{\linewidth}{!}{$\displaystyle
\Omega_d := \left\{(S, p, i) \mid S \in \mathcal{S}_d, p \in \mathcal{P}, i \in \mathcal{I}_{\text{seed}}\right\}, \;\; \forall d \in \{\mathrm{C100}, \mathrm{TIN}\}\,,$}
\end{align}
where each $(S,p,i)$ represents a CIL experiment defined by a sequence of incremental steps $S$, a rehearsal policy $p$, and a random seed $i$.

By construction, both benchmarks share two experimental factor sets: $\mathcal{P}$ and $\mathcal{I}_{\text{seed}}$.
The former includes three class-balanced uniform sampling policies that retain 8\%, 20\%, and 40\% of rehearsal samples per class.  The latter represents the machine-representable seed space, with each seed controlling the stochastic operations of R-SGD and the policies in $\mathcal{P}$.\footnote{Details on the reasons for including $\mathcal{I}_{\text{seed}}$ as a factor set are provided in Appendix \ref{sec:importance_random_seed_benchmarks}.} In contrast, the benchmarks differ in their sets of incremental-step sequences, denoted by $\mathcal{S}_{\mathrm{C100}}$ and $\mathcal{S}_{\mathrm{TIN}}$. These sets are derived from two widely used datasets in the CIL community: $\mathcal{S}_{\mathrm{C100}}$ from CIFAR-100 \cite{RN100}, and $\mathcal{S}_{\mathrm{TIN}}$ from the more challenging Tiny-ImageNet \cite{RN201}.\footnote{Descriptions of the CIFAR-100 and Tiny-ImageNet datasets are provided in Appendix \ref{sec:datasets_details}, along with the rationale for using Tiny-ImageNet instead of the more standard ILSVRC2012 ImageNet \cite{RN202}.} More precisely, these sets are defined as the union of three subsets:
\begin{align}
    \mathcal{S}_d &:= \mathcal{S}_d^{10\%} \cup \mathcal{S}_d^{20\%} \cup \mathcal{S}_d^{50\%}, \quad \forall d \in \{\mathrm{C100}, \mathrm{TIN}\}\,,
\end{align}
where $\mathcal{S}_{d}^{p\%}$ comprises \emph{all distinct} CIL sequences in which each step introduces $p\%$ of the dataset classes. The sequences have length three for $p \in \{10, 20\}$ and length two for $p = 50$, since introducing 50\% of the classes per step covers the entire class set after two steps.\footnote{We restrict the analysis to three incremental steps, as the trends observed in the next subsections are indicative of subsequent behavior. Additional steps would incur substantial computational cost with marginal benefit.} To ensure balanced contribution within each set, we downsample larger subsets so that all three contain the same number of sequences.

As is common practice in the CIL literature (e.g., \cite{RN183}), we use ResNet-32 and ResNet-18 \cite{RN184} as the main network backbones for the experiments in $\Omega_{\mathrm{C100}}$ and $\Omega_{\mathrm{TIN}}$, respectively. In addition, we set the weighting coefficient $\alpha^m$ (\ref{eq:rehearsal_objective}) to the standard value of $1/2$ in both benchmarks. The hyperparameters used for R-SGD are detailed in Appendix \ref{sec:training_details}.

Finally, for each $d \in \{\text{C100}, \text{TIN}\}$ and $k \in \{2,3\}$, we denote by $\Omega_d^k$ the set of all $k$-th incremental steps in $\Omega_d$. Note that $\Omega_d^2$ consists of nine disjoint partitions of equal size, corresponding to the combination of three class-introduction percentages (10\%, 20\%, 50\%) and three rehearsal retention rates (8\%, 20\%, 40\%). In contrast, $\Omega_d^3$ consists of six disjoint partitions of equal size, since only two class-introduction percentages (10\%, 20\%) are applicable at the third step. We further denote by $\Omega_d^{2;\,\{10\%, 20\%\}}$ the subset of $\Omega_d^{2}$ consisting of second steps in which $10\%$ or $20\%$ of new classes are introduced, thereby excluding those with $50\%$.  This subset enables a direct comparison between second and third steps without confounding due to differences in the number of classes introduced.\footnote{As shown in Appendix \ref{sec:fine_grained_analysis_imb_forgetting}, \ref{sec:fine_grained_analysis_indi_predi_strength}, and \ref{sec:finer_grained_anal_joint_pred_strength}, both imbalanced forgetting and the predictive power of our derived coefficients vary with the number of classes introduced per step, justifying the introduction of this subset.}

\subsection{Stratified Sampling of the Benchmarks}\label{sec:Stratified_Sampling_of_the_Benchmarks}
\noindent Exhaustive evaluation over all second and third incremental steps in our benchmarks is computationally infeasible. For each $d \in \{\mathrm{C100}, \mathrm{TIN}\}$ and $k \in \{2,3\}$, we therefore perform stratified sampling with equal allocation across all partitions in $\Omega^k_d$. Concretely, we draw 40 i.i.d. samples from each partition in $\Omega_d^k$, resulting in 360 second-step samples (9 partitions $\times$ 40) and 240 third-step samples (6 partitions $\times$ 40) per benchmark, for an overall total of 1{,}200 sampled steps across both $\Omega_{\mathrm{C100}}$ and $\Omega_{\mathrm{TIN}}$. Since the partitions have equal cardinality, this procedure is equivalent to i.i.d. sampling from the uniform distribution over $\Omega_d^k$. For convenience, we denote the resulting sampled benchmarks as $\hat{\Omega}_{\mathrm{C100}}$ and $\hat{\Omega}_{\mathrm{TIN}}$, adopting the same superscript convention used for $\Omega_{\mathrm{C100}}$ and $\Omega_{\mathrm{TIN}}$. 

The resulting samples are used to estimate population-level statistics of interest.
To quantify sampling uncertainty, we use 95\% confidence intervals (CIs) alongside all estimates. Details of their computation are provided in Appendix \ref{sec:confidence_intervals_computation}.

\subsection{Computation of the Last-Layer Imbalanced Forgetting Coefficients in the Benchmarks}\label{sec:comput_imbalanced_forg_coeff_benchmarks}
\noindent The \emph{Last-Layer Imbalanced Forgetting Coefficients} (Section \ref{sec:imbalanced_forgetting_coefficients}) are defined over the full R-SGD training trajectory of an incremental step, making their exact computation prohibitively expensive in our benchmarks. To overcome this, we approximate them using evenly spaced checkpoints, with the number of checkpoints set to the number of training epochs plus one. This includes the initial R-SGD step, along with those corresponding to the end of each subsequent epoch. Since gradient-based quantities typically vary smoothly across nearby iterations and exhibit strong temporal correlation (making consecutive steps often redundant), this procedure provides a suitable tractable approximation of the full training trajectory for the following ranking-based analyses.

\subsection{Analysis of Imbalanced Forgetting}\label{sec:analysis_imbalanced_forgetting}
\noindent We now introduce two complementary metrics to quantify the severity of imbalanced forgetting across our benchmarks. 
The first metric, termed the \emph{Forgetting Range} (FG-R), measures the difference in forgetting between the most forgotten class and the least forgotten one at the end of incremental step $s_m$:
\begin{equation}
    \text{FG-R}^m := \max_{c \in \mathcal{Y}^{1:m-1}}\text{FG}_{c}^m - \min_{c \in \mathcal{Y}^{1:m-1}}\text{FG}_c^m\,.
\end{equation}
The second metric, the \emph{Forgetting Half-Gap} (FG-HG), quantifies the difference between the average forgetting of the top $50\%$ most forgotten classes and that of the bottom $50\%$ at the end of step $s_m$:
\begin{equation}
    \text{FG-HG}^m := \frac{1}{|\mathcal{C}_{\text{top}}^m|}\sum_{c \in \mathcal{C}_{\text{top}}^m}\text{FG}_c^m  - \frac{1}{|\mathcal{C}_{\text{bot}}^m|}\sum_{c \in \mathcal{C}_{\text{bot}}^m}\text{FG}_c^m\,,
\end{equation}
where $\mathcal{C}_{\text{top}}^m$ and $\mathcal{C}_{\text{bot}}^m$ denote the top and bottom $50\%$ most forgotten classes at the end of step $s_m$, respectively. Intuitively, FG-R captures the extreme disparity in forgetting, while
FG-HG complements it by quantifying the typical gap between the most and least forgotten halves of classes. Both metrics equal zero if and only if forgetting is perfectly balanced across all previously learned classes.

The results in Table \ref{tab:FG-R_FG-HSG_overall} show that imbalanced forgetting is both substantial and consistent across our benchmarks. In absolute terms, the mean value of FG-R consistently exceeds 0.55, reaching a top value of 0.71 for the third incremental steps in  $\Omega_{\mathrm{TIN}}$. Since FG-R is typically bounded within $[0,1]$,\footnote{Across all incremental steps sampled from both $\Omega_{\mathrm{C100}}$ and $\Omega_{\mathrm{TIN}}$, the least-forgotten class exhibits negative forgetting in only 12\% of cases. These values are, however, close to zero, with an average of $-0.04$. Clearly, the most-forgotten class always exhibits positive forgetting.} these magnitudes imply that, on average, the most forgotten class forgets substantially more than the least forgotten one. Even the more conservative FG-HG metric reaches mean values between 0.25 and 0.28. This indicates that the average forgetting of the most forgotten half of classes exceeds that of the least forgotten half by roughly one quarter of the typical forgetting range, confirming that imbalance is not limited to extreme outliers but includes a substantial fraction of classes. Moreover, a notable structural pattern emerges when comparing the second and third incremental steps. Specifically, in both $\Omega_{\mathrm{C100}}$ and $\Omega_{\mathrm{TIN}}$, the mean value of FG-R increases from the second to the third incremental steps (with the differences being statistically significant at the 95\% level). In contrast, FG-HG remains relatively stable across steps. This pattern suggests that imbalanced forgetting progressively increases at each subsequent step, primarily due to a widening gap between the most forgotten class and the least forgotten one, rather than a growing separation between the top and bottom halves of classes. Furthermore, for both incremental steps, the average values of FG-R and FG-HG are higher in $\Omega_{\mathrm{TIN}}$ than in $\Omega_{\mathrm{C100}}$. Given that $\Omega_{\mathrm{TIN}}$ is a more complex CIL benchmark than $\Omega_{\mathrm{C100}}$, these results suggest that the severity of imbalanced forgetting scales with CIL complexity. Finally, the relatively small standard deviations of both FG-R and FG-HG (with FG-HG showing lower variability than FG-R) indicate that the discrepancies in forgetting captured by these metrics are consistent across experiments.

In summary, our empirical investigation shows that imbalanced forgetting is substantial and consistent across our benchmarks, providing evidence that this phenomenon arises systematically and severely in rehearsal-based CIL, despite equal per-class rehearsal allocation. Although consistently pronounced, the degree of imbalance varies depending on CIL complexity and depth of incremental steps. For completeness, Appendix \ref{sec:fine_grained_analysis_imb_forgetting} provides a finer-grained analysis of imbalanced forgetting.

\begin{table}[htbp]
  \centering
  \caption{Estimated mean $\pm$ standard deviation (with 95\% CIs) of FG-R and FG-HG for the second and third incremental steps in $\Omega_{\mathrm{C100}}$ and $\Omega_{\mathrm{TIN}}$.}
  \label{tab:FG-R_FG-HSG_overall}
  \resizebox{\linewidth}{!}{%
  \begin{tabular}{lcc}
    \toprule
    & FG-R & FG-HG \\
    \midrule
    \multicolumn{1}{l}{} &
\multicolumn{2}{c}{$\Omega_{\mathrm{C100}}$}\\
    \midrule
    $\Omega^2_{\mathrm{C100}}$
      & 0.59 {\tiny(0.57, 0.60)} $\pm$ 0.15 {\tiny(0.15, 0.17)}
      & 0.25 {\tiny(0.25, 0.26)} $\pm$ 0.07 {\tiny(0.06, 0.07)} \\
    \midrule
    $\Omega^{2;\,\{10\%, 20\%\}}_{\mathrm{C100}}$
      & 0.55 {\tiny(0.53, 0.57)} $\pm$ 0.16 {\tiny(0.15, 0.17)}
      & 0.25 {\tiny(0.24, 0.26)} $\pm$ 0.08 {\tiny(0.07, 0.08)} \\
    \midrule
    $\Omega^3_{\mathrm{C100}}$
      & 0.62 {\tiny(0.60, 0.64)} $\pm$ 0.15 {\tiny(0.14, 0.16)}
      & 0.26 {\tiny(0.25, 0.26)} $\pm$ 0.07 {\tiny(0.06, 0.07)} \\
    \midrule
    \multicolumn{1}{l}{} &
\multicolumn{2}{c}{$\Omega_{\mathrm{TIN}}$}\\
    \midrule
    $\Omega^2_{\mathrm{TIN}}$
      & 0.70 {\tiny(0.69, 0.71)} $\pm$ 0.13 {\tiny(0.12, 0.14)}
      & 0.28 {\tiny(0.27, 0.28)} $\pm$ 0.05 {\tiny(0.05, 0.06)} \\
    \midrule
    $\Omega^{2; \, \{10\%, 20\%\}}_{\mathrm{TIN}}$
      & 0.66 {\tiny(0.64, 0.67)} $\pm$ 0.13 {\tiny(0.12, 0.14)}
      & 0.27 {\tiny(0.27, 0.28)} $\pm$ 0.06 {\tiny(0.05, 0.06)} \\
    \midrule
    $\Omega^3_{\mathrm{TIN}}$
      & 0.71 {\tiny(0.69, 0.72)} $\pm$ 0.11 {\tiny(0.10, 0.12)}
      & 0.27 {\tiny(0.27, 0.28)} $\pm$ 0.05 {\tiny(0.04, 0.05)} \\
    \bottomrule
  \end{tabular}
  }
\end{table}

\subsection{Analysis of the Individual Predictive Strength of the Last-Layer Imbalanced Forgetting Coefficients}\label{sec:individual_pred_strength}
\noindent Here, we evaluate how effectively each \emph{Last-layer Imbalanced Forgetting Coefficient} predicts the ranking of past classes by their forgetting across our benchmarks. Specifically, for each coefficient and each incremental step, we analyze (i) the extent to which past classes exhibiting higher values of that coefficient \emph{during} training on that step tend to experience greater forgetting \emph{at the end} of that step, and (ii) whether this relationship persists after controlling for all other coefficients. To this end, we measure (i) using Spearman’s rank correlation $\rho$, which captures the \emph{marginal association} between a coefficient and class-wise forgetting, and (ii) using partial Spearman’s rank correlation $\rho_p$, which measures the \emph{conditional association} after controlling for the remaining coefficients. We use Spearman-based correlations because they capture monotonic relationships without assuming linearity and are widely used for evaluating rankings in statistical and machine learning analyses.\footnote{Since each coefficient is measured during training on an incremental step, whereas forgetting is measured after training ends, these correlations quantify how well each coefficient predicts the eventual forgetting-based ranking of past classes.}

The results in Table \ref{tab:spearman_individual_coefficients} show that SIC consistently exhibits the strongest average marginal and conditional associations across our benchmarks. Importantly, its mean marginal associations are only slightly higher than the conditional ones, indicating that SIC captures predictive information that is largely unique relative to the other coefficients. Moreover, SIC also exhibits small standard deviations relative to its mean associations---the smallest among the coefficients---indicating that its predictive strength is highly stable across the incremental steps in our benchmarks. Overall, these results indicate that SIC consistently emerges as the strongest predictor among the proposed coefficients. In more detail, SIC achieves very strong mean associations across $\Omega_{\mathrm{C100}}$ ($\rho \approx 0.86\text{-}0.88$; $\rho_p \approx 0.73$–$0.78$). In contrast, its average predictive performance is weaker---though still considerable---across $\Omega_{\mathrm{TIN}}$ ($\rho \approx 0.56\text{--}0.68$; $\rho_p \approx 0.54\text{--}0.61$), with the difference in mean associations relative to $\Omega_{\mathrm{C100}}$ being statistically significant at the 95\% level. This suggests that the predictive strength of SIC decreases with CIL complexity, likely due to noisier and more complex training dynamics. Notably, NIC also experiences a similar trend. Furthermore, across both $\Omega_{\mathrm{C100}}$ and $\Omega_{\mathrm{TIN}}$, the mean associations of SIC decrease between the second and third incremental steps, although this decline is statistically significant (at the 95\% level) only in the latter benchmark. This degradation likely reflects increasing noise in the training dynamics of later incremental steps, as the classification task involves an increasing number of classes and the model progressively becomes saturated. A similar pattern is also observed for NIC.

Regarding NIC, it exhibits the second strongest average marginal associations across our benchmarks. As evidenced by its relatively small standard deviations, these associations are also consistent across incremental steps. However, unlike SIC, its mean predictive strength decreases considerably after controlling for the other coefficients. This indicates that a substantial portion of the predictive information captured by NIC overlaps with that of the remaining coefficients, rather than reflecting uniquely predictive signal. Crucially, Appendix \ref{sec:disentangling_NIC_pred_overlap_SIC_CIC} shows that this overlap arises primarily with SIC rather than CIC, indicating that NIC largely captures predictive patterns already reflected by SIC. In absolute terms, NIC achieves the following mean associations in $\Omega_{\mathrm{C100}}$ ($\rho \approx 0.65\text{-}0.75$; $\rho_p \approx 0.22\text{-}0.30$) and $\Omega_{\mathrm{TIN}}$ ($\rho \approx 0.46\text{-}0.53$; $\rho_p \approx 0.13\text{-}0.20$). Finally, CIC exhibits the weakest mean marginal associations and the largest standard deviations relative to those means. This indicates that it provides the lowest and least stable predictive signal among the derived coefficients. Although its predictive strength is modest, CIC captures complementary predictive information not explained by the remaining coefficients, as reflected by the small difference between its mean marginal and conditional associations. Notably, unlike SIC and NIC, the predictive strength of CIC increases with CIL complexity. The reasons for this behavior remain unclear and warrant further investigation. Concretely, CIC achieves the following mean associations in $\Omega_{\mathrm{C100}}$ ($\rho \approx 0.19\text{-}0.20$; $\rho_p \approx 0.10\text{-}0.12$) and $\Omega_{\mathrm{TIN}}$ ($\rho \approx 0.36\text{-}0.40$; $\rho_p \approx 0.36\text{-}0.37$).

To summarize, this subsection shows that the \emph{Last-Layer Imbalanced Forgetting Coefficients} differ substantially from one another in how effectively they predict the forgetting-based ranking of past classes across our benchmarks. In more detail, SIC consistently emerges as the strongest predictor, capturing predictive information that is largely unique relative to the other coefficients. This suggests that SIC captures the most prominent last-layer gradient-level source of interference identified by our analysis. NIC also shows substantial predictive strength but much of the predictive signal it captures overlaps with SIC. Finally, CIC emerges as the weakest predictor; nevertheless, it captures predictive information that is largely complementary to that provided by the other coefficients. For completeness, Appendix \ref{sec:fine_grained_analysis_indi_predi_strength} provides a finer-grained analysis of the individual predictive strength of these coefficients. Furthermore, Appendix \ref{sec:temporal_evolution_indiv_strength_imbalanced_forgetting} examines the effect of computing our coefficients using progressively more R-SGD steps from the start of training on an incremental step. The results show that using only the early portion of the R-SGD trajectory already yields performance comparable to that obtained with the entire trajectory. This indicates that the last-layer gradient-level interactions captured by our coefficients during the early training phase are almost as informative as those captured over the full training trajectory.

\begin{table*}[htbp]
  \centering
  \caption{Estimated mean $\pm$ standard deviation (with 95\% CIs) of the predictive strength achieved by each \emph{Last-layer Imbalanced Forgetting Coefficient} for the second and third incremental steps in $\Omega_{\mathrm{C100}}$ and $\Omega_{\mathrm{TIN}}$. The predictive strength is with respect to ranking past classes by their forgetting and is measured using Spearman’s correlation $\rho$ and partial Spearman’s correlation $\rho_p$.}
  \label{tab:spearman_individual_coefficients}
  \resizebox{0.7\textwidth}{!}{%
  \begin{tabular}{llccc}
    \toprule
    & & SIC & CIC & NIC\\
    \midrule
    \multicolumn{2}{l}{} & \multicolumn{3}{c}{$\Omega_{\mathrm{C100}}$}\\
    \midrule
    \multirow{2}{*}{$\Omega^2_{\mathrm{C100}}$}
      & $\rho$ & 0.86 {\tiny(0.85, 0.87)} $\pm$ 0.12 {\tiny(0.11, 0.14)} & 0.20 {\tiny(0.17, 0.23)} $\pm$ 0.29 {\tiny(0.27, 0.31)}  & 0.74 {\tiny(0.72, 0.75)} $\pm$ 0.15 {\tiny(0.13, 0.17)}\\
      & $\rho_p$  & 0.73 {\tiny(0.70, 0.75)} $\pm$ 0.24 {\tiny(0.21, 0.26)} & 0.12 {\tiny(0.08, 0.15)} $\pm$ 0.29 {\tiny(0.27, 0.32)} & 0.30 {\tiny(0.26, 0.33)} $\pm$ 0.28 {\tiny(0.26, 0.31)}  \\
      \midrule
    \multirow{2}{*}{$\Omega^{2;\,\{10\%, 20\%\}}_{\mathrm{C100}}$}
      & $\rho$ & 0.88 {\tiny(0.87, 0.89)} $\pm$ 0.13 {\tiny(0.11, 0.15)} & 0.19 {\tiny(0.14, 0.24)} $\pm$ 0.33 {\tiny(0.30, 0.36)} & 0.75 {\tiny(0.73, 0.77)} $\pm$ 0.17 {\tiny(0.15, 0.20)}\\
      & $\rho_p$  & 0.75 {\tiny(0.72, 0.78)} $\pm$ 0.26 {\tiny(0.24, 0.30)} & 0.12 {\tiny(0.07, 0.17)} $\pm$ 0.34 {\tiny(0.31, 0.37)} & 0.26 {\tiny(0.22, 0.31)} $\pm$ 0.32 {\tiny(0.30, 0.35)}  \\
    \midrule
    \multirow{2}{*}{$\Omega^3_{\mathrm{C100}}$}
      & $\rho$ & 0.87 {\tiny(0.86, 0.88)} $\pm$ 0.09 {\tiny(0.08, 0.10)} & 0.19 {\tiny(0.16, 0.23)} $\pm$ 0.28 {\tiny(0.26, 0.30)} & 0.65 {\tiny(0.64, 0.67)} $\pm$ 0.14 {\tiny(0.12, 0.15)}\\
      & $\rho_p$  & 0.78 {\tiny(0.76, 0.80)} $\pm$ 0.17 {\tiny(0.15, 0.20)} & 0.10 {\tiny(0.07, 0.13)} $\pm$ 0.24 {\tiny(0.22, 0.26)} & 0.22 {\tiny(0.19, 0.25)} $\pm$ 0.23 {\tiny(0.22, 0.25)}  \\
    \midrule
    \multicolumn{2}{l}{} & \multicolumn{3}{c}{$\Omega_{\mathrm{TIN}}$}\\
    \midrule
    \multirow{2}{*}{$\Omega^2_{\mathrm{TIN}}$}
      & $\rho$ & 0.63 {\tiny(0.61, 0.65)} $\pm$ 0.15 {\tiny(0.14, 0.16)} & 0.36 {\tiny(0.34, 0.38)} $\pm$ 0.19 {\tiny(0.18, 0.21)} & 0.53 {\tiny(0.51, 0.54)} $\pm$ 0.14 {\tiny(0.13, 0.15)}\\
      & $\rho_p$  & 0.55 {\tiny(0.53, 0.57)} $\pm$ 0.16 {\tiny(0.15, 0.17)} & 0.36 {\tiny(0.33, 0.38)} $\pm$ 0.21 {\tiny(0.19, 0.23)} & 0.20 {\tiny(0.17, 0.22)} $\pm$ 0.19 {\tiny(0.18, 0.21)}  \\
     \midrule 
     \multirow{2}{*}{$\Omega^{2;\,\{10\%, 20\%\}}_{\mathrm{TIN}}$}
      & $\rho$ & 0.68 {\tiny(0.66, 0.70)} $\pm$ 0.14 {\tiny(0.13, 0.16)} & 0.37 {\tiny(0.34, 0.40)} $\pm$ 0.21 {\tiny(0.20, 0.23)} & 0.53 {\tiny(0.51, 0.55)} $\pm$ 0.15 {\tiny(0.14, 0.16)}\\
      & $\rho_p$  & 0.61 {\tiny(0.59, 0.63)} $\pm$ 0.15 {\tiny(0.13, 0.17)} & 0.37 {\tiny(0.34, 0.40)} $\pm$ 0.23 {\tiny(0.21, 0.25)} & 0.17 {\tiny(0.14, 0.19)} $\pm$ 0.22 {\tiny(0.20, 0.24)}  \\
     \midrule 
     \multirow{2}{*}{$\Omega^3_{\mathrm{TIN}}$}
      & $\rho$ & 0.56 {\tiny(0.54, 0.58)} $\pm$ 0.14 {\tiny(0.13, 0.15)} & 0.40 {\tiny(0.37, 0.42)} $\pm$ 0.21 {\tiny(0.20, 0.23)} & 0.46 {\tiny(0.44, 0.47)} $\pm$ 0.14 {\tiny(0.12, 0.15)}\\
      & $\rho_p$  & 0.54 {\tiny(0.52, 0.55)} $\pm$ 0.14 {\tiny(0.12, 0.15)} & 0.37 {\tiny(0.34, 0.40)} $\pm$ 0.20 {\tiny(0.18, 0.21)} & 0.13 {\tiny(0.11, 0.15)} $\pm$ 0.16 {\tiny(0.14, 0.17)}  \\
    \bottomrule
  \end{tabular}
  }
\end{table*}

\subsection{Analysis of the Joint Predictive Strength of the Last-Layer Imbalanced Forgetting Coefficients}\label{sec:analysis_joint_pred_strength}
\noindent The results of the previous subsection (Table \ref{tab:spearman_individual_coefficients}) show that, although CIC and NIC are weaker predictors than SIC, they still capture some complementary predictive information. This suggests that jointly leveraging all \emph{Last-Layer Imbalanced Forgetting Coefficients} may improve prediction of the forgetting-based ranking of past classes, compared with using SIC alone. To evaluate this, we adopt a linear model that predicts class-wise forgetting as a weighted combination of SIC, CIC, and NIC. More formally, we model the degree of forgetting of a past class $c$ at the end of step $s_m$ as follows:
\begin{equation}
    \text{FG}^m_c = \beta_1\text{SIC}^m_c + \beta_2\text{CIC}^m_c + \beta_3\text{NIC}^m_c + \beta_0\,,
\end{equation}
with class ranking within an incremental step obtained by ordering the resulting scores. This joint model is assessed using a leave-one-out procedure applied separately to the second and third incremental steps in $\hat{\Omega}_{\mathrm{C100}}$ and $\hat{\Omega}_{\mathrm{TIN}}$. We refer to this procedure as the \emph{step-wise leave-one-out} (SW-LOO) protocol.

Specifically, for each $d \in \{\mathrm{C100}, \mathrm{TIN}\}$ and $k \in \{2,3\}$, we proceed as follows: at each round, a different incremental step in $\hat{\Omega}_d^k$ is held out for validation, while the joint model is fitted on the remaining steps in $\hat{\Omega}_d^k$ that share the same percentage of introduced classes and rehearsal retention rate as the excluded step.\footnote{We restrict the fitting set to incremental steps that share the same proportion of introduced classes and rehearsal retention rate as the held-out step to ensure that the predictor-target relationship is estimated under comparable conditions.} In more detail, the model is fitted using a mixed-effects formulation \cite{RN204} with random intercepts and slopes, which is statistically preferable to pooling all observations into a single dataset, as it explicitly models variability across incremental steps and respects the hierarchical data structure. The resulting model is then evaluated by how well it predicts the ranking of past classes by their forgetting on the held-out step. This is measured using Spearman's rank correlation between the model predictions and the ground-truth forgetting values of past classes.

Table \ref{tab:SW-LOO_protocol} reports the results of the above evaluation procedure, together with the predictive strength obtained on the same incremental steps using only raw SIC values.\footnote{The SIC results coincide with those in Table \ref{tab:spearman_individual_coefficients}, since the estimates in that table were computed using the sampled benchmarks $\hat{\Omega}_{\mathrm{C100}}$ and $\hat{\Omega}_{\mathrm{TIN}}$.} Notably, the joint model consistently exhibits stronger average predictive performance than SIC alone. In more detail, the improvement is modest on $\hat{\Omega}_{\mathrm{C100}}$, where the mean predictive strength of SIC is already strong. In contrast, the gap is substantially larger on $\hat{\Omega}_{\mathrm{TIN}}$. These findings indicate that jointly leveraging all coefficients yields stronger predictive performance than relying on SIC alone, with the advantage becoming more pronounced in more complex CIL settings. In absolute terms, the joint model achieves the following mean predictive performance on $\hat{\Omega}_{\mathrm{C100}}$ ($\rho \approx 0.88\text{-}0.91$) and $\hat{\Omega}_{\mathrm{TIN}}$  ($\rho \approx 0.72\text{-}0.77$), while that of SIC is as follows on $\hat{\Omega}_{\mathrm{C100}}$ ($\rho \approx 0.86\text{-}0.88$) and $\hat{\Omega}_{\mathrm{TIN}}$  ($\rho \approx 0.56\text{-}0.68$). Moreover, the joint model exhibits smaller standard deviations than SIC, indicating that its predictive strength is also more stable across incremental steps. Finally, as observed for SIC, the average predictive strength of the joint model is lower on $\hat{\Omega}_{\mathrm{TIN}}$ than on $\hat{\Omega}_{\mathrm{C100}}$, and decreases from the second to the third incremental steps in both sets. These results suggest that ranking predictions become more difficult both at later incremental steps and in more challenging CIL settings, not only when SIC is used alone but also when jointly leveraging all coefficients. We hypothesize that this occurs for the same reasons discussed for SIC in the previous subsection.

Notably, Appendix \ref{sec:complementary_joint_pred_strength} shows that the joint model also produces class-wise forgetting estimates that are reasonably close to the ground truth, with the average mean absolute error (MAE) ranging from 0.06 to 0.10 across our benchmarks. As further demonstrated in Appendix \ref{sec:complementary_joint_pred_strength}, although the weights of this model vary across benchmark partitions,\footnote{This suggests that the relationship between the \emph{Last-Layer Imbalanced Forgetting Coefficients} and class-wise forgetting depends on the benchmark, percentage of newly introduced classes and rehearsal-retention rate.} they are consistently positive. This indicates that each \emph{Last-layer Imbalanced Forgetting Coefficient} has a positive partial effect on class-wise forgetting.

In summary, this subsection shows that a linear model incorporating all \emph{Last-Layer Imbalanced Forgetting Coefficients} predicts the forgetting-based ranking of past classes more accurately than SIC alone, indicating that despite their lower individual predictive power, CIC and NIC capture complementary predictive information beyond that of SIC. Furthermore, the reliable predictive performance of this linear model supports interpreting our coefficients as offering a plausible mechanistic account linking last-layer gradient-level interactions during training to class-level forgetting outcomes.

\begin{table}[htbp]
  \centering
  \caption{Mean $\pm$ standard deviation of the predictive strength achieved by a linear model using all \emph{Last-Layer Imbalanced Forgetting Coefficients}, evaluated under the SW-LOO protocol on the second and third incremental steps in $\hat{\Omega}_{\mathrm{C100}}$ and $\hat{\Omega}_{\mathrm{TIN}}$. The predictive strength is with respect to ranking past classes by their forgetting and is measured using Spearman's correlation $\rho$. The predictive strength obtained on the same incremental steps using only the raw SIC values is also reported.}
  \label{tab:SW-LOO_protocol}
  \renewcommand{\arraystretch}{0.5}
  \resizebox{0.60\linewidth}{!}{%
  \begin{tabular}{lcc}
    \toprule
    & All coefficients & SIC \\
    \midrule
    & \multicolumn{2}{c}{ $\hat{\Omega}_{\mathrm{C100}}$}\\
    \midrule
    $\hat{\Omega}^2_{\mathrm{C100}}$
        & 0.90 $\pm$ 0.08 & 0.86 $\pm$ 0.12 \\ 
    \midrule
    $\hat{\Omega}^{2;\,\{10\%, 20\%\}}_{\mathrm{C100}}$
        & 0.91 $\pm$ 0.09 & 0.88 $\pm$ 0.13 \\
    \midrule
    $\hat{\Omega}^3_{\mathrm{C100}}$
        & 0.88 $\pm$ 0.08 & 0.87 $\pm$ 0.09 \\
    \midrule
    & \multicolumn{2}{c}{ $\hat{\Omega}_{\mathrm{TIN}}$}\\
    \midrule
    $\hat{\Omega}^2_{\mathrm{TIN}}$
        & 0.74 $\pm$ 0.11 & 0.63 $\pm$ 0.15 \\
    \midrule
    $\hat{\Omega}^{2; \, \{10\%, 20\%\}}_{\mathrm{TIN}}$
        & 0.77 $\pm$ 0.11 & 0.68 $\pm$ 0.14 \\
    \midrule
    $\hat{\Omega}^3_{\mathrm{TIN}}$
        & 0.72 $\pm$ 0.10 & 0.56 $\pm$ 0.14 \\
    \bottomrule
\end{tabular}
  }
\end{table}

\subsection{Analysis of the Relationship between NIC and SIC}\label{sec:ana_relationship_SIC_NIC}
\noindent Section \ref{sec:individual_pred_strength} shows that SIC is the strongest predictor, suggesting that it captures the dominant gradient-level source of interference among the \emph{Last-Layer Imbalanced Forgetting Coefficients}. Motivated by this, we here investigate why certain past classes exhibit higher SIC values than others.

In this regard, as shown in Table \ref{tab:spearman_SIC_NIC}, we observe that NIC is closely and consistently associated with SIC across our benchmarks.\footnote{For completeness, Appendix \ref{sec:pairwise_associations_imbalanced_forgetting} reports the pairwise associations among SIC, CIC, and NIC across our benchmarks.} That is, past classes experiencing greater interference from new classes during an incremental step, as measured by NIC, also tend to exhibit higher self-induced interference arising from the bias between their rehearsal and original last-layer gradients. Importantly, the mean NIC-SIC association follows the same pattern observed previously: it is lower on $\Omega_{\mathrm{TIN}}$ than on $\Omega_{\mathrm{C100}}$, and decreases from the second to the third incremental steps in both benchmarks.\footnote{A more fine-grained analysis of the association between SIC and NIC is provided in Appendix \ref{sec:finer_grained_anal_assoc_NIC_SIC}.}

We further posit that the observed association arises because the source of interference captured by NIC influences that captured by SIC. Specifically, we hypothesize that higher NIC values for a past class during an incremental step lead to higher SIC by steering the training trajectory toward regions of the parameter space where rehearsal samples from that class induce greater interfering bias in the last layer. To evaluate this hypothesis, we conduct a set of controlled experiments in which, for randomly selected incremental steps in our benchmarks, we repeatedly re-run their training using different sets of new classes while holding all other factors unchanged (i.e., the rehearsal set, the random seed, and the initial network parameters). Notably, we observe that across repeated re-runs of the same step, the rehearsal set of a past class tends to yield higher SIC in runs where the new classes interfere more strongly with that class, as measured by NIC. More specifically, for the selected incremental steps, the mean class-wise Spearman's correlation between NIC and SIC computed across re-runs ranges from 0.40 to 0.76. Since all factors other than the new classes are identical across re-runs, these findings provide empirical support for our hypothesis. From a causal perspective, they therefore suggest that SIC may mediate the effect of NIC on class-wise forgetting, which would explain why NIC exhibits a high Spearman's correlation with class-wise forgetting, but this association drops dramatically when controlling for SIC (Appendix \ref{sec:disentangling_NIC_pred_overlap_SIC_CIC}). Further details and results from these controlled experiments are provided in Appendix \ref{sec:controlled_exps_NIC_SIC}.

In summary, this subsection investigates why certain past classes exhibit larger SIC than others and identifies NIC as a potential contributing factor. However, the imperfect association between NIC and SIC suggests that NIC alone does not fully explain the observed discrepancies. We further hypothesize that the extent to which a class can be effectively represented by a limited number of rehearsal samples also influences its SIC. In particular, classes with lower distributional complexity in the original training data are likely to incur lower SIC than those with higher complexity under the same level of NIC. Exploring the joint effect of this factor and NIC on SIC is left for future work.

\begin{table}[htbp]
  \centering
  \caption{Estimated mean $\pm$ standard deviation (with 95\% CIs) of the association between NIC and SIC, measured using Spearman's correlation $\rho$, for the second and third incremental steps in $\Omega_{\mathrm{C100}}$ and $\Omega_{\mathrm{TIN}}$.}
  \label{tab:spearman_SIC_NIC}
  \renewcommand{\arraystretch}{0.5}
  \resizebox{0.65\linewidth}{!}{%
  \begin{tabular}{lcc}
    \toprule
    &  & $\rho$ \\
    \midrule
    \multicolumn{3}{c}{ $\Omega_{\mathrm{C100}}$}\\
    \midrule
    $\Omega^2_{\mathrm{C100}}$ &  & 0.71 {\tiny(0.69, 0.73)} $\pm$ 0.17 {\tiny(0.16, 0.19)} \\ 
    \midrule
    $\Omega^{2;\,\{10\%, 20\%\}}_{\mathrm{C100}}$ &  & 0.74 {\tiny(0.72, 0.76)} $\pm$ 0.19 {\tiny(0.17, 0.22)}\\
    \midrule
    $\Omega^3_{\mathrm{C100}}$ &  & 0.62 {\tiny(0.60, 0.64)} $\pm$ 0.15 {\tiny(0.13, 0.17)}\\ 
    \midrule
    \multicolumn{3}{c}{ $\Omega_{\mathrm{TIN}}$}\\
    \midrule
    $\Omega^2_{\mathrm{TIN}}$ &  & 0.57 {\tiny(0.56, 0.59)} $\pm$ 0.14 {\tiny(0.13, 0.15)}\\
    \midrule
    $\Omega^{2;\, \{10\%, 20\%\}}_{\mathrm{TIN}}$ &  & 0.58 {\tiny(0.56, 0.60)} $\pm$ 0.16 {\tiny(0.14, 0.17)}\\
    \midrule
    $\Omega^3_{\mathrm{TIN}}$ &  & 0.46 {\tiny(0.45, 0.48)} $\pm$ 0.14 {\tiny(0.13, 0.15)}\\
    \bottomrule
\end{tabular}
  }
\end{table}

\section{Conclusion}\label{sec:Conclusion}
\noindent This work empirically demonstrates that imbalanced forgetting---a phenomenon whereby some classes experience substantially greater forgetting than others---arises systematically and severely in rehearsal-based CIL, despite equal per-class allocation of rehearsal samples. Motivated by the practical relevance of this phenomenon, we further investigate why certain past classes are forgotten more than others. To this end, we construct, from a principled analysis based on the \emph{Class-Wise R-SGD lemma}, three last-layer coefficients---SIC, CIC, and NIC---that capture different gradient-level sources of interference affecting each past class \emph{during training} on an incremental step. We then demonstrate that, taken together, they reliably predict how past classes will rank in terms of forgetting \emph{at the end} of that step. Although predictive performance alone cannot demonstrate causality, these findings are consistent with interpreting the coefficients as a plausible mechanistic explanation linking last-layer gradient-level interactions during training to class-level forgetting outcomes, and hence to imbalanced forgetting.

Overall, our findings offer actionable insights for the CIL community and inform future research directions. In particular, the identification of imbalanced forgetting as a systematic failure mode---despite equal per-class allocation of rehearsal samples---suggests that future rehearsal-based methods should explicitly account for class-wise disparities in forgetting. Promoting more uniform retention across past classes is not only desirable from a fairness standpoint, but also beneficial for overall performance: reducing forgetting in the most forgotten classes, while keeping performance on the others roughly unchanged, effectively lowers overall forgetting. Therefore, addressing imbalanced forgetting is not a secondary concern, but a principled pathway toward mitigating catastrophic forgetting itself.

Furthermore, our analysis points to promising directions for achieving this goal based on reducing class-wise disparities in the identified last-layer gradient-level sources of interference. Among these, prioritizing disparities in the source of interference captures by SIC is especially critical, as it appears to be the dominant one. Since SIC captures interference induced by the mismatch between rehearsal and original last-layer gradients for a given class, one natural strategy among several possible approaches is to adapt the allocation of rehearsal samples based on SIC values. Specifically, allocating more samples to classes with higher SIC may help mitigate this imbalance, suggesting that uniform per-class rehearsal allocation is suboptimal and that adaptive, class-aware allocation schemes may be more effective. However, it is unclear whether the SIC values that past classes will experience during the next incremental step can be reliably anticipated at the end of the current one, which is when the rehearsal set must be constructed. This challenge arises because, although we hypothesize that SIC depends in part on the distributional complexity of the original training data, it appears to be strongly influenced by NIC, which in turn depends on the nature of newly introduced classes. Consequently, rather than directly targeting SIC (e.g., via unequal rehearsal allocation), balancing NIC across classes---e.g., through constrained optimization---may provide a more practical indirect pathway for reducing SIC disparities. More broadly, our findings call for a shift from uniform mitigation strategies to approaches that explicitly account for class-dependent dynamics in catastrophic forgetting.

\bibliographystyle{IEEEtran}
\bibliography{refs}

@inproceedings{RN186,
   author = {Ahn, Hongjoon and Kwak, Jihwan and Lim, Subin and Bang, Hyeonsu and Kim, Hyojun and Moon, Taesup},
   title = {Ss-il: Separated softmax for incremental learning},
   booktitle = {Proceedings of the IEEE/CVF International conference on computer vision},
   pages = {844-853},
   year = {2021},
   type = {Conference Proceedings}
}

@article{RN9,
   author = {Aljundi, Rahaf and Lin, Min and Goujaud, Baptiste and Bengio, Yoshua},
   title = {Gradient based sample selection for online continual learning},
   journal = {Advances in neural information processing systems},
   volume = {32},
   year = {2019},
   type = {Journal Article}
}

@inproceedings{RN165,
   author = {Bang, Jihwan and Kim, Heesu and Yoo, YoungJoon and Ha, Jung-Woo and Choi, Jonghyun},
   title = {Rainbow memory: Continual learning with a memory of diverse samples},
   booktitle = {Proceedings of the IEEE/CVF conference on computer vision and pattern recognition},
   pages = {8218-8227},
   year = {2021},
   type = {Conference Proceedings}
}

@inproceedings{RN185,
   author = {Belouadah, Eden and Popescu, Adrian},
   title = {Scail: Classifier weights scaling for class incremental learning},
   booktitle = {Proceedings of the IEEE/CVF winter conference on applications of computer vision},
   pages = {1266-1275},
   year = {2020},
   type = {Conference Proceedings}
}

@article{RN22,
   author = {Benkő, Beatrix},
   title = {Example forgetting and rehearsal in continual learning},
   journal = {Pattern Recognition Letters},
   volume = {179},
   pages = {65-72},
   ISSN = {0167-8655},
   year = {2024},
   type = {Journal Article}
}

@article{RN164,
   author = {Borsos, Zalán and Mutny, Mojmir and Krause, Andreas},
   title = {Coresets via bilevel optimization for continual learning and streaming},
   journal = {Advances in neural information processing systems},
   volume = {33},
   pages = {14879-14890},
   year = {2020},
   type = {Journal Article}
}

@article{RN7,
   author = {Buzzega, Pietro and Boschini, Matteo and Porrello, Angelo and Abati, Davide and Calderara, Simone},
   title = {Dark experience for general continual learning: a strong, simple baseline},
   journal = {Advances in neural information processing systems},
   volume = {33},
   pages = {15920-15930},
   year = {2020},
   type = {Journal Article}
}

@inproceedings{RN176,
   author = {Caccia, Lucas and Aljundi, Rahaf and Asadi, Nader and Tuytelaars, Tinne and Pineau, Joelle and Belilovsky, Eugene},
   title = {New Insights on Reducing Abrupt Representation Change in Online Continual Learning},
   booktitle = {ICLR 2022 - 10th Conference on Learning Representations},
   year = {2022},
   type = {Conference Proceedings}
}

@inproceedings{RN161,
   author = {Castro, Francisco M and Marín-Jiménez, Manuel J and Guil, Nicolás and Schmid, Cordelia and Alahari, Karteek},
   title = {End-to-end incremental learning},
   booktitle = {Proceedings of the European conference on computer vision (ECCV)},
   pages = {233-248},
   year = {2018},
   type = {Conference Proceedings}
}

@inproceedings{RN163,
   author = {Chaudhry, Arslan and Dokania, Puneet K and Ajanthan, Thalaiyasingam and Torr, Philip HS},
   title = {Riemannian walk for incremental learning: Understanding forgetting and intransigence},
   booktitle = {Proceedings of the European conference on computer vision (ECCV)},
   pages = {532-547},
   year = {2018},
   type = {Conference Proceedings}
}

@inproceedings{RN121,
   author = {Chaudhry, A. and Marc'Aurelio, R. and Rohrbach, M. and Elhoseiny, M.},
   title = {Efficient lifelong learning with A-GEM},
   booktitle = {7th International Conference on Learning Representations, ICLR 2019},
   publisher = {International Conference on Learning Representations, ICLR},
   year = {2019},
   type = {Conference Proceedings}
}

@inproceedings{RN202,
   author = {Deng, Jia and Dong, Wei and Socher, Richard and Li, Li-Jia and Li, Kai and Fei-Fei, Li},
   title = {Imagenet: A large-scale hierarchical image database},
   booktitle = {2009 IEEE conference on computer vision and pattern recognition},
   publisher = {Ieee},
   pages = {248-255},
   ISBN = {1424439922},
   year = {2009},
   type = {Conference Proceedings}
}

@article{RN196,
   author = {Efron, Bradley},
   title = {Better bootstrap confidence intervals},
   journal = {Journal of the American statistical Association},
   volume = {82},
   number = {397},
   pages = {171-185},
   ISSN = {0162-1459},
   year = {1987},
   type = {Journal Article}
}

@inproceedings{RN157,
   author = {Evron, Itay and Moroshko, Edward and Ward, Rachel and Srebro, Nathan and Soudry, Daniel},
   title = {How catastrophic can catastrophic forgetting be in linear regression?},
   booktitle = {Conference on Learning Theory},
   publisher = {PMLR},
   pages = {4028-4079},
   ISBN = {2640-3498},
   year = {2022},
   type = {Conference Proceedings}
}

@article{RN12,
   author = {Feldman, Dan},
   title = {Introduction to core-sets: an updated survey},
   journal = {arXiv preprint arXiv:2011.09384},
   year = {2020},
   type = {Journal Article}
}

@book{RN204,
   author = {Gelman, Andrew and Hill, Jennifer},
   title = {Data analysis using regression and multilevel/hierarchical models},
   publisher = {Cambridge university press},
   ISBN = {052168689X},
   year = {2007},
   type = {Book}
}

@article{RN10,
   author = {Ghadimi, Saeed and Lan, Guanghui},
   title = {Stochastic first-and zeroth-order methods for nonconvex stochastic programming},
   journal = {SIAM journal on optimization},
   volume = {23},
   number = {4},
   pages = {2341-2368},
   ISSN = {1052-6234},
   year = {2013},
   type = {Journal Article}
}

@inproceedings{RN154,
   author = {Goldfarb, Daniel and Evron, Itay and Weinberger, Nir and Soudry, Daniel and Hand, Paul},
   title = {The Joint Effect of Task Similarity and Overparameterization on Catastrophic Forgetting--An Analytical Model},
   booktitle = {ICLR 2024 - 12th International Conference on Learning Representations},
   year = {2024},
   type = {Conference Proceedings}
}

@article{RN19,
   author = {Goodfellow, Ian J and Mirza, Mehdi and Xiao, Da and Courville, Aaron and Bengio, Yoshua},
   title = {An empirical investigation of catastrophic forgetting in gradient-based neural networks},
   journal = {arXiv preprint arXiv:1312.6211},
   year = {2013},
   type = {Journal Article}
}

@article{RN11,
   author = {Harun, Md Yousuf and Gallardo, Jhair and Hayes, Tyler L and Kemker, Ronald and Kanan, Christopher},
   title = {Siesta: Efficient online continual learning with sleep},
   journal = {arXiv preprint arXiv:2303.10725},
   year = {2023},
   type = {Journal Article}
}

@inproceedings{RN184,
   author = {He, Kaiming and Zhang, Xiangyu and Ren, Shaoqing and Sun, Jian},
   title = {Deep residual learning for image recognition},
   booktitle = {Proceedings of the IEEE conference on computer vision and pattern recognition},
   pages = {770-778},
   year = {2016},
   type = {Conference Proceedings}
}

@article{RN172,
   author = {Hinton, Geoffrey and Vinyals, Oriol and Dean, Jeff},
   title = {Distilling the knowledge in a neural network},
   journal = {arXiv preprint arXiv:1503.02531},
   year = {2015},
   type = {Journal Article}
}

@inproceedings{RN160,
   author = {Hou, Saihui and Pan, Xinyu and Loy, Chen Change and Wang, Zilei and Lin, Dahua},
   title = {Learning a unified classifier incrementally via rebalancing},
   booktitle = {Proceedings of the IEEE/CVF conference on computer vision and pattern recognition},
   pages = {831-839},
   year = {2019},
   type = {Conference Proceedings}
}

@inproceedings{RN167,
   author = {Iscen, Ahmet and Zhang, Jeffrey and Lazebnik, Svetlana and Schmid, Cordelia},
   title = {Memory-efficient incremental learning through feature adaptation},
   booktitle = {Computer Vision–ECCV 2020: 16th European Conference, Glasgow, UK, August 23–28, 2020, Proceedings, Part XVI 16},
   publisher = {Springer},
   pages = {699-715},
   ISBN = {3030585166},
   year = {2020},
   type = {Conference Proceedings}
}

@inproceedings{RN170,
   author = {Jiang, Jian and Cetin, Edoardo and Celiktutan, Oya},
   title = {Ib-drr-incremental learning with information-back discrete representation replay},
   booktitle = {Proceedings of the IEEE/CVF Conference on Computer Vision and Pattern Recognition},
   pages = {3533-3542},
   year = {2021},
   type = {Conference Proceedings}
}

@inproceedings{RN171,
   author = {Jodelet, Quentin and Liu, Xin and Phua, Yin Jun and Murata, Tsuyoshi},
   title = {Class-incremental learning using diffusion model for distillation and replay},
   booktitle = {Proceedings of the IEEE/CVF International Conference on Computer Vision},
   pages = {3425-3433},
   year = {2023},
   type = {Conference Proceedings}
}

@article{RN100,
   author = {Krizhevsky, Alex and Hinton, Geoffrey},
   title = {Learning multiple layers of features from tiny images},
   year = {2009},
   type = {Journal Article}
}

@article{RN201,
   author = {Le, Yann and Yang, Xuan},
   title = {Tiny imagenet visual recognition challenge},
   journal = {CS 231N},
   volume = {7},
   number = {7},
   pages = {3},
   year = {2015},
   type = {Journal Article}
}

@inproceedings{RN153,
   author = {Lee, Sebastian and Goldt, Sebastian and Saxe, Andrew},
   title = {Continual learning in the teacher-student setup: Impact of task similarity},
   booktitle = {International Conference on Machine Learning},
   publisher = {PMLR},
   pages = {6109-6119},
   ISBN = {2640-3498},
   year = {2021},
   type = {Conference Proceedings}
}

@inproceedings{RN205,
   author = {Li, Ziyan and Hiratani, Naoki},
   title = {Optimal Task Order for Continual Learning of Multiple Tasks},
   booktitle = {Forty-second International Conference on Machine Learning},
   type = {Conference Proceedings}
}

@article{RN58,
   author = {Li, Zhizhong and Hoiem, Derek},
   title = {Learning without forgetting},
   journal = {IEEE transactions on pattern analysis and machine intelligence},
   volume = {40},
   number = {12},
   pages = {2935-2947},
   ISSN = {0162-8828},
   year = {2017},
   type = {Journal Article}
}

@inproceedings{RN156,
   author = {Lin, Sen and Ju, Peizhong and Liang, Yingbin and Shroff, Ness},
   title = {Theory on forgetting and generalization of continual learning},
   booktitle = {International Conference on Machine Learning},
   publisher = {PMLR},
   pages = {21078-21100},
   ISBN = {2640-3498},
   year = {2023},
   type = {Conference Proceedings}
}

@inproceedings{RN119,
   author = {Lopez-Paz, D. and Ranzato, M.},
   title = {Gradient episodic memory for continual learning},
   booktitle = {Advances in Neural Information Processing Systems},
   editor = {Guyon, I. and Fergus, R. and Wallach, H. and Wallach, H. and Guyon, I. and Vishwanathan, S. V. N. and von Luxburg, U. and Garnett, R. and Vishwanathan, S. V. N. and Bengio, S. and Fergus, R.},
   publisher = {Neural information processing systems foundation},
   volume = {2017-December},
   pages = {6468-6477},
   ISBN = {10495258 (ISSN)},
   year = {2017},
   type = {Conference Proceedings}
}

@inproceedings{RN200,
   author = {Loshchilov, I. and Hutter, F.},
   title = {SGDR: Stochastic gradient descent with warm restarts},
   booktitle = {5th International Conference on Learning Representations, ICLR 2017 - Conference Track Proceedings},
   publisher = {International Conference on Learning Representations, ICLR},
   year = {2017},
   type = {Conference Proceedings}
}

@article{RN198,
   author = {MacKinnon, James G and White, Halbert},
   title = {Some heteroskedasticity-consistent covariance matrix estimators with improved finite sample properties},
   journal = {Journal of econometrics},
   volume = {29},
   number = {3},
   pages = {305-325},
   ISSN = {0304-4076},
   year = {1985},
   type = {Journal Article}
}

@article{RN66,
   author = {McClelland, James L and McNaughton, Bruce L and O'Reilly, Randall C},
   title = {Why there are complementary learning systems in the hippocampus and neocortex: insights from the successes and failures of connectionist models of learning and memory},
   journal = {Psychological review},
   volume = {102},
   number = {3},
   pages = {419},
   ISSN = {1939-1471},
   year = {1995},
   type = {Journal Article}
}

@inbook{RN77,
   author = {McCloskey, Michael and Cohen, Neal J},
   title = {Catastrophic interference in connectionist networks: The sequential learning problem},
   booktitle = {Psychology of learning and motivation},
   publisher = {Elsevier},
   volume = {24},
   pages = {109-165},
   ISBN = {0079-7421},
   year = {1989},
   type = {Book Section}
}

@article{RN67,
   author = {O’Reilly, Randall C and Bhattacharyya, Rajan and Howard, Michael D and Ketz, Nicholas},
   title = {Complementary learning systems},
   journal = {Cognitive science},
   volume = {38},
   number = {6},
   pages = {1229-1248},
   ISSN = {0364-0213},
   year = {2014},
   type = {Journal Article}
}

@article{RN151,
   author = {O'Reilly, Randall C and Norman, Kenneth A},
   title = {Hippocampal and neocortical contributions to memory: Advances in the complementary learning systems framework},
   journal = {Trends in cognitive sciences},
   volume = {6},
   number = {12},
   pages = {505-510},
   ISSN = {1364-6613},
   year = {2002},
   type = {Journal Article}
}

@inproceedings{RN169,
   author = {Ostapenko, Oleksiy and Puscas, Mihai and Klein, Tassilo and Jahnichen, Patrick and Nabi, Moin},
   title = {Learning to remember: A synaptic plasticity driven framework for continual learning},
   booktitle = {Proceedings of the IEEE/CVF conference on computer vision and pattern recognition},
   pages = {11321-11329},
   year = {2019},
   type = {Conference Proceedings}
}

@article{RN155,
   author = {Ramasesh, Vinay V and Dyer, Ethan and Raghu, Maithra},
   title = {Anatomy of catastrophic forgetting: Hidden representations and task semantics},
   journal = {arXiv preprint arXiv:2007.07400},
   year = {2020},
   type = {Journal Article}
}

@article{RN150,
   author = {Ratcliff, Roger},
   title = {Connectionist models of recognition memory: constraints imposed by learning and forgetting functions},
   journal = {Psychological review},
   volume = {97},
   number = {2},
   pages = {285},
   ISSN = {1939-1471},
   year = {1990},
   type = {Journal Article}
}

@inproceedings{RN113,
   author = {Rebuffi, S. A. and Kolesnikov, A. and Sperl, G. and Lampert, C. H.},
   title = {iCaRL: Incremental classifier and representation learning},
   booktitle = {Proceedings - 30th IEEE Conference on Computer Vision and Pattern Recognition, CVPR 2017},
   publisher = {Institute of Electrical and Electronics Engineers Inc.},
   volume = {2017-January},
   pages = {5533-5542},
   ISBN = {978-153860457-1 (ISBN)},
   DOI = {10.1109/CVPR.2017.587},
   year = {2017},
   type = {Conference Proceedings}
}

@article{RN158,
   author = {Robins, Anthony},
   title = {Catastrophic forgetting, rehearsal and pseudorehearsal},
   journal = {Connection Science},
   volume = {7},
   number = {2},
   pages = {123-146},
   ISSN = {0954-0091},
   year = {1995},
   type = {Journal Article}
}

@article{RN168,
   author = {Shin, Hanul and Lee, Jung Kwon and Kim, Jaehong and Kim, Jiwon},
   title = {Continual learning with deep generative replay},
   journal = {Advances in neural information processing systems},
   volume = {30},
   year = {2017},
   type = {Journal Article}
}

@inproceedings{RN6,
   author = {Tiwari, Rishabh and Killamsetty, Krishnateja and Iyer, Rishabh and Shenoy, Pradeep},
   title = {Gcr: Gradient coreset based replay buffer selection for continual learning},
   booktitle = {Proceedings of the IEEE/CVF Conference on Computer Vision and Pattern Recognition},
   pages = {99-108},
   year = {2022},
   type = {Conference Proceedings}
}

@article{RN8,
   author = {Toneva, Mariya and Sordoni, Alessandro and Combes, Remi Tachet des and Trischler, Adam and Bengio, Yoshua and Gordon, Geoffrey J},
   title = {An empirical study of example forgetting during deep neural network learning},
   journal = {arXiv preprint arXiv:1812.05159},
   year = {2018},
   type = {Journal Article}
}

@inproceedings{RN178,
   author = {Verwimp, Eli and De Lange, Matthias and Tuytelaars, Tinne},
   title = {Rehearsal revealed: The limits and merits of revisiting samples in continual learning},
   booktitle = {Proceedings of the IEEE/CVF International Conference on Computer Vision},
   pages = {9385-9394},
   year = {2021},
   type = {Conference Proceedings}
}

@inproceedings{RN174,
   author = {Wang, Fu-Yun and Zhou, Da-Wei and Ye, Han-Jia and Zhan, De-Chuan},
   title = {Foster: Feature boosting and compression for class-incremental learning},
   booktitle = {European conference on computer vision},
   publisher = {Springer},
   pages = {398-414},
   year = {2022},
   type = {Conference Proceedings}
}

@inproceedings{RN181,
   author = {Wei, Colin and Ma, Tengyu},
   title = {Improved sample complexities for deep neural networks and robust classification via an all-layer margin},
   booktitle = {International Conference on Learning Representations},
   year = {2020},
   type = {Conference Proceedings}
}

@inproceedings{RN173,
   author = {Wu, Yue and Chen, Yinpeng and Wang, Lijuan and Ye, Yuancheng and Liu, Zicheng and Guo, Yandong and Fu, Yun},
   title = {Large scale incremental learning},
   booktitle = {Proceedings of the IEEE/CVF conference on computer vision and pattern recognition},
   pages = {374-382},
   year = {2019},
   type = {Conference Proceedings}
}

@inproceedings{RN152,
   author = {Xu, Shixiong and Meng, Gaofeng and Nie, Xing and Ni, Bolin and Fan, Bin and Xiang, Shiming},
   title = {Defying imbalanced forgetting in class incremental learning},
   booktitle = {Proceedings of the AAAI Conference on Artificial Intelligence},
   volume = {38},
   pages = {16211-16219},
   ISBN = {2374-3468},
   year = {2024},
   type = {Conference Proceedings}
}

@inproceedings{RN18,
   author = {Yan, Shipeng and Xie, Jiangwei and He, Xuming},
   title = {Der: Dynamically expandable representation for class incremental learning},
   booktitle = {Proceedings of the IEEE/CVF conference on computer vision and pattern recognition},
   pages = {3014-3023},
   year = {2021},
   type = {Conference Proceedings}
}

@article{RN177,
   author = {Yu, Longhui and Hu, Tianyang and Lanqing, HONG and Liu, Zhen and Weller, Adrian and Liu, Weiyang},
   title = {Continual Learning by Modeling Intra-Class Variation},
   journal = {Transactions on Machine Learning Research},
   ISSN = {2835-8856},
   year = {2023},
   type = {Journal Article}
}

@article{RN206,
   author = {Yu, Tianhe and Kumar, Saurabh and Gupta, Abhishek and Levine, Sergey and Hausman, Karol and Finn, Chelsea},
   title = {Gradient surgery for multi-task learning},
   journal = {Advances in neural information processing systems},
   volume = {33},
   pages = {5824-5836},
   year = {2020},
   type = {Journal Article}
}

@inproceedings{RN175,
   author = {Zhan, Da-Wei Zhou and Qi-Wei Wang and Han-Jia Ye and De-Chuan},
   title = {A Model or 603 Exemplars: Towards Memory-Efficient Class-Incremental Learning},
   booktitle = {ICLR 2023 - 11th Conference on Learning Representations},
   year = {2023},
   type = {Conference Proceedings}
}

@inproceedings{RN162,
   author = {Zhao, Bowen and Xiao, Xi and Gan, Guojun and Zhang, Bin and Xia, Shu-Tao},
   title = {Maintaining discrimination and fairness in class incremental learning},
   booktitle = {Proceedings of the IEEE/CVF conference on computer vision and pattern recognition},
   pages = {13208-13217},
   year = {2020},
   type = {Conference Proceedings}
}

@article{RN166,
   author = {Zhao, Hanbin and Wang, Hui and Fu, Yongjian and Wu, Fei and Li, Xi},
   title = {Memory-efficient class-incremental learning for image classification},
   journal = {IEEE Transactions on Neural Networks and Learning Systems},
   volume = {33},
   number = {10},
   pages = {5966-5977},
   ISSN = {2162-237X},
   year = {2021},
   type = {Journal Article}
}

@inproceedings{RN180,
   author = {Zheng, Bowen and Zhou, Da-Wei and Ye, Han-Jia and Zhan, De-Chuan},
   title = {Multi-layer rehearsal feature augmentation for class-incremental learning},
   booktitle = {Forty-first International Conference on Machine Learning},
   year = {2024},
   type = {Conference Proceedings}
}

@article{RN183,
   author = {Zhou, Da-Wei and Wang, Qi-Wei and Qi, Zhi-Hong and Ye, Han-Jia and Zhan, De-Chuan and Liu, Ziwei},
   title = {Class-incremental learning: A survey},
   journal = {IEEE Transactions on Pattern Analysis and Machine Intelligence},
   ISSN = {0162-8828},
   year = {2024},
   type = {Journal Article}
}

\appendices

\section{Rehearsal-based Approaches in Class-Incremental Learning}\label{sec:extensions_rehearsal}
\noindent In addition to the standard formulation of rehearsal considered in the main body of this paper (see Section \ref{sec:rehearsal_description}), several other variants have been proposed in the CIL literature. For completeness, we provide a brief overview of these below.

In order to enhance memory efficiency of rehearsal in CIL scenarios, two approaches, proposed in \cite{RN167, RN166}, retain features and low-fidelity images, respectively, instead of keeping raw ones. Other works \cite{RN168, RN169, RN170, RN171} propose avoiding the storage of past samples altogether, opting instead to use generative models to generate past samples. These generative methods perform well on simple datasets but struggle significantly with more complex ones.

Instead of focusing on memory efficiency, other approaches aim to enhance rehearsal by incorporating additional techniques to mitigate forgetting. Inspired by \cite{RN172, RN58}, the knowledge distillation methods \cite{RN113, RN161, RN160, RN7} leverage the samples in the rehearsal set for computing a knowledge distillation term. This additional term facilitates knowledge transfer from the previous version of the model to the current one, thereby helping the current model retain previously learned information. A second family of approaches \cite{RN160, RN173, RN162} improves rehearsal by mitigating the bias of the final classifier layer toward new classes. LUCIR \cite{RN160} uses a cosine classifier to reduce the impact of larger weights assigned to new classes. BiC \cite{RN173} adds an auxiliary rectification layer to adjust the output predictions. WA \cite{RN162} normalizes the classifier weights after every optimization step and applies weight clipping to ensure that predicted probabilities remain proportional to the classifier weights.

While effective, deep neural networks eventually saturate as the number of incrementally learned classes exceeds their capacity, leading to interference and degradation of previously acquired knowledge. To overcome this, dynamic network approaches \cite{RN18, RN174, RN175} adaptively increase network capacity along with preserving a subset of past data for rehearsal. DER \cite{RN18} introduces a new backbone at each incremental step and aggregates the features from all backbones, which are then passed to a shared final classifier. FOSTER \cite{RN174} improves efficiency by employing a model compression process based on knowledge distillation \cite{RN172}. MEMO \cite{RN175} further optimizes the expansion protocol by selectively expanding only specialized blocks.

For a more comprehensive survey of the rehearsal-based approaches proposed in the CIL literature, the reader is referred to \cite{RN183}.

\section{Rehearsal Policies in Class-Incremental Learning}\label{sec:adv_rehearal_policies}
\noindent Beyond class-balanced uniform sampling and herding \cite{RN113}, the CIL literature has proposed alternative rehearsal policies, including sampling exemplars near the decision boundary and with high prediction entropy \cite{RN163}; selecting samples with the largest prediction uncertainty based on the augmentation of multiple instances \cite{RN165}; maximizing the gradient diversity in the rehearsal set \cite{RN9}; and preserving a weighted coreset \cite{RN12} designed to approximate the full gradient at convergence of training \cite{RN6}. Additionally, samples with the lowest example forgetting frequency \cite{RN8} are retained in \cite{RN22}, and cardinality-constrained bilevel optimization is used in \cite{RN164}. 

As outlined in the main body of this paper, these alternative rehearsal policies, however, provide only marginal improvements over class-balanced uniform sampling and herding, while incurring substantially higher computational costs and algorithmic complexity. For a more comprehensive survey of rehearsal policies proposed in the CIL literature, the reader is referred to \cite{RN183}.

\section{Derivation of the Class-Wise R-SGD Lemma}\label{sec:proof_class_wise_SGD_lemma}
\noindent In what follows, we provide a full derivation of the \emph{Class-Wise R-SGD lemma} (Lemma \ref{lemma:per_iteration_class_wise_GD_inequality}).

For each past class $c \in \mathcal{Y}^{1:m-1}$, under the $L_c^m$-smoothness assumption (Assumption \ref{ass:L_smootness}), the following holds when a single R-SGD update step (\ref{eq:update_rule}) is performed at iterate $\theta_t$ during incremental step $s_m$:
\begin{align}
    \Delta^c_t &\leq -\lambda_t^m\langle\nabla_\theta \mathcal{L}(\theta_t;D_c),\,\bar{g}(\theta_t;k_t^m,\xi_t^{\mathcal{R}^{m-1}}, \xi_t^{D^m}) \rangle \nonumber\\
    &\quad + \frac{(\lambda_t^m)^2 L_c^m}{2}\left|\left|\bar{g}(\theta_t;k_t^m,\xi_t^{\mathcal{R}^{m-1}}, \xi_t^{D^m}) \right|\right|^2_2\,,
\end{align}
where $\Delta^c_t$ is defined as in Lemma \ref{lemma:per_iteration_class_wise_GD_inequality}, i.e., $\Delta^c_t := \mathcal{L}(\theta_{t+1}; D_c) - \mathcal{L}(\theta_{t}; D_c)$.

By expanding the overall stochastic gradient in the first term on the right-hand side as in (\ref{eq:equivalent_overall_stochastic_gradient}), and applying the linearity property of the inner product, we obtain:
\begin{align}
    \Delta^c_t &\leq  -\lambda_t^m \frac{k_{t,c}^m}{K_t^m}\left|\left| \nabla_\theta \mathcal{L}(\theta_t; D_c) \right|\right|_2^2 \nonumber \\
    &\quad -\lambda_t^m \frac{k_{t,c}^m}{K_t^m}\langle B(\theta_t; \xi_t^{\mathcal{R}_c^{m-1}}),\, \nabla_\theta \mathcal{L}(\theta_t; D_c)\rangle \nonumber \\
    &\quad  -\lambda_t^m \sum_{y \in \mathcal{Y}^{1:m-1}_{-c}}\frac{k_{t,y}^m}{K_t^m}\langle B(\theta_t; \xi_t^{\mathcal{R}_y^{m-1}}),\,\nabla_\theta \mathcal{L}(\theta_t; D_c)\rangle \nonumber \\
    &\quad -\lambda_t^m \sum_{y \in \mathcal{Y}^{1:m-1}_{-c}}\frac{k_{t,y}^m}{K_t^m}\langle \nabla_\theta \mathcal{L}(\theta_t; D_y),\,\nabla_\theta \mathcal{L}(\theta_t; D_c)\rangle \nonumber \\
    &\quad -\lambda_t^m (1-\alpha^m)\langle g(\theta_t; \xi_t^{D^m}),\,\nabla_\theta \mathcal{L}(\theta_t; D_c)\rangle \nonumber \\
    &\quad + \frac{(\lambda_t^m)^2 L_c^m}{2}\left|\left|\bar{g}(\theta_t;k_t^m,\xi_t^{\mathcal{R}^{m-1}}, \xi_t^{D^m}) \right|\right|^2_2\,,
\end{align}
where $\mathcal{Y}^{1:m-1}_{-c}$ is defined as in Lemma \ref{lemma:per_iteration_class_wise_GD_inequality}, i.e., $\mathcal{Y}^{1:m-1}_{-c} := \mathcal{Y}^{1:m-1} \setminus \{c\}$.

Next, taking the joint expectation over all random mini-batches drawn at iteration $t$, as well as over the random mini-batch sizes associated with each past class, yields the following:

\[
\resizebox{\linewidth}{!}{$
\begin{aligned}
    \mathbb{E}_{z_t}\left[\Delta^c_t\right] &\leq -\lambda_t^m \frac{\mathbb{E}_{k_{t,c}^m}\left[k_{t,c}^m\right]}{K_t^m}\left|\left| \nabla_\theta \mathcal{L}(\theta_t; D_c) \right|\right|_2^2 \nonumber \\
    &\quad -\lambda_t^m \mathbb{E}_{k_{t,c}^m}\left[\frac{k_{t,c}^m}{K_t^m}\langle \mathbb{E}_{\xi_t^{\mathcal{R}_c^{m-1}}|k_{t,c}^m}\left[B(\theta_t; \xi_t^{\mathcal{R}_c^{m-1}})\right],\, \nabla_\theta \mathcal{L}(\theta_t; D_c)\rangle \right]\nonumber \\
    &\quad  -\lambda_t^m \sum_{y \in \mathcal{Y}^{1:m-1}_{-c}}\mathbb{E}_{k_{t,y}^m}\left[\frac{k_{t,y}^m}{K_t^m}\langle \mathbb{E}_{\xi_t^{\mathcal{R}_y^{m-1}}|k_{t,y}^m}\left[B(\theta_t; \xi_t^{\mathcal{R}_y^{m-1}})\right],\,\nabla_\theta \mathcal{L}(\theta_t; D_c)\rangle\right] \nonumber \\
    &\quad -\lambda_t^m \sum_{y \in \mathcal{Y}^{1:m-1}_{-c}}\frac{\mathbb{E}_{k_{t,y}^m}[k_{t,y}^m]}{K_t^m}\langle \nabla_\theta \mathcal{L}(\theta_t; D_y),\,\nabla_\theta \mathcal{L}(\theta_t; D_c)\rangle \nonumber \\
    &\quad -\lambda_t^m (1-\alpha^m)\langle \mathbb{E}_{\xi_t^{D^m}}\left[g(\theta_t; \xi_t^{D^m})\right],\,\nabla_\theta \mathcal{L}(\theta_t; D_c)\rangle \nonumber \\
    &\quad + \frac{(\lambda_t^m)^2 L_c^m}{2}\mathbb{E}_{z_t}\left[\left|\left|\bar{g}(\theta_t;k_t^m,\xi_t^{\mathcal{R}^{m-1}}, \xi_t^{D^m}) \right|\right|^2_2\right]\,,
\end{aligned}
$}
\tag{\theequation}\stepcounter{equation}
\]
where, as in Lemma \ref{lemma:per_iteration_class_wise_GD_inequality},  $\mathbb{E}_{z_t}[\cdot]$ denotes the joint expectation over $k_t^m$, $\xi_t^{\mathcal{R}^{m-1}}$ and $\xi_t^{D^m}$. Observe that, on the right-hand side of the inequality above, the first, fourth, and fifth terms follow from the marginalisation property of expectation, i.e.:
\begin{equation}
\mathbb{E}_{X,Y}\left[f(X)\right] = \mathbb{E}_{X}\left[f(X)\right]\,
\end{equation}
as well as from the linearity of both expectation and the inner product. In the second and third terms, we applied the marginalisation property of expectation together with the tower rule, i.e.:
\begin{equation}
    \mathbb{E}_{X,Y}\left[f(X,Y)\right] = \mathbb{E}_{X}\left[\mathbb{E}_{Y|X}\left[f(X,Y)\right]\right]\,,
\end{equation}
and again made use of the linearity of expectation and the inner product.

Note that, because of the multinomial distribution in (\ref{eq:multinomial}), for each past class $c$, the expected number of samples $k_{t,c}^m$ drawn at iteration $t$ is:
\begin{equation}\label{eq:expected_size_mini_batch_per_class}
    \mathbb{E}_{k_{t,c}^m}\left[k_{t,c}^m\right] = \alpha^m K_t^m p_c^m\,, \quad \forall c \in \mathcal{Y}^{1:m-1}\,.
\end{equation} 
Second, for any random mini-batch $\xi_t^D \in \xi_t^{\mathcal{R}^{m-1}} \cup \{\xi_t^{D^m}\}$, the expected stochastic noise term at iterate $\theta_t$ vanishes as a direct result of the i.i.d. sampling process described in Section \ref{sec:variant_SGD}:
\begin{equation}\label{eq:expected_stochastic_noise}
    \mathbb{E}_{\xi_t^D}\left[n(\theta_t, \xi_t^D)\right] = \mathbf{0}\,.
\end{equation}
Lastly, at iteration $t$ and for each past class $c$, the expectation of the stochastic noise term conditioned on $k_{t,c}^m$ is identical to the unconditional expectation, since, as described in Section \ref{sec:variant_SGD}, the number of samples drawn from $\mathcal{R}^{m-1}_c$ does not affect the i.i.d. nature of the sampling process:
\begin{align}
\mathbb{E}_{\xi_t^{\mathcal{R}^{m-1}_c}|k_{t,c}^m}\left[n(\theta_t, \xi_t^{\mathcal{R}^{m-1}_c})\right] &= \mathbb{E}_{\xi_t^{\mathcal{R}^{m-1}_c}}\left[n(\theta_t, \xi_t^{\mathcal{R}^{m-1}_c})\right]\, \label{eq:conditional_expectation_stochastic_noise_1}\\
&= \mathbf{0}, \quad \forall c \in \mathcal{Y}^{1:m-1}\,.\label{eq:conditional_expectation_stochastic_noise_2}
\end{align}
with (\ref{eq:conditional_expectation_stochastic_noise_2}) following directly from  (\ref{eq:expected_stochastic_noise}). By leveraging (\ref{eq:expected_size_mini_batch_per_class}), (\ref{eq:expected_stochastic_noise}), (\ref{eq:conditional_expectation_stochastic_noise_1}), (\ref{eq:conditional_expectation_stochastic_noise_2}) along with the definition in (\ref{eq:random_bias}), the following holds:

\begin{align}
    \mathbb{E}_{z_t}\left[\Delta^c_t\right] &\leq -\lambda_t^m \alpha^m p^m_c\left|\left| \nabla_\theta \mathcal{L}(\theta_t; D_c) \right|\right|_2^2 \nonumber \\
    &\quad -\lambda_t^m \alpha^m p^m_c \langle b_c(\theta_t),\, \nabla_\theta \mathcal{L}(\theta_t; D_c) \rangle\nonumber \\
    &\quad  -\lambda_t^m \sum_{y \in \mathcal{Y}^{1:m-1}_{-c}}\alpha^m p^m_y \langle b_y(\theta_t),\,\nabla_\theta \mathcal{L}(\theta_t; D_c)\rangle \nonumber \\
    &\quad -\lambda_t^m \sum_{y \in \mathcal{Y}^{1:m-1}_{-c}}\alpha^m p^m_y \langle \nabla_\theta \mathcal{L}(\theta_t; D_y),\,\nabla_\theta \mathcal{L}(\theta_t; D_c)\rangle \nonumber \\
    &\quad -\lambda_t^m (1-\alpha^m)\langle \nabla_\theta \mathcal{L}(\theta_t; D^m),\,\nabla_\theta \mathcal{L}(\theta_t; D_c)\rangle \nonumber \\
    &\quad + \frac{(\lambda_t^m)^2 L_c^m}{2}\mathbb{E}_{z_t}\left[\left|\left|\bar{g}(\theta_t;k_t^m,\xi_t^{\mathcal{R}^{m-1}}, \xi_t^{D^m}) \right|\right|^2_2\right]\,.
\end{align}
Finally, by incorporating the term $I^m_c(\theta_t)$ (\ref{eq:overall_interference_term}) in the previous inequality, we obtain the \emph{Class-Wise R-SGD lemma} as stated in Lemma \ref{lemma:per_iteration_class_wise_GD_inequality}:
\begin{align}
        \mathbb{E}_{z_t}\left[\Delta_t^c\right] &\leq - \lambda_t^m \alpha^m p_c^m \,|| \nabla_\theta \mathcal{L}(\theta_t;D_c)||_2^2 \nonumber \\
        & \quad +\lambda_t^m \, || \nabla_\theta \mathcal{L}(\theta_t;D_c)||_2 \,I_c^m(\theta_t) \nonumber \\
        & \quad + \frac{(\lambda_t^m)^2L_c^m}{2}\,\mathbb{E}_{z_t}\left[\left|\left|\bar{g}(\theta_t; k_t^m, \xi_t^{\mathcal{R}^{m-1}},\,\xi_t^{D^m} )\right|\right|^2_2\right] \,.
    \end{align}
    
\hfill $\square$

\section{Empirical Comparison between NIC and ALL-NIC}\label{sec:empirical_comp_NIC_ALL_NIC}
\noindent As shown in Section \ref{sec:imbalanced_forgetting_coefficients}, NIC is defined using gradients with respect to the last-layer parameters associated with a specific past class, while SIC and CIC are defined based on gradients with respect to all last-layer parameters. This distinction is motivated by empirical evidence indicating that restricting gradients to class-specific last-layer parameters leads to substantially improved performance for NIC. In the following, we denote as ALL-NIC the counterpart of NIC that is identically formulated but considers all last-layer parameters.

In more detail, using the same evaluation framework as in Section \ref{sec:individual_pred_strength}, we observe that NIC is substantially more effective than ALL-NIC in predicting the ranking of past classes by their degree of forgetting across our benchmarks. This is summarized in Table \ref{tab:spearman_ALL-NIC_and_NIC}. Importantly, NIC outperforms ALL-NIC not only in terms of mean marginal associations but also in terms of mean conditional associations, indicating that NIC also captures more unique predictive information than ALL-NIC relative to the other coefficients (SIC and CIC). Lastly, the predictive signal of NIC is also more consistent than that of ALL-NIC, as evidenced by its lower standard deviations. These results suggest that gradient-level interference from new classes on the last-layer parameters associated with a specific old class during an incremental step plays a more significant role in that class’s forgetting at the end of that step than interference on that class measured over all last-layer parameters. This may be because the last-layer parameters associated with a specific past class are the most directly involved in correctly classifying that class.

Furthermore, as shown in Table \ref{tab:spearman_interference_bias}, the associations between NIC and SIC are stronger and more consistent than those between ALL-NIC and SIC across our benchmarks. Additional details on the relationship between NIC and SIC are provided in Section \ref{sec:ana_relationship_SIC_NIC}.

\begin{table}[htbp]
  \centering
  \caption{Estimated mean $\pm$ standard deviation (with 95\% CIs) of the predictive strength achieved by NIC and ALL-NIC for the second and third incremental steps in $\Omega_{\mathrm{C100}}$ and $\Omega_{\mathrm{TIN}}$. The predictive strength is with respect to ranking past classes by their forgetting and is measured using Spearman’s correlation $\rho$ and partial Spearman’s correlation $\rho_p$.}
  \label{tab:spearman_ALL-NIC_and_NIC}
\resizebox{\linewidth}{!}{%
  \begin{tabular}{llcc}
    \toprule
    & & NIC & ALL-NIC\\
    \midrule
    & \multicolumn{3}{c}{$\Omega_{\mathrm{C100}}$}\\
    \midrule
    \multirow{2}{*}{$\Omega^2_{\mathrm{C100}}$}
      & $\rho$ & 0.74 {\tiny(0.72, 0.75)} $\pm$ 0.15 {\tiny(0.13, 0.17)} & 0.50 {\tiny(0.48, 0.53)} $\pm$ 0.25 {\tiny(0.23, 0.27)}\\
      & $\rho_p$  & 0.30 {\tiny(0.26, 0.33)} $\pm$ 0.28 {\tiny(0.26, 0.31)} & 0.23 {\tiny(0.19, 0.26)} $\pm$ 0.29 {\tiny(0.27, 0.32)}  \\
      \midrule
    \multirow{2}{*}{$\Omega^{2;\, \{10\%, 20\%\}}_{\mathrm{C100}}$}
      & $\rho$ & 0.75 {\tiny(0.73, 0.77)} $\pm$ 0.17 {\tiny(0.15, 0.20)} & 0.53 {\tiny(0.49, 0.56)} $\pm$ 0.26 {\tiny(0.24, 0.30)}\\
      & $\rho_p$  & 0.26 {\tiny(0.22, 0.31)} $\pm$ 0.32 {\tiny(0.30, 0.35)} & 0.20 {\tiny(0.16, 0.25)} $\pm$ 0.32 {\tiny(0.29, 0.35)}  \\
    \midrule
    \multirow{2}{*}{$\Omega^3_{\mathrm{C100}}$}
      & $\rho$ & 0.65 {\tiny(0.64, 0.67)} $\pm$ 0.14 {\tiny(0.12, 0.15)} & 0.37 {\tiny(0.34, 0.40)} $\pm$ 0.23 {\tiny(0.21, 0.25)}\\
      & $\rho_p$  & 0.22 {\tiny(0.19, 0.25)} $\pm$ 0.23 {\tiny(0.22, 0.25)} & 0.12 {\tiny(0.09, 0.16)} $\pm$ 0.25 {\tiny(0.23, 0.27)}  \\
    \midrule
    & \multicolumn{3}{c}{$\Omega_{\mathrm{TIN}}$}\\
    \midrule
    \multirow{2}{*}{$\Omega^2_{\mathrm{TIN}}$}
      & $\rho$ & 0.53 {\tiny(0.51, 0.54)} $\pm$ 0.14 {\tiny(0.13, 0.15)} & 0.26 {\tiny(0.24, 0.28)} $\pm$ 0.22 {\tiny(0.21, 0.24)}\\
      & $\rho_p$  & 0.20 {\tiny(0.17, 0.22)} $\pm$ 0.19 {\tiny(0.18, 0.21)} & -0.02 {\tiny(-0.04, 0.00)} $\pm$ 0.20 {\tiny(0.19, 0.22)}  \\
     \midrule 
     \multirow{2}{*}{$\Omega^{2;\, \{10\%, 20\%\}}_{\mathrm{TIN}}$}
      & $\rho$ & 0.53 {\tiny(0.51, 0.55)} $\pm$ 0.15 {\tiny(0.14, 0.16)} & 0.24 {\tiny(0.21, 0.28)} $\pm$ 0.24 {\tiny(0.22, 0.26)}\\
      & $\rho_p$  & 0.17 {\tiny(0.14, 0.19)} $\pm$ 0.22 {\tiny(0.20, 0.24)} & -0.05 {\tiny(-0.07, -0.01)} $\pm$ 0.22 {\tiny(0.21, 0.25)}  \\
     \midrule 
     \multirow{2}{*}{$\Omega^3_{\mathrm{TIN}}$}
      & $\rho$ & 0.46 {\tiny(0.44, 0.47)} $\pm$ 0.14 {\tiny(0.12, 0.15)} & 0.15 {\tiny(0.12, 0.18)} $\pm$ 0.20 {\tiny(0.19, 0.22)}\\
      & $\rho_p$  & 0.13 {\tiny(0.11, 0.15)} $\pm$ 0.16 {\tiny(0.14, 0.17)} & -0.08 {\tiny(-0.10, -0.06)} $\pm$ 0.14 {\tiny(0.13, 0.16)}  \\
    \bottomrule
  \end{tabular}
  }
\end{table}

\begin{table}[htbp]
  \centering
  \caption{Estimated mean $\pm$ standard deviation (with 95\% CIs) of the association between NIC and SIC, and ALL-NIC and SIC, measured using Spearman's correlation $\rho$, for the second and third incremental steps in $\Omega_{\mathrm{C100}}$ and $\Omega_{\mathrm{TIN}}$.}
  \label{tab:spearman_interference_bias}
  \resizebox{\linewidth}{!}{%
  \begin{tabular}{lcc}
    \toprule
    & NIC & ALL-NIC \\
    \midrule
    & \multicolumn{2}{c}{$\Omega_{\mathrm{C100}}$}\\
    \midrule
    $\Omega^2_{\mathrm{C100}}$ & 0.71 {\tiny(0.69, 0.73)} $\pm$ 0.17 {\tiny(0.16, 0.19)} & 0.45 {\tiny(0.42, 0.47)} $\pm$ 0.23 {\tiny(0.21, 0.25)} \\ 
    \midrule
    $\Omega^{2;\,\{10\%, 20\%\}}_{\mathrm{C100}}$ & 0.74 {\tiny(0.72, 0.76)} $\pm$ 0.19 {\tiny(0.17, 0.22)} & 0.49 {\tiny(0.45, 0.52)} $\pm$ 0.26 {\tiny(0.23, 0.29)}\\
    \midrule
    $\Omega^3_{\mathrm{C100}}$ & 0.62 {\tiny(0.60, 0.64)} $\pm$ 0.15 {\tiny(0.13, 0.17)} & 0.32 {\tiny(0.29, 0.35)} $\pm$ 0.21 {\tiny(0.20, 0.23)}\\ 
    \midrule
    & \multicolumn{2}{c}{$\Omega_{\mathrm{TIN}}$}\\
    \midrule
    $\Omega^2_{\mathrm{TIN}}$ & 0.57 {\tiny(0.56, 0.59)} $\pm$ 0.14 {\tiny(0.13, 0.15)} & 0.49 {\tiny(0.47, 0.51)} $\pm$ 0.16 {\tiny(0.15, 0.18)}\\
    \midrule
    $\Omega^{2;\, \{10\%, 20\%\}}_{\mathrm{TIN}}$ & 0.58 {\tiny(0.56, 0.60)} $\pm$ 0.16 {\tiny(0.14, 0.17)} & 0.46 {\tiny(0.44, 0.49)} $\pm$ 0.18 {\tiny(0.17, 0.20)}\\
    \midrule
    $\Omega^3_{\mathrm{TIN}}$ & 0.46 {\tiny(0.45, 0.48)} $\pm$ 0.14 {\tiny(0.13, 0.15)} & 0.40 {\tiny(0.38, 0.41)} $\pm$ 0.15 {\tiny(0.13, 0.16)}\\
    \bottomrule
  \end{tabular}
  }
\end{table}

\section{Dataset Details}\label{sec:datasets_details}
\noindent As outlined in the main body of this work, the CIL sequences of the experiments in $\Omega_{\mathrm{C100}}$ and $\Omega_{\mathrm{TIN}}$ are derived from the CIFAR-100 \cite{RN100} and Tiny-ImageNet \cite{RN201} datasets, respectively. Here, we provide more details about these datasets.

The CIFAR-100 dataset comprises 100 classes, each with 500 training and 100 test images. Each image is RGB with a resolution of $32 \times 32$ pixels. The Tiny-ImageNet dataset is derived from the ILSVRC2012 ImageNet classification benchmark \cite{RN202}. It comprises 200 classes, each with 500 training and 50 test images. All images are RGB with a resolution of $64 \times 64$ pixels. This dataset is generally considered more challenging than CIFAR-100 due to its greater visual complexity and class diversity.

We use Tiny-ImageNet in one of our full-factorial benchmarks instead of its superset, ILSVRC2012 ImageNet, because it contains fewer training images per class. This reduction enables a more systematic empirical investigation that would be less tractable with ILSVRC2012 ImageNet.

\section{Importance of the Random Seed Set in our Benchmarks}\label{sec:importance_random_seed_benchmarks}
\noindent As detailed in Section \ref{sec:benchmarks}, the full suite of experiments in our two benchmarks is constructed as the Cartesian product of three sets: (i) a set of incremental-step sequences (which differs between the two benchmarks), (ii) a set of rehearsal policies (shared by both benchmarks), and (iii) a set of random seeds (also shared). The inclusion of the latter is essential for a comprehensive empirical evaluation, as differences in random seeds lead to different optimization behaviours even when two experiments share the same sequence of incremental steps and the same rehearsal policy. This variation arises because the seed governs all sources of stochasticity in R-SGD throughout training, including the randomness in each optimization step, the initialization of network weights prior to the first incremental step, and the initialization of newly added output nodes in the final layer at each subsequent step.\footnote{As discussed in Appendix \ref{sec:training_details}, the network architecture remains fixed across incremental steps in our benchmarks; only the number of output nodes in the final linear layer increases to account for newly introduced classes.} Crucially, since all policies in the shared set of rehearsal policies rely on class-balanced uniform sampling, the random seed also determines which samples are selected for rehearsal from each previously learned class.

\section{R-SGD Hyperparameters}\label{sec:training_details}
\noindent For the hyperparameters of R-SGD in both $\Omega_{\mathrm{C100}}$ and $\Omega_{\mathrm{TIN}}$, we largely follow the setting described in \cite{RN183}. Specifically, we use an initial learning rate of $0.1$, a momentum coefficient of $0.9$, and a weight decay of $5 \times 10^{-4}$. A cosine annealing learning rate schedule \cite{RN200}, without warm restarts, is applied to gradually reduce the learning rate after each epoch toward a minimum value of $0.0$. The overall mini-batch size is set to 64 during the first incremental step and increased to 128 in subsequent steps; in the latter case, each mini-batch consists of 64 samples drawn from the new training data and 64 samples from the rehearsal set. The standard cross-entropy loss is used for optimization. Basic data augmentation techniques are applied, including random cropping, horizontal flipping, and color jittering. The number of training epochs is set to 170 for the incremental steps in $\Omega_\mathrm{C100}$ and 100 for those in $\Omega_{\mathrm{TIN}}$.

Lastly, we emphasize that the ResNet32 and ResNet18 network backbones \cite{RN184}---used for the experiments in $\Omega_{\mathrm{C100}}$ and $\Omega_{\mathrm{TIN}}$, respectively---undergo no architectural modifications throughout the incremental learning process. Only their classifiers (the last layer) are expanded with additional output nodes to accommodate newly introduced classes at each subsequent incremental step. As is standard practice in CIL, the weights and biases of these newly added output nodes are initialized from the following uniform distribution:
\begin{equation}
    \mathcal{U}\left(-\frac{1}{\sqrt{\textit{in\_feature}}}, \frac{1}{\sqrt{\textit{in\_feature}}} \right)\,,
\end{equation}
where \textit{in\_features} denotes the dimensionality of the input feature vectors to the final classifier.

\section{Confidence Intervals Computation}\label{sec:confidence_intervals_computation}
\noindent As reported in Section \ref{sec:Stratified_Sampling_of_the_Benchmarks}, we use 95\% CIs alongside all estimates to quantify sampling uncertainty.
For sample means, we use the Student’s t-distribution, relying on the Central Limit Theorem to justify approximate normality of the sampling distribution.
For sample standard deviations, whose sampling distribution is generally non-normal and asymmetric, we employ bootstrap confidence intervals. Specifically, the bias-corrected and accelerated (BCa) method \cite{RN196} is used with 1,000,000 resamples to reduce Monte Carlo variability.

\section{Finer-Grained Analysis of Imbalanced Forgetting}\label{sec:fine_grained_analysis_imb_forgetting}
\noindent In Section \ref{sec:analysis_imbalanced_forgetting}, we analyzed the severity of imbalanced forgetting for the second and third incremental steps in both $\Omega_{\mathrm{C100}}$ and $\Omega_{\mathrm{TIN}}$, aggregating across different percentages of rehearsal retention and newly introduced classes. Here, we provide a finer-grained analysis that explicitly characterizes how this severity varies with these proportions.

Specifically, Table \ref{tab:FG-R_FG-HG_fine_grained} reports the estimated means and standard deviations of both FG-R and FG-HG for each partition in $\Omega_d^k$, with $d \in \{\mathrm{C100}, \mathrm{TIN}\}$ and $k \in \{2,3\}$. The results show that both decreasing rehearsal percentage and increasing proportion of new classes leads to higher mean values of both metrics and, consequently, to greater severity of imbalanced forgetting on average. Moreover, the standard deviations are relatively small compared to the means, indicating that incremental steps within the same partitions exhibit similar levels of imbalanced forgetting.

Furthermore, for each $\Omega_d^k$, we also conduct two type-III two-way ANOVAs---one for FG-R and one for FG-HG---to assess whether rehearsal retention percentage, percentage of newly introduced classes, and their interaction have statistically significant effects on the two metrics in $\Omega_d^k$, and to quantify the corresponding effect sizes. To ensure robustness to potential heteroskedasticity, we perform the ANOVAs using the HC3 heteroskedasticity-consistent covariance estimator \cite{RN198}, which can be viewed as a generalization of Welch’s approach for handling unequal variances. The results of these ANOVAs are reported in Table \ref{tab:anova_FG-R_FG-HSG}.

\begin{figure*}[!t]
  \centering
  \includegraphics[width=0.9\textwidth]{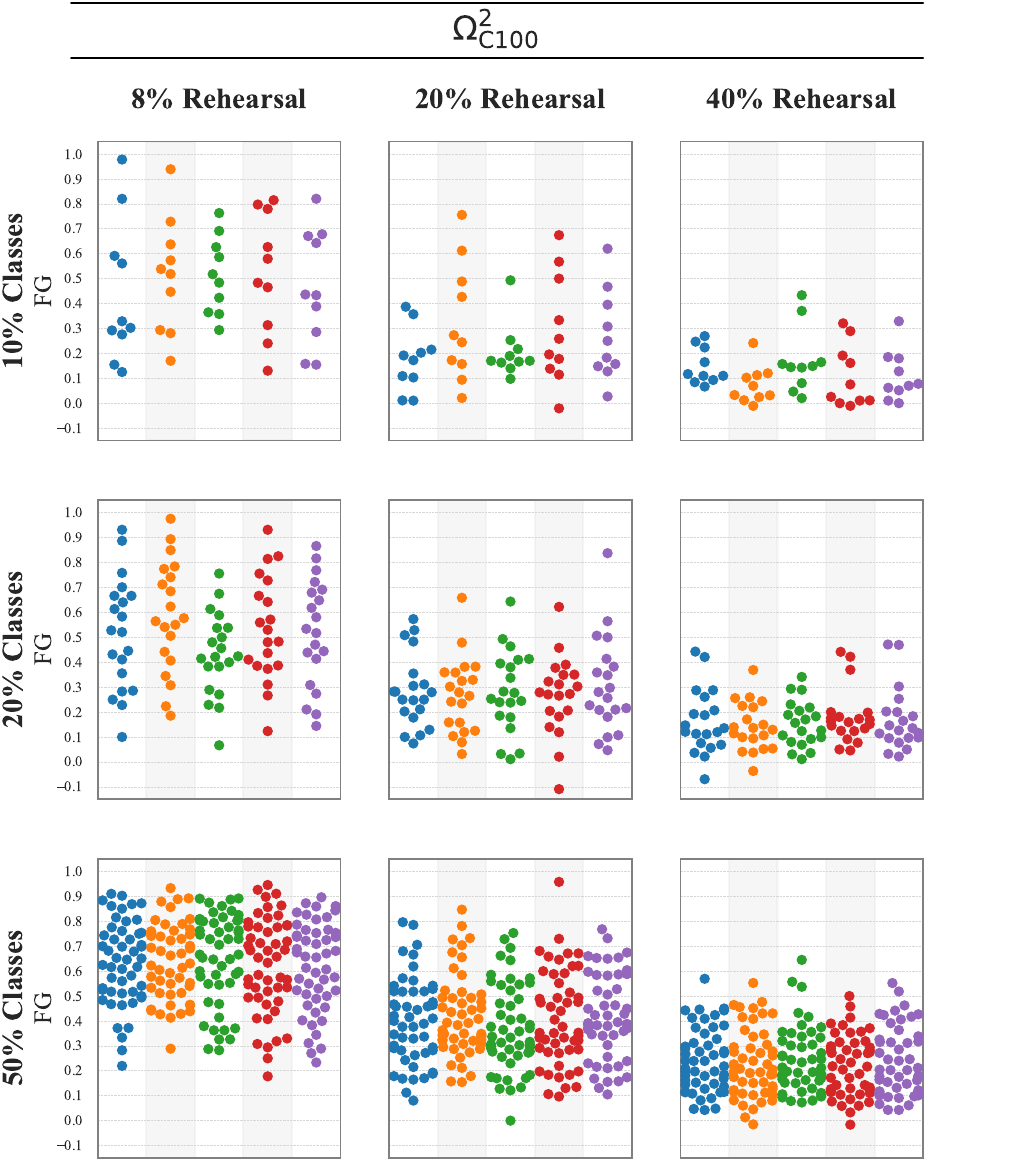}
  \caption{Swarm plots illustrating the distribution of forgetting across past classes, as measured by FG (\ref{eq:forgetting}), for randomly selected incremental steps from each partition in $\Omega_{\mathrm{C100}}^2$. Each cell contains five vertical swarm plots, each corresponding to a different randomly selected incremental step from the respective partition. Each swarm plot includes one point per past class present in the corresponding incremental step.}
  \label{fig:swarm_plots_forgetting_cifar100_1}
\end{figure*}

\begin{figure*}[!t]
  \centering
  \includegraphics[width=0.9\textwidth]{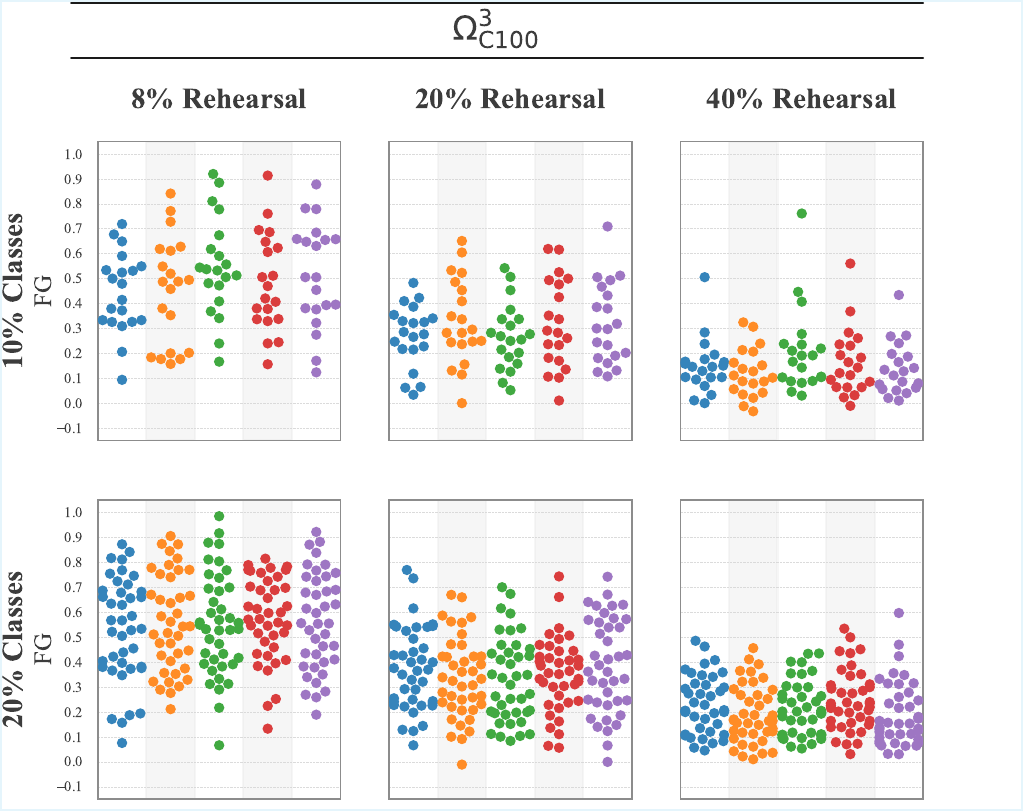}
  \caption{Swarm plots illustrating the distribution of forgetting across past classes, as measured by FG (\ref{eq:forgetting}), for randomly selected incremental steps from each partition in $\Omega_{\mathrm{C100}}^3$. Each cell contains five vertical swarm plots, each corresponding to a different randomly selected incremental step from the respective partition. Each swarm plot includes one point per past class present in the corresponding incremental step.}
  \label{fig:swarm_plots_forgetting_cifar100_2}
\end{figure*}

\begin{figure*}[!t]
  \centering
  \includegraphics[width=0.9\textwidth]{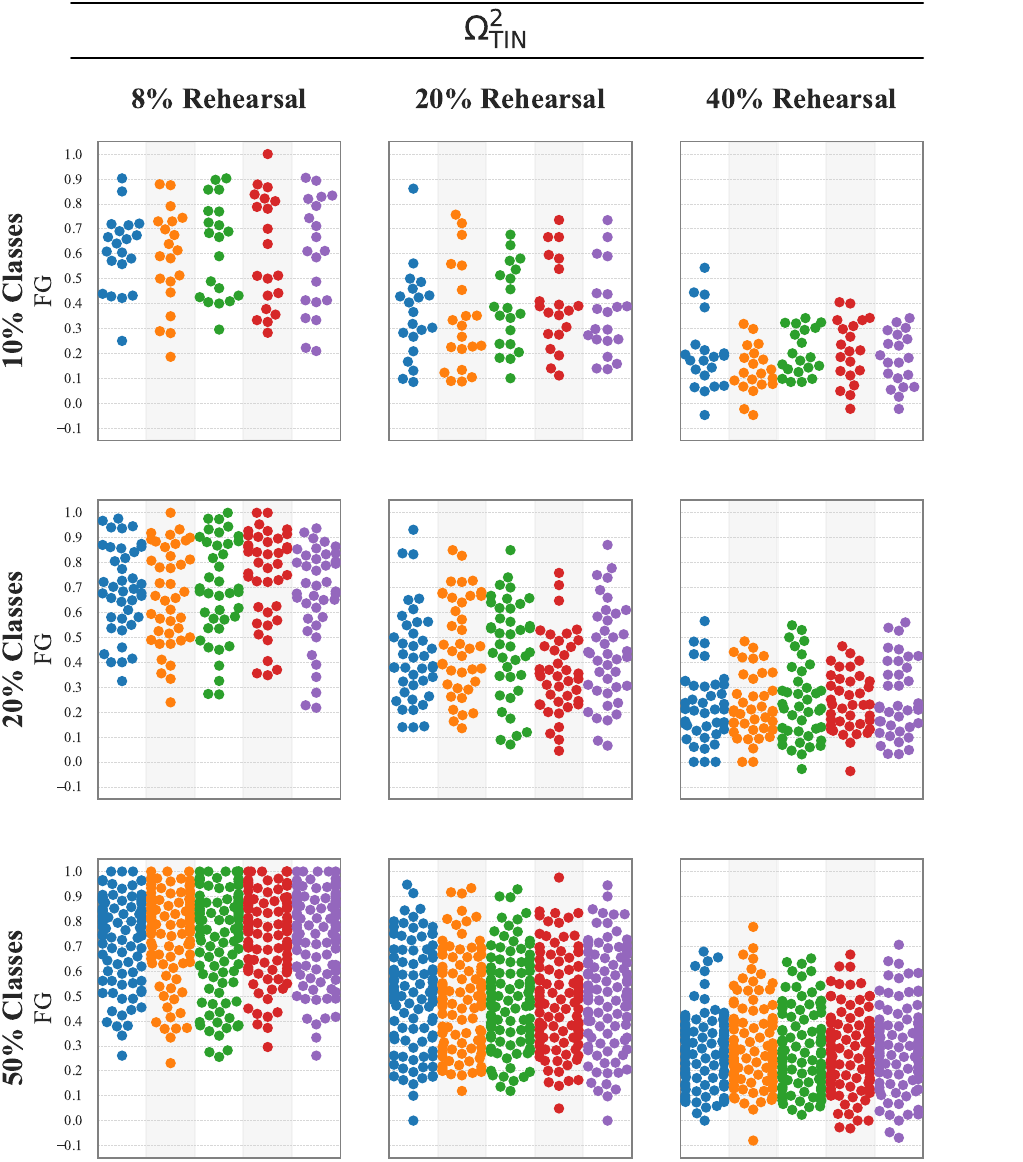}
  \caption{Swarm plots illustrating the distribution of forgetting across past classes, as measured by FG (\ref{eq:forgetting}), for randomly selected incremental steps from each partition in $\Omega_{\mathrm{TIN}}^2$. Each cell contains five vertical swarm plots, each corresponding to a different randomly selected incremental step from the respective partition. Each swarm plot includes one point per past class present in the corresponding incremental step.}
  \label{fig:swarm_plots_forgetting_tinyimagenet_1}
\end{figure*}

\begin{figure*}[!t]
  \centering
  \includegraphics[width=0.9\textwidth]{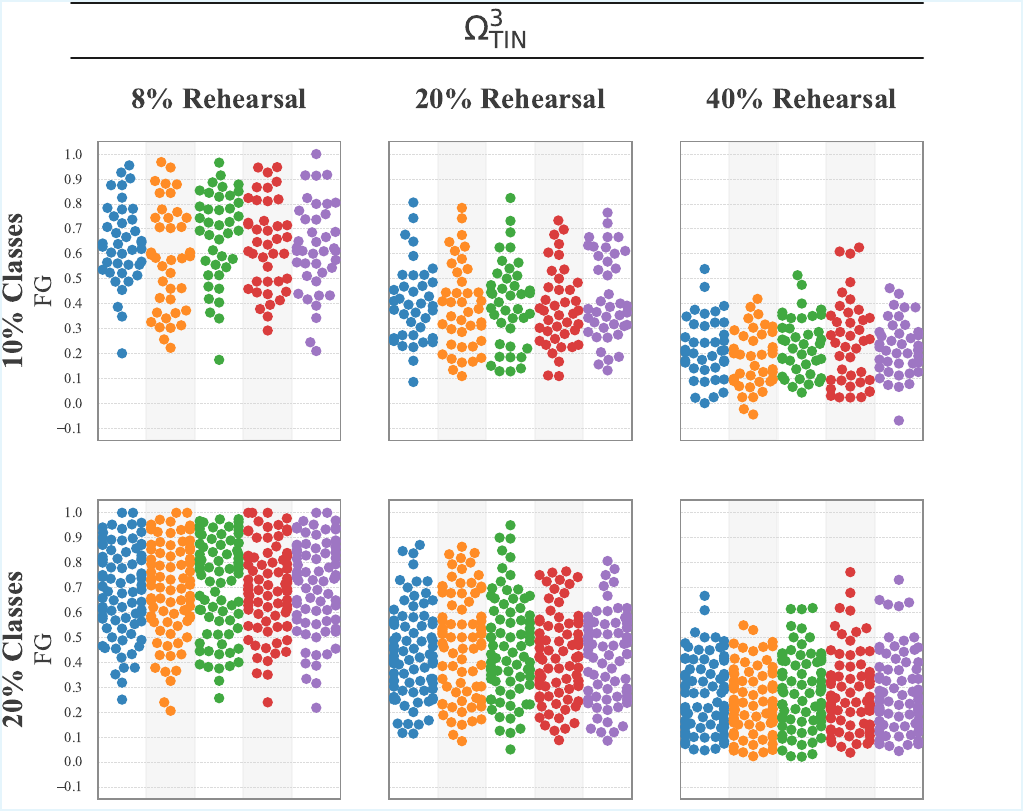}
  \caption{Swarm plots illustrating the distribution of forgetting across past classes, as measured by FG (\ref{eq:forgetting}), for randomly selected incremental steps from each partition in $\Omega_{\mathrm{TIN}}^3$. Each cell contains five vertical swarm plots, each corresponding to a different randomly selected incremental step from the respective partition. Each swarm plot includes one point per past class present in the corresponding incremental step.}
  \label{fig:swarm_plots_forgetting_tinyimagenet_2}
\end{figure*}

For completeness, Figure \ref{fig:swarm_plots_forgetting_cifar100_1}, \ref{fig:swarm_plots_forgetting_cifar100_2}, \ref{fig:swarm_plots_forgetting_tinyimagenet_1}, and \ref{fig:swarm_plots_forgetting_tinyimagenet_2} present a set of plots illustrating the distribution of forgetting across past classes, as measured by FG (\ref{eq:forgetting}), for randomly selected incremental steps in our benchmarks.

\begin{table*}[htbp]
  \centering
  \caption{Estimated mean $\pm$ standard deviation (with 95\% CIs) of FG-R and FG-HG for each partition in our benchmarks. "Pooled" estimates are obtained from the union of partitions sharing the same rehearsal retention percentage or percentage of newly introduced classes.}
  \label{tab:FG-R_FG-HG_fine_grained}
  \resizebox{\textwidth}{!}{%
  \begin{tabular}{llccc@{\hskip 20pt}c}
    \toprule
    & & 10\% Classes & 20\% Classes & 50\% Classes  & \textbf{Pooled}\\
    \midrule
    \multicolumn{6}{c}{\textbf{Benchmark:} $\Omega_{\mathrm{C100}}$}\\
    \midrule
    \multicolumn{6}{c}{$\Omega_{\mathrm{C100}}^2$}\\
    \midrule
    \multirow{2}{*}{8\% Reh.}
        & FG-R & $0.62$ {\tiny $(0.59, 0.66)$} $\pm$ $0.12$ {\tiny $(0.10, 0.14)$} & $0.75$ {\tiny $(0.72, 0.78)$} $\pm$ $0.09$ {\tiny $(0.07, 0.11)$} & $0.73$ {\tiny $(0.70, 0.76)$} $\pm$ $0.08$ {\tiny $(0.06, 0.14)$} & $0.70$ {\tiny $(0.68, 0.72)$} $\pm$ $0.11$ {\tiny $(0.10, 0.13)$}\\
        & FG-HG & $0.31$ {\tiny $(0.29, 0.33)$} $\pm$ $0.05$ {\tiny $(0.05, 0.07)$}& $0.33$ {\tiny $(0.32, 0.34)$} $\pm$ $0.04$ {\tiny $(0.03, 0.05)$} & $0.30$ {\tiny $(0.29, 0.31)$} $\pm$ $0.03$ {\tiny $(0.02, 0.03)$} & $0.31$ {\tiny $(0.31, 0.32)$} $\pm$ $0.04$ {\tiny $(0.04, 0.05)$}\\
    \midrule
    \multirow{2}{*}{20\% Reh.}
        & FG-R & $0.51$ {\tiny $(0.47, 0.55)$} $\pm$ $0.13$ {\tiny $(0.11, 0.15)$}& $0.60$ {\tiny $(0.57, 0.63)$} $\pm$ $0.09$ {\tiny $(0.07, 0.10)$} & $0.75$ {\tiny $(0.72, 0.78)$} $\pm$ $0.09$ {\tiny $(0.07, 0.11)$} & $0.62$ {\tiny $(0.60, 0.65)$} $\pm$ $0.14$ {\tiny $(0.13, 0.16)$}\\
        & FG-HG & $0.25$ {\tiny $(0.23, 0.27)$} $\pm$ $0.06$ {\tiny $(0.05, 0.08)$} & $0.27$ {\tiny $(0.25, 0.28)$} $\pm$ $0.04$ {\tiny $(0.03, 0.04)$} & $0.29$ {\tiny $(0.28, 0.29)$} $\pm$ $0.03$ {\tiny $(0.02, 0.03)$} & $0.27$ {\tiny $(0.26, 0.28)$} $\pm$ $0.05$ {\tiny $(0.04, 0.06)$}\\
    \midrule
    \multirow{2}{*}{40\% Reh.}
        & FG-R & $0.35$ {\tiny $(0.32, 0.38)$} $\pm$ $0.09$ {\tiny $(0.07, 0.11)$}& $0.45$ {\tiny $(0.43, 0.47)$} $\pm$ $0.06$ {\tiny $(0.05, 0.07)$} & $0.58$ {\tiny $(0.55, 0.60)$} $\pm$ $0.08$ {\tiny $(0.07, 0.10)$} & $0.46$ {\tiny $(0.44, 0.48)$} $\pm$ $0.12$ {\tiny $(0.11, 0.14)$}\\
        & FG-HG & $0.17$ {\tiny $(0.16, 0.18)$} $\pm$ $0.04$ {\tiny $(0.03, 0.05)$} & $0.18$ {\tiny $(0.17, 0.19)$} $\pm$ $0.03$ {\tiny $(0.03, 0.04)$} & $0.21$ {\tiny $(0.20, 0.22)$} $\pm$ $0.02$ {\tiny $(0.02, 0.02)$} & $0.19$ {\tiny $(0.18, 0.19)$} $\pm$ $0.04$ {\tiny $(0.03, 0.04)$}\\
    \midrule
     \addlinespace[7pt]
     \multirow{2}{*}{\textbf{Pooled}}
        & FG-R & $0.50$ {\tiny $(0.47, 0.52)$} $\pm$ $0.16$ {\tiny $(0.14, 0.18)$} & $0.60$ {\tiny $(0.57, 0.63)$} $\pm$ $0.15$ {\tiny $(0.14, 0.16)$}  & $0.69$ {\tiny $(0.67, 0.71)$} $\pm$ $0.11$ {\tiny $(0.10, 0.13)$} & $-$\\
        & FG-HG & $0.24$ {\tiny $(0.23, 0.26)$} $\pm$ $0.08$ {\tiny $(0.07, 0.09)$}  & $0.26$ {\tiny $(0.25, 0.27)$} $\pm$ $0.07$ {\tiny $(0.07, 0.08)$} & $0.27$ {\tiny $(0.26, 0.27)$} $\pm$ $0.05$ {\tiny $(0.04, 0.05)$} & $-$\\
    \midrule
    \multicolumn{6}{c}{$\Omega_{\mathrm{C100}}^3$}\\
    \midrule
    \multirow{2}{*}{8\% Reh.}
        & FG-R & $0.72$ {\tiny $(0.69, 0.75)$} $\pm$ $0.09$ {\tiny $(0.08, 0.11)$} & $0.78$ {\tiny $(0.75, 0.81)$} $\pm$ $0.08$ {\tiny $(0.07, 0.11)$} & $-$ & $0.75$ {\tiny $(0.73, 0.77)$} $\pm$ $0.09$ {\tiny $(0.08, 0.11)$}\\
        & FG-HG & $0.33$ {\tiny $(0.31, 0.34)$} $\pm$ $0.05$ {\tiny $(0.04, 0.06)$} & $0.33$ {\tiny $(0.32, 0.34)$} $\pm$ $0.03$ {\tiny $(0.03, 0.04)$} & $-$ & $0.33$ {\tiny $(0.32, 0.34)$} $\pm$ $0.04$ {\tiny $(0.03, 0.05)$}\\
    \midrule
    \multirow{2}{*}{20\% Reh.}
        & FG-R & $0.57$ {\tiny $(0.55, 0.60)$} $\pm$ $0.09$ {\tiny $(0.07, 0.11)$} & $0.68$ {\tiny $(0.64, 0.71)$} $\pm$ $0.10$ {\tiny $(0.09, 0.13)$} & $-$ & $0.62$ {\tiny $(0.60, 0.65)$} $\pm$ $0.11$ {\tiny $(0.09, 0.13)$}\\
        & FG-HG & $0.25$ {\tiny $(0.24, 0.26)$} $\pm$ $0.04$ {\tiny $(0.03, 0.04)$} & $0.27$ {\tiny $(0.26, 0.27)$} $\pm$ $0.03$ {\tiny $(0.03, 0.04)$} & $-$ & $0.26$ {\tiny $(0.25, 0.27)$} $\pm$ $0.03$ {\tiny $(0.03, 0.04)$}\\
    \midrule
    \multirow{2}{*}{40\% Reh.}
        & FG-R & $0.43$ {\tiny $(0.40, 0.46)$} $\pm$ $0.11$ {\tiny $(0.08, 0.14)$} & $0.53$ {\tiny $(0.50, 0.56)$} $\pm$ $0.08$ {\tiny $(0.07, 0.10)$} & $-$ & $0.48$ {\tiny $(0.46, 0.50)$} $\pm$ $0.11$ {\tiny $(0.10, 0.12)$}\\
        & FG-HG & $0.17$ {\tiny $(0.16, 0.18)$} $\pm$ $0.03$ {\tiny $(0.03, 0.04)$} & $0.19$ {\tiny $(0.19, 0.20)$} $\pm$ $0.02$ {\tiny $(0.01, 0.02)$} & $-$ & $0.18$ {\tiny $(0.18, 0.19)$} $\pm$ $0.03$ {\tiny $(0.02, 0.03)$}\\
    \midrule
    \addlinespace[7pt]
    \multirow{2}{*}{\textbf{Pooled}}
        & FG-R & $0.57$ {\tiny $(0.55, 0.60)$} $\pm$ $0.15$ {\tiny $(0.14, 0.17)$}  & $0.66$ {\tiny $(0.64, 0.69)$} $\pm$ $0.14$ {\tiny $(0.12, 0.15)$} & $-$ & $-$\\
        & FG-HG & $0.25$ {\tiny $(0.24, 0.26)$} $\pm$ $0.07$ {\tiny $(0.07, 0.08)$} & $0.26$ {\tiny $(0.25, 0.27)$} $\pm$ $0.06$ {\tiny $(0.06, 0.07)$} & $-$ & $-$\\
    \midrule
    \multicolumn{6}{c}{\textbf{Benchmark:} $\Omega_{\mathrm{TIN}}$}\\
    \midrule
    \multicolumn{6}{c}{ $\Omega_{\mathrm{TIN}}^2$}\\
    \midrule
    \multirow{2}{*}{8\% Reh.}
        & FG-R & $0.62$ {\tiny $(0.58, 0.67)$} $\pm$ $0.08$ {\tiny $(0.06, 0.11)$} & $0.72$ {\tiny $(0.69, 0.76)$} $\pm$ $0.06$ {\tiny $(0.05, 0.10)$} & $0.76$ {\tiny $(0.73, 0.79)$} $\pm$ $0.06$ {\tiny $(0.05, 0.08)$} & $0.70$ {\tiny $(0.67, 0.73)$} $\pm$ $0.09$ {\tiny $(0.08, 0.11)$}\\
        & FG-HG & $0.29$ {\tiny $(0.27, 0.32)$} $\pm$ $0.04$ {\tiny $(0.03, 0.05)$} & $0.31$ {\tiny $(0.29, 0.32)$} $\pm$ $0.03$ {\tiny $(0.02, 0.04)$}& $0.29$ {\tiny $(0.28, 0.29)$} $\pm$ $0.01$ {\tiny $(0.01, 0.02)$} & $0.30$ {\tiny $(0.29, 0.30)$} $\pm$ $0.03$ {\tiny $(0.02, 0.04)$}\\
    \midrule
    \multirow{2}{*}{20\% Reh.}
        & FG-R & $0.58$ {\tiny $(0.53, 0.62)$} $\pm$ $0.08$ {\tiny $(0.06, 0.11)$} & $0.71$ {\tiny $(0.68, 0.74)$} $\pm$ $0.05$ {\tiny $(0.03, 0.08)$}& $0.86$ {\tiny $(0.84, 0.88)$} $\pm$ $0.04$ {\tiny $(0.03, 0.06)$} & $0.71$ {\tiny $(0.67, 0.75)$} $\pm$ $0.13$ {\tiny $(0.11, 0.15)$}\\
        & FG-HG & $0.25$ {\tiny $(0.23, 0.28)$} $\pm$ $0.04$ {\tiny $(0.03, 0.05)$} & $0.30$ {\tiny $(0.29, 0.31)$} $\pm$ $0.02$ {\tiny $(0.02, 0.03)$} & $0.31$ {\tiny $(0.30, 0.32)$} $\pm$ $0.02$ {\tiny $(0.01, 0.02)$} & $0.29$ {\tiny $(0.28, 0.30)$} $\pm$ $0.04$ {\tiny $(0.03, 0.05)$}\\
    \midrule
    \multirow{2}{*}{40\% Reh.}
        & FG-R & $0.45$ {\tiny $(0.39, 0.51)$} $\pm$ $0.10$ {\tiny $(0.06, 0.16)$} & $0.57$ {\tiny $(0.53, 0.62)$} $\pm$ $0.07$ {\tiny $(0.05, 0.11)$} & $0.74$ {\tiny $(0.69, 0.79)$} $\pm$ $0.08$ {\tiny $(0.06, 0.12)$} & $0.59$ {\tiny $(0.54, 0.63)$} $\pm$ $0.15$ {\tiny $(0.13, 0.18)$}\\
        & FG-HG & $0.19$ {\tiny $(0.17, 0.22)$} $\pm$ $0.04$ {\tiny $(0.03, 0.05)$} & $0.22$ {\tiny $(0.21, 0.23)$} $\pm$ $0.02$ {\tiny $(0.01, 0.03)$} & $0.25$ {\tiny $(0.25, 0.26)$} $\pm$ $0.01$ {\tiny $(0.01, 0.02)$} & $0.22$ {\tiny $(0.21, 0.23)$} $\pm$ $0.04$ {\tiny $(0.03, 0.04)$}\\
    \midrule
    \addlinespace[7pt]
    \multirow{2}{*}{\textbf{Pooled}}
        & FG-R & $0.55$ {\tiny $(0.51, 0.58)$} $\pm$ $0.11$ {\tiny $(0.10, 0.14)$} & $0.67$ {\tiny $(0.64, 0.70)$} $\pm$ $0.09$ {\tiny $(0.08, 0.11)$} & $0.79$ {\tiny $(0.76, 0.81)$} $\pm$ $0.08$ {\tiny $(0.07, 0.10)$} & $-$\\
        & FG-HG & $0.25$ {\tiny $(0.23, 0.26)$} $\pm$ $0.06$ {\tiny $(0.05, 0.07)$} & $0.27$ {\tiny $(0.26, 0.29)$} $\pm$ $0.05$ {\tiny $(0.04, 0.05)$} & $0.28$ {\tiny $(0.28, 0.29)$} $\pm$ $0.03$ {\tiny $(0.02, 0.04)$} & $-$\\
    \midrule
    \multicolumn{6}{c}{ $\Omega_{\mathrm{TIN}}^3$}\\
    \midrule
    \multirow{2}{*}{8\% Reh.}
        & FG-R & $0.75$ {\tiny $(0.73, 0.77)$} $\pm$ $0.06$ {\tiny $(0.05, 0.07)$} & $0.75$ {\tiny $(0.73, 0.76)$} $\pm$ $0.05$ {\tiny $(0.05, 0.07)$} & $-$ & $0.75$ {\tiny $(0.74, 0.76)$} $\pm$ $0.06$ {\tiny $(0.05, 0.07)$}\\
        & FG-HG & $0.31$ {\tiny $(0.30, 0.32)$} $\pm$ $0.03$ {\tiny $(0.03, 0.04)$} & $0.29$ {\tiny $(0.29, 0.30)$} $\pm$ $0.02$ {\tiny $(0.02, 0.03)$} & $-$ & $0.30$ {\tiny $(0.30, 0.31)$} $\pm$ $0.03$ {\tiny $(0.02, 0.03)$}\\
    \midrule
    \multirow{2}{*}{20\% Reh.}
        & FG-R & $0.71$ {\tiny $(0.68, 0.73)$} $\pm$ $0.07$ {\tiny $(0.06, 0.09)$} & $0.79$ {\tiny $(0.76, 0.82)$} $\pm$ $0.08$ {\tiny $(0.07, 0.10)$} & $-$ & $0.75$ {\tiny $(0.73, 0.77)$} $\pm$ $0.09$ {\tiny $(0.08, 0.10)$}\\
        & FG-HG & $0.29$ {\tiny $(0.28, 0.30)$} $\pm$ $0.03$ {\tiny $(0.02, 0.03)$} & $0.30$ {\tiny $(0.29, 0.30)$} $\pm$ $0.03$ {\tiny $(0.02, 0.03)$} & $-$ & $0.29$ {\tiny $(0.29, 0.30)$} $\pm$ $0.03$ {\tiny $(0.02, 0.03)$}\\
    \midrule
    \multirow{2}{*}{40\% Reh.}
        & FG-R & $0.55$ {\tiny $(0.53, 0.58)$} $\pm$ $0.08$ {\tiny $(0.06, 0.10)$} & $0.69$ {\tiny $(0.65, 0.72)$} $\pm$ $0.11$ {\tiny $(0.09, 0.16)$} & $-$ & $0.62$ {\tiny $(0.59, 0.65)$} $\pm$ $0.12$ {\tiny $(0.10, 0.15)$}\\
        & FG-HG & $0.21$ {\tiny $(0.20, 0.22)$} $\pm$ $0.03$ {\tiny $(0.02, 0.04)$}  & $0.23$ {\tiny $(0.22, 0.23)$} $\pm$ $0.02$ {\tiny $(0.01, 0.02)$} & $-$ & $0.22$ {\tiny $(0.21, 0.22)$} $\pm$ $0.02$ {\tiny $(0.02, 0.03)$}\\
    \midrule
    \addlinespace[7pt]
    \multirow{2}{*}{\textbf{Pooled}}
        & FG-R & $0.67$ {\tiny $(0.65, 0.69)$} $\pm$ $0.11$ {\tiny $(0.10, 0.12)$} & $0.74$ {\tiny $(0.72, 0.76)$} $\pm$ $0.10$ {\tiny $(0.08, 0.11)$} & $-$ & $-$\\
        & FG-HG & $0.27$ {\tiny $(0.26, 0.28)$} $\pm$ $0.05$ {\tiny $(0.05, 0.06)$} & $0.27$ {\tiny $(0.26, 0.28)$} $\pm$ $0.04$ {\tiny $(0.04, 0.04)$} & $-$ & $-$\\
    \bottomrule
  \end{tabular}
  }
  
\end{table*}

\begin{table}[htbp]
  \centering
  \caption{Type-III two-way ANOVA results for FG-R and FG-HG on our benchmarks, using rehearsal retention percentage and percentage of new classes as factors, together with their interaction term. Reported statistics include the $F$-statistic, $p$-value, and partial eta squared ($\eta^2_p$).}
  \label{tab:anova_FG-R_FG-HSG}
  \resizebox{\linewidth}{!}{%
  \begin{tabular}{llccc}
    \toprule
    & & {$F$-statistic} & $p$-value & {$\eta^2_p$}  \\
    \midrule
    \multicolumn{5}{c}{\textbf{Benchmark:} $\Omega_{\mathrm{C100}}$}\\
    \midrule
    \multicolumn{5}{c}{$\Omega_{\mathrm{C100}}^2$}\\
    \midrule
    \multirow{3}{*}{FG-R}
        & Rehearsal retention percentage & 70.17 & $2.31 \times 10^{-26}$ & 0.29\\
        & Percentage new classes & 15.45 & $3.69 \times 10^{-7}$ & 0.08\\
        & Interaction & 9.70 & $1.89 \times 10^{-7}$ & 0.10\\
    \midrule
    \multirow{3}{*}{FG-HG}
        & Rehearsal retention percentage & 87.62 & $1.36 \times 10^{-31}$ & 0.33\\
        & Percentage new classes & 6.67 & $1.43 \times 10^{-3}$ & 0.04 \\
        & Interaction & 8.76 & $9.46 \times 10^{-7}$ & 0.09\\
    \midrule
    \multicolumn{5}{c}{$\Omega_{\mathrm{C100}}^{2; \,\{10\%, 20\%\}}$}\\
    \midrule
    \multirow{3}{*}{FG-R}
        & Rehearsal retention percentage & 70.17 & $1.34 \times 10^{-24}$ & 0.37\\
        & Percentage new classes & 29.00 & $1.76 \times 10^{-7}$ & 0.11 \\
        & Interaction & 0.89 & $4.10 \times 10^{-1}$ & 0.01 \\
    \midrule
    \multirow{3}{*}{FG-HG}
        & Rehearsal retention percentage & 87.62 & $3.95 \times 10^{-29}$ & 0.43 \\
        & Percentage new classes & 3.82 & $5.18 \times 10^{-2}$ & $0.02$ \\
        & Interaction & 0.34 & $7.15 \times 10^{-1}$ & 0.00 \\
    \midrule
    \multicolumn{5}{c}{$\Omega_{\mathrm{C100}}^3$}\\
    \midrule
    \multirow{3}{*}{FG-R}
        & Rehearsal retention percentage & 84.87 & $1.92 \times 10^{-28}$ & 0.42 \\
        & Percentage new classes & 8.88 & $3.19 \times 10^{-3}$ & 0.04 \\
        & Interaction & 1.31 & $2.71 \times 10^{-1}$ & 0.01 \\
    \midrule
    \multirow{3}{*}{FG-HG}
        & Rehearsal retention percentage & 164.31 & $2.65 \times 10^{-45}$ & 0.58 \\
        & Percentage new classes & 0.00 & $9.83 \times 10^{-1}$ & 0.00 \\
        & Interaction & 1.79 & $1.70 \times 10^{-1}$ & 0.02 \\
    \midrule
    \multicolumn{5}{c}{\textbf{Benchmark:} $\Omega_{\mathrm{TIN}}$}\\
    \midrule
    \multicolumn{5}{c}{$\Omega_{\mathrm{TIN}}^2$}\\
    \midrule
    \multirow{3}{*}{FG-R}
        & Rehearsal retention percentage & 22.56 & $6.07 \times 10^{-10}$ & 0.11 \\
        & Percentage new classes & 7.90 & $4.42 \times 10^{-4}$ & 0.04 \\
        & Interaction & 17.82 & $2.49 \times 10^{-13}$ & 0.17 \\
    \midrule
    \multirow{3}{*}{FG-HG}
        & Rehearsal retention percentage & 65.79 & $5.42 \times 10^{-25}$ & 0.27 \\
        & Percentage new classes & 15.65 & $3.09 \times 10^{-7}$ & 0.08 \\
        & Interaction & 21.29 & $9.73 \times 10^{-16}$ & 0.20 \\
    \midrule
    \multicolumn{5}{c}{$\Omega_{\mathrm{TIN}}^{2; \,\{10\%, 20\%\}}$}\\
    \midrule
    \multirow{3}{*}{FG-R}
        & Rehearsal retention percentage & 22.56 & $1.10 \times 10^{-9}$ & 0.16 \\
        & Percentage new classes & 7.51 & $6.59 \times 10^{-3}$ & 0.03 \\
        & Interaction & 5.48 & $4.73 \times 10^{-3}$ & 0.04 \\
    \midrule
    \multirow{3}{*}{FG-HG}
        & Rehearsal retention percentage & 65.79 & $2.13 \times 10^{-23}$ & 0.36 \\
        & Percentage new classes & 1.32 & $2.52 \times 10^{-1}$ & 0.01 \\
        & Interaction & 4.79 & $9.17 \times 10^{-3}$ & 0.04 \\
    \midrule
    \multicolumn{5}{c}{$\Omega_{\mathrm{TIN}}^3$}\\
    \midrule
    \multirow{3}{*}{FG-R}
        & Rehearsal retention percentage & 81.14 & $1.70 \times 10^{-27}$ & 0.41 \\
        & Percentage new classes & 0.13 & $7.21 \times 10^{-1}$ & 0.00\\
        & Interaction & 18.21 & $4.46 \times 10^{-8}$ & 0.13\\
    \midrule
    \multirow{3}{*}{FG-HG}
        & Rehearsal retention percentage & 139.31 & $1.42 \times 10^{-40}$ & 0.54 \\
        & Percentage new classes & 8.75 & $3.41 \times 10^{-3}$ & 0.04 \\
        & Interaction & 10.11 & $6.17 \times 10^{-5}$ & 0.08 \\
    \bottomrule
\end{tabular}
  }
\end{table}

\section{Finer-Grained Analysis of the Individual Predictive Strength of the Last-Layer Imbalanced Forgetting Coefficients}\label{sec:fine_grained_analysis_indi_predi_strength}

\noindent In Section \ref{sec:individual_pred_strength}, we analyzed how effectively each \emph{Last-layer Imbalanced Forgetting Coefficient} predicts the ranking of past classes by their forgetting across our benchmarks. This was accomplished by computing the marginal and conditional associations between each coefficient and class-wise forgetting for the second and third incremental steps in both $\Omega_{\mathrm{C100}}$ and $\Omega_{\mathrm{TIN}}$, aggregating across different percentages of rehearsal retention and newly introduced classes. Here, we provide a finer-grained analysis that explicitly characterizes how these associations vary with these proportions. Specifically, Table \ref{tab:spearman_SIC}, \ref{tab:spearman_CIC}, and \ref{tab:spearman_NIC} report the estimated means and standard deviations of these associations for each partition in $\Omega_d^k$, with $d \in \{\mathrm{C100},\mathrm{TIN}\}$ and $k \in \{2,3\}$. 

\paragraph{SIC} Increasing the proportion of new classes leads to lower mean values for both marginal and conditional associations in all benchmarks. With respect to rehearsal retention, lower levels result in higher mean marginal and conditional associations in $\Omega_{\mathrm{C100}}$. In contrast, both associations increase from $8\%$ to $20\%$ retention, followed by a decrease from $20\%$ to $40\%$ in $\Omega_{\mathrm{TIN}}$.

\paragraph{CIC} Reducing rehearsal retention percentage leads to higher mean marginal and conditional associations in all benchmarks. In contrast, no consistent trend is observed with respect to variations in the proportion of new classes.

\paragraph{NIC} Increasing rehearsal retention percentage generally leads to higher mean conditional associations in all benchmarks, while the mean marginal ones increase from $8\%$ to $20\%$ retention, followed by a decrease from $20\%$ to $40\%$. Regarding the percentage of new classes, higher values result in higher mean conditional associations in all benchmarks, whereas no consistent trend is observed with respect to the marginal ones.

Furthermore, for each coefficient and each $\Omega_d^k$, we also conduct two type-III two-way ANOVAs---one for the marginal associations and one for the difference between the marginal and conditional associations---to assess whether rehearsal retention percentage, percentage of newly introduced classes, and their interaction have statistically significant effects on the two quantities, and to quantify the corresponding effect sizes. To ensure robustness to potential heteroskedasticity, we perform the ANOVAs using the HC3 heteroskedasticity-consistent covariance estimator. The results of these ANOVAs are reported in Table \ref{tab:anova_SIC}, \ref{tab:anova_CIC}, and \ref{tab:anova_NIC}. Overall, rehearsal retention percentage emerges as the dominant factor across SIC, CIC, and NIC, both in terms of effect size and frequency of statistical significance.

\begin{table*}[htbp]
  \centering
   \caption{Estimated mean $\pm$ standard deviation (with 95\% CIs) of the predictive strength achieved by SIC for each partition in our benchmarks. The predictive strength is with respect to ranking past classes by their forgetting and is measured using Spearman’s correlation $\rho$ and partial Spearman’s correlation $\rho_p$. "Pooled" estimates are obtained from the union of partitions sharing the same rehearsal retention percentage or percentage of newly introduced classes.}
  \label{tab:spearman_SIC}
  \resizebox{\textwidth}{!}{%
  \begin{tabular}{llccc@{\hskip 20pt}c}
    \toprule
    \multicolumn{6}{c}{\textbf{\emph{Last-Layer Imbalanced Forgetting Coefficient}:} SIC}\\
    \midrule
    & & 10\% Classes & 20\% Classes & 50\% Classes  & \textbf{Pooled}\\
    \midrule
    \multicolumn{6}{c}{\textbf{Benchmark:} $\Omega_{\mathrm{C100}}$}\\
    \midrule
    \multicolumn{6}{c}{$\Omega_{\mathrm{C100}}^2$}\\
    \midrule
    \multirow{2}{*}{8\% Reh.}
        & $\rho$ & $0.92$ {\tiny $(0.90, 0.93)$} $\pm$ $0.07$ {\tiny $(0.05, 0.09)$} & $0.92$ {\tiny $(0.90, 0.93)$} $\pm$ $0.05$ {\tiny $(0.04, 0.07)$} & $0.88$ {\tiny $(0.87, 0.89)$} $\pm$ $0.04$ {\tiny $(0.04, 0.06)$} & $0.91$ {\tiny $(0.90, 0.91)$} $\pm$ $0.06$ {\tiny $(0.05, 0.07)$}\\
        & $\rho_p$ & $0.86$ {\tiny $(0.81, 0.89)$} $\pm$ $0.15$ {\tiny $(0.11, 0.22)$} & $0.84$ {\tiny $(0.82, 0.87)$} $\pm$ $0.08$ {\tiny $(0.07, 0.10)$} & $0.80$ {\tiny $(0.78, 0.81)$} $\pm$ $0.06$ {\tiny $(0.05, 0.07)$} & $0.83$ {\tiny $(0.81, 0.85)$} $\pm$ $0.11$ {\tiny $(0.09, 0.14)$}\\
    \midrule
    \multirow{2}{*}{20\% Reh.}
        & $\rho$ & $0.91$ {\tiny $(0.88, 0.94)$} $\pm$ $0.16$ {\tiny $(0.10, 0.27)$} & $0.89$ {\tiny $(0.87, 0.90)$} $\pm$ $0.06$ {\tiny $(0.05, 0.07)$} & $0.83$ {\tiny $(0.82, 0.85)$} $\pm$ $0.06$ {\tiny $(0.05, 0.07)$} & $0.88$ {\tiny $(0.87, 0.90)$} $\pm$ $0.11$ {\tiny $(0.08, 0.17)$}\\
        & $\rho_p$ & $0.80$ {\tiny $(0.71, 0.86)$} $\pm$ $0.31$ {\tiny $(0.23, 0.45)$} & $0.74$ {\tiny $(0.69, 0.79)$} $\pm$ $0.15$ {\tiny $(0.13, 0.18)$} & $0.68$ {\tiny $(0.66, 0.71)$} $\pm$ $0.07$ {\tiny $(0.06, 0.09)$} & $0.75$ {\tiny $(0.71, 0.78)$} $\pm$ $0.20$ {\tiny $(0.16, 0.29)$}\\
    \midrule
    \multirow{2}{*}{40\% Reh.}
        & $\rho$ & $0.82$ {\tiny $(0.77, 0.85)$} $\pm$ $0.14$ {\tiny $(0.10, 0.17)$} & $0.78$ {\tiny $(0.74, 0.82)$} $\pm$ $0.13$ {\tiny $(0.11, 0.16)$} & $0.70$ {\tiny $(0.67, 0.73)$} $\pm$ $0.08$ {\tiny $(0.07, 0.10)$} & $0.77$ {\tiny $(0.75, 0.79)$} $\pm$ $0.12$ {\tiny $(0.11, 0.14)$}\\
        & $\rho_p$ & $0.56$ {\tiny $(0.47, 0.64)$} $\pm$ $0.28$ {\tiny $(0.23, 0.35)$} & $0.52$ {\tiny $(0.45, 0.59)$} $\pm$ $0.23$ {\tiny $(0.18, 0.30)$} & $0.47$ {\tiny $(0.42, 0.51)$} $\pm$ $0.14$ {\tiny $(0.11, 0.17)$} & $0.52$ {\tiny $(0.48, 0.56)$} $\pm$ $0.22$ {\tiny $(0.20, 0.26)$}\\
    \midrule
     \addlinespace[7pt]
     \multirow{2}{*}{\textbf{Pooled}}
        & $\rho$ & $0.89$ {\tiny $(0.87, 0.91)$} $\pm$ $0.14$ {\tiny $(0.11, 0.18)$} & $0.87$ {\tiny $(0.86, 0.89)$} $\pm$ $0.11$ {\tiny $(0.09, 0.14)$} & $0.82$ {\tiny $(0.80, 0.83)$} $\pm$ $0.10$ {\tiny $(0.09, 0.11)$} & $-$\\
        & $\rho_p$ & $0.76$ {\tiny $(0.71, 0.80)$} $\pm$ $0.29$ {\tiny $(0.24, 0.35)$} & $0.73$ {\tiny $(0.69, 0.76)$} $\pm$ $0.21$ {\tiny $(0.18, 0.26)$} & $0.67$ {\tiny $(0.64, 0.70)$} $\pm$ $0.17$ {\tiny $(0.15, 0.19)$} & $-$\\
    \midrule
    \multicolumn{6}{c}{$\Omega_{\mathrm{C100}}^3$}\\
    \midrule
    \multirow{2}{*}{8\% Reh.}
        & $\rho$ & $0.92$ {\tiny $(0.91, 0.93)$} $\pm$ $0.04$ {\tiny $(0.03, 0.05)$} & $0.92$ {\tiny $(0.91, 0.92)$} $\pm$ $0.03$ {\tiny $(0.02, 0.04)$} & $-$ & $0.92$ {\tiny $(0.91, 0.92)$} $\pm$ $0.04$ {\tiny $(0.03, 0.04)$}\\
        & $\rho_p$ & $0.87$ {\tiny $(0.84, 0.89)$} $\pm$ $0.08$ {\tiny $(0.07, 0.12)$} & $0.86$ {\tiny $(0.85, 0.87)$} $\pm$ $0.04$ {\tiny $(0.04, 0.05)$} & $-$ & $0.87$ {\tiny $(0.85, 0.88)$} $\pm$ $0.07$ {\tiny $(0.05, 0.09)$}\\
    \midrule
    \multirow{2}{*}{20\% Reh.}
        & $\rho$ & $0.90$ {\tiny $(0.87, 0.91)$} $\pm$ $0.08$ {\tiny $(0.05, 0.14)$} & $0.87$ {\tiny $(0.86, 0.89)$} $\pm$ $0.05$ {\tiny $(0.04, 0.07)$} & $-$ & $0.88$ {\tiny $(0.87, 0.90)$} $\pm$ $0.07$ {\tiny $(0.05, 0.11)$}\\
        & $\rho_p$ & $0.81$ {\tiny $(0.77, 0.85)$} $\pm$ $0.13$ {\tiny $(0.10, 0.17)$} & $0.78$ {\tiny $(0.75, 0.81)$} $\pm$ $0.09$ {\tiny $(0.07, 0.11)$} & $-$ & $0.80$ {\tiny $(0.77, 0.82)$} $\pm$ $0.11$ {\tiny $(0.09, 0.14)$}\\
    \midrule
    \multirow{2}{*}{40\% Reh.}
        & $\rho$ & $0.80$ {\tiny $(0.77, 0.83)$} $\pm$ $0.09$ {\tiny $(0.07, 0.13)$} & $0.76$ {\tiny $(0.74, 0.78)$} $\pm$ $0.07$ {\tiny $(0.06, 0.08)$} & $-$ & $0.78$ {\tiny $(0.76, 0.80)$} $\pm$ $0.08$ {\tiny $(0.07, 0.10)$}\\
        & $\rho_p$ & $0.63$ {\tiny $(0.56, 0.70)$} $\pm$ $0.23$ {\tiny $(0.18, 0.29)$} & $0.58$ {\tiny $(0.54, 0.62)$} $\pm$ $0.12$ {\tiny $(0.10, 0.15)$} & $-$ & $0.61$ {\tiny $(0.56, 0.65)$} $\pm$ $0.18$ {\tiny $(0.15, 0.23)$}\\
    \midrule
    \addlinespace[7pt]
    \multirow{2}{*}{\textbf{Pooled}}
        & $\rho$ & $0.88$ {\tiny $(0.87, 0.89)$} $\pm$ $0.09$ {\tiny $(0.08, 0.12)$} & $0.86$ {\tiny $(0.85, 0.87)$} $\pm$ $0.09$ {\tiny $(0.07, 0.10)$} & $-$ & $-$\\
        & $\rho_p$ & $0.79$ {\tiny $(0.76, 0.82)$} $\pm$ $0.19$ {\tiny $(0.16, 0.24)$} & $0.76$ {\tiny $(0.74, 0.79)$} $\pm$ $0.15$ {\tiny $(0.13, 0.17)$} & $-$ & $-$\\
    \midrule
    \multicolumn{6}{c}{\textbf{Benchmark:} $\Omega_{\mathrm{TIN}}$}\\
    \midrule
    \multicolumn{6}{c}{ $\Omega_{\mathrm{TIN}}^2$}\\
    \midrule
    \multirow{2}{*}{8\% Reh.}
        & $\rho$ & $0.67$ {\tiny $(0.64, 0.71)$} $\pm$ $0.11$ {\tiny $(0.09, 0.13)$} & $0.56$ {\tiny $(0.51, 0.60)$} $\pm$ $0.13$ {\tiny $(0.11, 0.16)$} & $0.42$ {\tiny $(0.39, 0.44)$} $\pm$ $0.09$ {\tiny $(0.07, 0.12)$} & $0.56$ {\tiny $(0.53, 0.59)$} $\pm$ $0.15$ {\tiny $(0.13, 0.17)$}\\
        & $\rho_p$ & $0.59$ {\tiny $(0.54, 0.65)$} $\pm$ $0.17$ {\tiny $(0.14, 0.22)$} & $0.54$ {\tiny $(0.49, 0.57)$} $\pm$ $0.12$ {\tiny $(0.10, 0.15)$} & $0.39$ {\tiny $(0.35, 0.43)$} $\pm$ $0.12$ {\tiny $(0.09, 0.15)$} & $0.51$ {\tiny $(0.48, 0.54)$} $\pm$ $0.16$ {\tiny $(0.14, 0.18)$}\\
    \midrule
    \multirow{2}{*}{20\% Reh.}
        & $\rho$ & $0.78$ {\tiny $(0.75, 0.82)$} $\pm$ $0.12$ {\tiny $(0.09, 0.14)$} & $0.71$ {\tiny $(0.68, 0.74)$} $\pm$ $0.10$ {\tiny $(0.08, 0.14)$} & $0.59$ {\tiny $(0.57, 0.61)$} $\pm$ $0.07$ {\tiny $(0.06, 0.10)$} & $0.70$ {\tiny $(0.68, 0.73)$} $\pm$ $0.12$ {\tiny $(0.11, 0.14)$}\\
        & $\rho_p$ & $0.69$ {\tiny $(0.64, 0.74)$} $\pm$ $0.16$ {\tiny $(0.12, 0.23)$} & $0.61$ {\tiny $(0.57, 0.64)$} $\pm$ $0.11$ {\tiny $(0.09, 0.13)$} & $0.49$ {\tiny $(0.47, 0.51)$} $\pm$ $0.07$ {\tiny $(0.06, 0.09)$} & $0.60$ {\tiny $(0.57, 0.63)$} $\pm$ $0.14$ {\tiny $(0.12, 0.16)$}\\
    \midrule
    \multirow{2}{*}{40\% Reh.}
        & $\rho$ & $0.70$ {\tiny $(0.64, 0.75)$} $\pm$ $0.17$ {\tiny $(0.14, 0.20)$} & $0.61$ {\tiny $(0.58, 0.65)$} $\pm$ $0.12$ {\tiny $(0.09, 0.16)$} & $0.52$ {\tiny $(0.49, 0.54)$} $\pm$ $0.07$ {\tiny $(0.06, 0.08)$} & $0.62$ {\tiny $(0.59, 0.64)$} $\pm$ $0.14$ {\tiny $(0.12, 0.16)$}\\
        & $\rho_p$ & $0.60$ {\tiny $(0.54, 0.65)$} $\pm$ $0.19$ {\tiny $(0.15, 0.25)$} & $0.51$ {\tiny $(0.48, 0.54)$} $\pm$ $0.10$ {\tiny $(0.08, 0.12)$} & $0.34$ {\tiny $(0.31, 0.36)$} $\pm$ $0.07$ {\tiny $(0.06, 0.09)$} & $0.49$ {\tiny $(0.46, 0.52)$} $\pm$ $0.16$ {\tiny $(0.15, 0.18)$}\\
    \midrule
    \addlinespace[7pt]
    \multirow{2}{*}{\textbf{Pooled}}
        & $\rho$ & $0.72$ {\tiny $(0.70, 0.75)$} $\pm$ $0.14$ {\tiny $(0.13, 0.16)$}  & $0.63$ {\tiny $(0.61, 0.65)$} $\pm$ $0.13$ {\tiny $(0.12, 0.15)$} & $0.51$ {\tiny $(0.49, 0.53)$} $\pm$ $0.11$ {\tiny $(0.10, 0.12)$} & $-$\\
        & $\rho_p$ & $0.63$ {\tiny $(0.60, 0.66)$} $\pm$ $0.18$ {\tiny $(0.16, 0.21)$} & $0.55$ {\tiny $(0.53, 0.57)$} $\pm$ $0.12$ {\tiny $(0.10, 0.13)$} & $0.41$ {\tiny $(0.39, 0.43)$} $\pm$ $0.11$ {\tiny $(0.10, 0.13)$} & $-$\\
    \midrule
    \multicolumn{6}{c}{ $\Omega_{\mathrm{TIN}}^3$}\\
    \midrule
    \multirow{2}{*}{8\% Reh.}
        & $\rho$ & $0.56$ {\tiny $(0.53, 0.60)$} $\pm$ $0.11$ {\tiny $(0.09, 0.12)$} & $0.38$ {\tiny $(0.35, 0.41)$} $\pm$ $0.10$ {\tiny $(0.08, 0.13)$} & $-$ & $0.48$ {\tiny $(0.44, 0.51)$} $\pm$ $0.14$ {\tiny $(0.12, 0.16)$}\\
        & $\rho_p$ & $0.56$ {\tiny $(0.52, 0.60)$} $\pm$ $0.13$ {\tiny $(0.10, 0.17)$} & $0.45$ {\tiny $(0.42, 0.48)$} $\pm$ $0.09$ {\tiny $(0.08, 0.11)$} & $-$ & $0.51$ {\tiny $(0.48, 0.54)$} $\pm$ $0.12$ {\tiny $(0.10, 0.15)$}\\
    \midrule
    \multirow{2}{*}{20\% Reh.}
        & $\rho$ & $0.72$ {\tiny $(0.68, 0.75)$} $\pm$ $0.10$ {\tiny $(0.08, 0.13)$} & $0.55$ {\tiny $(0.52, 0.57)$} $\pm$ $0.08$ {\tiny $(0.07, 0.10)$} & $-$ & $0.64$ {\tiny $(0.61, 0.67)$} $\pm$ $0.12$ {\tiny $(0.11, 0.14)$}\\
        & $\rho_p$ & $0.65$ {\tiny $(0.62, 0.69)$} $\pm$ $0.11$ {\tiny $(0.09, 0.14)$} & $0.51$ {\tiny $(0.48, 0.54)$} $\pm$ $0.08$ {\tiny $(0.06, 0.10)$} & $-$ & $0.59$ {\tiny $(0.56, 0.61)$} $\pm$ $0.12$ {\tiny $(0.10, 0.14)$}\\
    \midrule
    \multirow{2}{*}{40\% Reh.}
        & $\rho$ & $0.60$ {\tiny $(0.56, 0.64)$} $\pm$ $0.12$ {\tiny $(0.10, 0.14)$} & $0.48$ {\tiny $(0.46, 0.50)$} $\pm$ $0.07$ {\tiny $(0.06, 0.09)$} & $-$ & $0.55$ {\tiny $(0.52, 0.57)$} $\pm$ $0.11$ {\tiny $(0.10, 0.13)$}\\
        & $\rho_p$ & $0.52$ {\tiny $(0.47, 0.57)$} $\pm$ $0.16$ {\tiny $(0.13, 0.23)$} & $0.36$ {\tiny $(0.33, 0.39)$} $\pm$ $0.09$ {\tiny $(0.08, 0.11)$} & $-$ & $0.45$ {\tiny $(0.41, 0.48)$} $\pm$ $0.15$ {\tiny $(0.13, 0.18)$}\\
    \midrule
    \addlinespace[7pt]
    \multirow{2}{*}{\textbf{Pooled}}
        & $\rho$ & $0.63$ {\tiny $(0.61, 0.65)$} $\pm$ $0.13$ {\tiny $(0.11, 0.14)$} & $0.47$ {\tiny $(0.45, 0.49)$} $\pm$ $0.11$ {\tiny $(0.10, 0.13)$} & $-$ & $-$\\
        & $\rho_p$ & $0.58$ {\tiny $(0.56, 0.61)$} $\pm$ $0.15$ {\tiny $(0.12, 0.18)$} & $0.45$ {\tiny $(0.43, 0.46)$} $\pm$ $0.11$ {\tiny $(0.10, 0.12)$} & $-$ & $-$\\
    \bottomrule
  \end{tabular}
  }

\end{table*}

\begin{table*}[htbp]
  \centering
  \caption{Estimated mean $\pm$ standard deviation (with 95\% CIs) of the predictive strength achieved by CIC for each partition in our benchmarks. The predictive strength is with respect to ranking past classes by their forgetting and is measured using Spearman’s correlation $\rho$ and partial Spearman’s correlation $\rho_p$. "Pooled" estimates are obtained from the union of partitions sharing the same rehearsal retention percentage or percentage of newly introduced classes.}
  \label{tab:spearman_CIC}
  \resizebox{\textwidth}{!}{%
  \begin{tabular}{llccc@{\hskip 20pt}c}
    \toprule
    \multicolumn{6}{c}{\textbf{\emph{Last-Layer Imbalanced Forgetting Coefficient}:} CIC}\\
    \midrule
    & & 10\% Classes & 20\% Classes & 50\% Classes  & \textbf{Pooled}\\
    \midrule
    \multicolumn{6}{c}{\textbf{Benchmark:} $\Omega_{\mathrm{C100}}$}\\
    \midrule
    \multicolumn{6}{c}{$\Omega_{\mathrm{C100}}^2$}\\
    \midrule
    \multirow{2}{*}{8\% Reh.}
        & $\rho$ & $0.27$ {\tiny $(0.12, 0.42)$} $\pm$ $0.38$ {\tiny $(0.33, 0.45)$} & $0.35$ {\tiny $(0.27, 0.43)$} $\pm$ $0.24$ {\tiny $(0.20, 0.29)$} & $0.35$ {\tiny $(0.31, 0.40)$} $\pm$ $0.13$ {\tiny $(0.11, 0.16)$} & $0.33$ {\tiny $(0.27, 0.38)$} $\pm$ $0.28$ {\tiny $(0.24, 0.32)$}\\
        & $\rho_p$ & $0.18$ {\tiny $(0.00, 0.34)$} $\pm$ $0.44$ {\tiny $(0.36, 0.53)$} & $0.20$ {\tiny $(0.11, 0.29)$} $\pm$ $0.26$ {\tiny $(0.21, 0.33)$} & $0.20$ {\tiny $(0.15, 0.25)$} $\pm$ $0.15$ {\tiny $(0.12, 0.17)$} & $0.19$ {\tiny $(0.12, 0.26)$} $\pm$ $0.30$ {\tiny $(0.26, 0.36)$}\\
    \midrule
    \multirow{2}{*}{20\% Reh.}
        & $\rho$ & $0.10$ {\tiny $(-0.04, 0.23)$} $\pm$ $0.37$ {\tiny $(0.32, 0.43)$} & $0.19$ {\tiny $(0.11, 0.27)$} $\pm$ $0.24$ {\tiny $(0.20, 0.29)$} & $0.23$ {\tiny $(0.17, 0.28)$} $\pm$ $0.16$ {\tiny $(0.13, 0.19)$} & $0.17$ {\tiny $(0.12, 0.22)$} $\pm$ $0.28$ {\tiny $(0.24, 0.32)$}\\
        & $\rho_p$ & $0.04$ {\tiny $(-0.10, 0.19)$} $\pm$ $0.39$ {\tiny $(0.33, 0.46)$} & $0.14$ {\tiny $(0.07, 0.22)$} $\pm$ $0.22$ {\tiny $(0.18, 0.28)$} & $0.11$ {\tiny $(0.06, 0.16)$} $\pm$ $0.15$ {\tiny $(0.13, 0.18)$} & $0.10$ {\tiny $(0.04, 0.15)$} $\pm$ $0.27$ {\tiny $(0.24, 0.32)$}\\
    \midrule
    \multirow{2}{*}{40\% Reh.}
        & $\rho$ & $0.09$ {\tiny $(-0.05, 0.23)$} $\pm$ $0.37$ {\tiny $(0.31, 0.46)$} & $0.11$ {\tiny $(0.03, 0.20)$} $\pm$ $0.25$ {\tiny $(0.21, 0.32)$} & $0.07$ {\tiny $(0.02, 0.12)$} $\pm$ $0.15$ {\tiny $(0.12, 0.19)$}  & $0.09$ {\tiny $(0.04, 0.15)$} $\pm$ $0.27$ {\tiny $(0.24, 0.32)$}\\
        & $\rho_p$ & $0.09$ {\tiny $(-0.06, 0.24)$} $\pm$ $0.40$ {\tiny $(0.35, 0.47)$} & $0.17$ {\tiny $(0.08, 0.25)$} $\pm$ $0.24$ {\tiny $(0.21, 0.28)$} & $0.06$ {\tiny $(0.02, 0.10)$} $\pm$ $0.13$ {\tiny $(0.11, 0.17)$} & $0.11$ {\tiny $(0.05, 0.16)$} $\pm$ $0.28$ {\tiny $(0.25, 0.32)$}\\
    \midrule
     \addlinespace[7pt]
     \multirow{2}{*}{\textbf{Pooled}}
        & $\rho$ & $0.16$ {\tiny $(0.07, 0.24)$} $\pm$ $0.38$ {\tiny $(0.35, 0.42)$} & $0.22$ {\tiny $(0.17, 0.27)$} $\pm$ $0.26$ {\tiny $(0.24, 0.30)$} & $0.22$ {\tiny $(0.19, 0.25)$} $\pm$ $0.18$ {\tiny $(0.17, 0.21)$} & $-$\\
        & $\rho_p$ & $0.10$ {\tiny $(0.02, 0.19)$} $\pm$ $0.41$ {\tiny $(0.37, 0.45)$} & $0.17$ {\tiny $(0.12, 0.22)$} $\pm$ $0.24$ {\tiny $(0.22, 0.27)$} & $0.12$ {\tiny $(0.10, 0.15)$} $\pm$ $0.15$ {\tiny $(0.14, 0.17)$} & $-$\\
    \midrule
    \multicolumn{6}{c}{$\Omega_{\mathrm{C100}}^3$}\\
    \midrule
    \multirow{2}{*}{8\% Reh.}
        & $\rho$ & $0.32$ {\tiny $(0.22, 0.42)$} $\pm$ $0.31$ {\tiny $(0.25, 0.38)$} & $0.43$ {\tiny $(0.39, 0.48)$} $\pm$ $0.14$ {\tiny $(0.11, 0.19)$} & $-$ & $0.38$ {\tiny $(0.32, 0.43)$} $\pm$ $0.25$ {\tiny $(0.20, 0.31)$}\\
        & $\rho_p$ & $0.14$ {\tiny $(0.06, 0.23)$} $\pm$ $0.24$ {\tiny $(0.20, 0.30)$} & $0.24$ {\tiny $(0.19, 0.28)$} $\pm$ $0.14$ {\tiny $(0.12, 0.17)$} & $-$ & $0.19$ {\tiny $(0.14, 0.24)$} $\pm$ $0.20$ {\tiny $(0.17, 0.25)$}\\
    \midrule
    \multirow{2}{*}{20\% Reh.}
        & $\rho$ & $0.20$ {\tiny $(0.12, 0.28)$} $\pm$ $0.23$ {\tiny $(0.20, 0.28)$} & $0.21$ {\tiny $(0.16, 0.27)$} $\pm$ $0.17$ {\tiny $(0.15, 0.20)$} & $-$ & $0.21$ {\tiny $(0.16, 0.25)$} $\pm$ $0.20$ {\tiny $(0.18, 0.24)$}\\
        & $\rho_p$ & $0.05$ {\tiny $(-0.04, 0.15)$} $\pm$ $0.27$ {\tiny $(0.23, 0.34)$} & $0.11$ {\tiny $(0.06, 0.16)$} $\pm$ $0.15$ {\tiny $(0.12, 0.18)$} & $-$ & $0.08$ {\tiny $(0.03, 0.13)$} $\pm$ $0.22$ {\tiny $(0.19, 0.27)$}\\
    \midrule
    \multirow{2}{*}{40\% Reh.}
        & $\rho$ & $-0.07$ {\tiny $(-0.17, 0.03)$} $\pm$ $0.28$ {\tiny $(0.24, 0.34)$} & $0.02$ {\tiny $(-0.04, 0.08)$} $\pm$ $0.17$ {\tiny $(0.15, 0.22)$} & $-$ & $-0.02$ {\tiny $(-0.08, 0.03)$} $\pm$ $0.24$ {\tiny $(0.21, 0.28)$}\\
        & $\rho_p$ & $0.04$ {\tiny $(-0.07, 0.16)$} $\pm$ $0.32$ {\tiny $(0.27, 0.39)$} & $0.03$ {\tiny $(-0.03, 0.08)$} $\pm$ $0.17$ {\tiny $(0.13, 0.21)$} & $-$ & $0.03$ {\tiny $(-0.03, 0.10)$} $\pm$ $0.25$ {\tiny $(0.22, 0.30)$}\\
    \midrule
    \addlinespace[7pt]
    \multirow{2}{*}{\textbf{Pooled}}
        & $\rho$ & $0.16$ {\tiny $(0.09, 0.22)$} $\pm$ $0.31$ {\tiny $(0.29, 0.35)$} & $0.23$ {\tiny $(0.19, 0.27)$} $\pm$ $0.23$ {\tiny $(0.21, 0.26)$} & $-$ & $-$\\
        & $\rho_p$ & $0.08$ {\tiny $(0.02, 0.14)$} $\pm$ $0.28$ {\tiny $(0.25, 0.32)$} & $0.13$ {\tiny $(0.09, 0.16)$} $\pm$ $0.17$ {\tiny $(0.15, 0.20)$} & $-$ & $-$\\
    \midrule
    \multicolumn{6}{c}{\textbf{Benchmark:} $\Omega_{\mathrm{TIN}}$}\\
    \midrule
    \multicolumn{6}{c}{ $\Omega_{\mathrm{TIN}}^2$}\\
    \midrule
    \multirow{2}{*}{8\% Reh.}
        & $\rho$ & $0.44$ {\tiny $(0.36, 0.51)$} $\pm$ $0.23$ {\tiny $(0.19, 0.28)$} & $0.47$ {\tiny $(0.42, 0.52)$} $\pm$ $0.15$ {\tiny $(0.12, 0.19)$} & $0.48$ {\tiny $(0.46, 0.50)$} $\pm$ $0.07$ {\tiny $(0.05, 0.08)$} & $0.46$ {\tiny $(0.43, 0.49)$} $\pm$ $0.17$ {\tiny $(0.14, 0.20)$}\\
        & $\rho_p$ & $0.55$ {\tiny $(0.48, 0.62)$} $\pm$ $0.20$ {\tiny $(0.16, 0.24)$} & $0.58$ {\tiny $(0.54, 0.62)$} $\pm$ $0.14$ {\tiny $(0.11, 0.19)$} & $0.57$ {\tiny $(0.54, 0.60)$} $\pm$ $0.08$ {\tiny $(0.07, 0.11)$} & $0.57$ {\tiny $(0.54, 0.60)$} $\pm$ $0.15$ {\tiny $(0.13, 0.17)$}\\
    \midrule
    \multirow{2}{*}{20\% Reh.}
        & $\rho$ & $0.41$ {\tiny $(0.32, 0.49)$} $\pm$ $0.24$ {\tiny $(0.20, 0.31)$} & $0.39$ {\tiny $(0.34, 0.44)$} $\pm$ $0.15$ {\tiny $(0.13, 0.18)$} & $0.34$ {\tiny $(0.31, 0.36)$} $\pm$ $0.08$ {\tiny $(0.07, 0.11)$} & $0.38$ {\tiny $(0.35, 0.41)$} $\pm$ $0.17$ {\tiny $(0.15, 0.21)$}\\
        & $\rho_p$ & $0.35$ {\tiny $(0.26, 0.43)$} $\pm$ $0.24$ {\tiny $(0.20, 0.30)$} & $0.46$ {\tiny $(0.40, 0.51)$} $\pm$ $0.17$ {\tiny $(0.13, 0.25)$} & $0.42$ {\tiny $(0.38, 0.45)$} $\pm$ $0.10$ {\tiny $(0.08, 0.12)$} & $0.41$ {\tiny $(0.37, 0.44)$} $\pm$ $0.19$ {\tiny $(0.16, 0.22)$}\\
    \midrule
    \multirow{2}{*}{40\% Reh.}
        & $\rho$ & $0.29$ {\tiny $(0.22, 0.37)$} $\pm$ $0.22$ {\tiny $(0.18, 0.28)$} & $0.20$ {\tiny $(0.15, 0.25)$} $\pm$ $0.16$ {\tiny $(0.13, 0.21)$} & $0.17$ {\tiny $(0.14, 0.19)$} $\pm$ $0.09$ {\tiny $(0.08, 0.11)$} & $0.22$ {\tiny $(0.19, 0.25)$} $\pm$ $0.17$ {\tiny $(0.15, 0.20)$}\\
        & $\rho_p$ & $0.25$ {\tiny $(0.17, 0.33)$} $\pm$ $0.24$ {\tiny $(0.20, 0.30)$} & $0.26$ {\tiny $(0.21, 0.30)$} $\pm$ $0.14$ {\tiny $(0.12, 0.17)$} & $0.18$ {\tiny $(0.16, 0.21)$} $\pm$ $0.07$ {\tiny $(0.06, 0.09)$} & $0.23$ {\tiny $(0.20, 0.26)$} $\pm$ $0.17$ {\tiny $(0.15, 0.20)$}\\
    \midrule
    \addlinespace[7pt]
    \multirow{2}{*}{\textbf{Pooled}}
        & $\rho$ & $0.38$ {\tiny $(0.34, 0.43)$} $\pm$ $0.24$ {\tiny $(0.21, 0.27)$} & $0.36$ {\tiny $(0.32, 0.39)$} $\pm$ $0.19$ {\tiny $(0.17, 0.22)$} & $0.33$ {\tiny $(0.30, 0.36)$} $\pm$ $0.15$ {\tiny $(0.14, 0.17)$} & $-$\\
        & $\rho_p$ & $0.39$ {\tiny $(0.34, 0.44)$} $\pm$ $0.25$ {\tiny $(0.23, 0.29)$} & $0.44$ {\tiny $(0.40, 0.48)$} $\pm$ $0.20$ {\tiny $(0.18, 0.23)$} & $0.40$ {\tiny $(0.37, 0.44)$} $\pm$ $0.18$ {\tiny $(0.17, 0.20)$} & $-$\\
    \midrule
    \multicolumn{6}{c}{ $\Omega_{\mathrm{TIN}}^3$}\\
    \midrule
    \multirow{2}{*}{8\% Reh.}
        & $\rho$ & $0.56$ {\tiny $(0.51, 0.60)$} $\pm$ $0.13$ {\tiny $(0.11, 0.17)$} & $0.59$ {\tiny $(0.56, 0.61)$} $\pm$ $0.06$ {\tiny $(0.06, 0.08)$} & $-$ & $0.57$ {\tiny $(0.55, 0.59)$} $\pm$ $0.11$ {\tiny $(0.09, 0.13)$}\\
        & $\rho_p$ & $0.60$ {\tiny $(0.56, 0.64)$} $\pm$ $0.12$ {\tiny $(0.10, 0.15)$} & $0.64$ {\tiny $(0.62, 0.66)$} $\pm$ $0.05$ {\tiny $(0.04, 0.06)$} & $-$ & $0.62$ {\tiny $(0.60, 0.64)$} $\pm$ $0.10$ {\tiny $(0.08, 0.12)$}\\
    \midrule
    \multirow{2}{*}{20\% Reh.}
        & $\rho$ & $0.44$ {\tiny $(0.39, 0.48)$} $\pm$ $0.13$ {\tiny $(0.11, 0.17)$} & $0.41$ {\tiny $(0.38, 0.44)$} $\pm$ $0.10$ {\tiny $(0.08, 0.12)$} & $-$ & $0.42$ {\tiny $(0.40, 0.45)$} $\pm$ $0.12$ {\tiny $(0.10, 0.14)$}\\
        & $\rho_p$ & $0.40$ {\tiny $(0.35, 0.44)$} $\pm$ $0.14$ {\tiny $(0.12, 0.18)$} & $0.43$ {\tiny $(0.40, 0.46)$} $\pm$ $0.10$ {\tiny $(0.08, 0.12)$} & $-$ & $0.41$ {\tiny $(0.39, 0.44)$} $\pm$ $0.12$ {\tiny $(0.11, 0.14)$}\\
    \midrule
    \multirow{2}{*}{40\% Reh.}
        & $\rho$ & $0.16$ {\tiny $(0.11, 0.21)$} $\pm$ $0.16$ {\tiny $(0.13, 0.20)$} & $0.14$ {\tiny $(0.10, 0.18)$} $\pm$ $0.12$ {\tiny $(0.10, 0.16)$} & $-$ & $0.15$ {\tiny $(0.12, 0.18)$} $\pm$ $0.14$ {\tiny $(0.12, 0.17)$}\\
        & $\rho_p$ & $0.20$ {\tiny $(0.14, 0.25)$} $\pm$ $0.17$ {\tiny $(0.14, 0.20)$} & $0.12$ {\tiny $(0.08, 0.16)$} $\pm$ $0.12$ {\tiny $(0.10, 0.15)$} & $-$ & $0.16$ {\tiny $(0.12, 0.19)$} $\pm$ $0.15$ {\tiny $(0.13, 0.17)$}\\
    \midrule
    \addlinespace[7pt]
    \multirow{2}{*}{\textbf{Pooled}}
        & $\rho$ & $0.40$ {\tiny $(0.36, 0.44)$} $\pm$ $0.22$ {\tiny $(0.19, 0.24)$} & $0.39$ {\tiny $(0.35, 0.43)$} $\pm$ $0.21$ {\tiny $(0.19, 0.23)$} & $-$ & $-$\\
        & $\rho_p$ & $0.41$ {\tiny $(0.37, 0.45)$} $\pm$ $0.22$ {\tiny $(0.20, 0.24)$} & $0.42$ {\tiny $(0.38, 0.46)$} $\pm$ $0.23$ {\tiny $(0.21, 0.25)$} & $-$ & $-$\\
    \bottomrule
  \end{tabular}
  }
\end{table*}

\begin{table*}[htbp]
  \centering
  \caption{Estimated mean $\pm$ standard deviation (with 95\% CIs) of the predictive strength achieved by NIC for each partition in our benchmarks. The predictive strength is with respect to ranking past classes by their forgetting and is measured using Spearman’s correlation $\rho$ and partial Spearman’s correlation $\rho_p$. "Pooled" estimates are obtained from the union of partitions sharing the same rehearsal retention percentage or percentage of newly introduced classes.}
  \label{tab:spearman_NIC}
  \resizebox{\textwidth}{!}{%
  \begin{tabular}{llccc@{\hskip 20pt}c}
    \toprule
    \multicolumn{6}{c}{\textbf{\emph{Last-Layer Imbalanced Forgetting Coefficient}:} NIC}\\
    \midrule
    & & 10\% Classes & 20\% Classes & 50\% Classes  & \textbf{Pooled}\\
    \midrule
    \multicolumn{6}{c}{\textbf{Benchmark:} $\Omega_{\mathrm{C100}}$}\\
    \midrule
    \multicolumn{6}{c}{$\Omega_{\mathrm{C100}}^2$}\\
    \midrule
    \multirow{2}{*}{8\% Reh.}
        & $\rho$ & $0.67$ {\tiny $(0.58, 0.74)$} $\pm$ $0.24$ {\tiny $(0.20, 0.31)$} & $0.70$ {\tiny $(0.66, 0.74)$} $\pm$ $0.13$ {\tiny $(0.11, 0.17)$} & $0.66$ {\tiny $(0.62, 0.69)$} $\pm$ $0.11$ {\tiny $(0.09, 0.14)$} & $0.68$ {\tiny $(0.65, 0.71)$} $\pm$ $0.17$ {\tiny $(0.15, 0.22)$}\\
        & $\rho_p$ & $0.12$ {\tiny $(-0.02, 0.26)$} $\pm$ $0.37$ {\tiny $(0.31, 0.46)$} & $0.20$ {\tiny $(0.12, 0.28)$} $\pm$ $0.24$ {\tiny $(0.20, 0.29)$} & $0.24$ {\tiny $(0.19, 0.28)$} $\pm$ $0.14$ {\tiny $(0.10, 0.21)$} & $0.19$ {\tiny $(0.13, 0.24)$} $\pm$ $0.28$ {\tiny $(0.24, 0.32)$}\\
    \midrule
    \multirow{2}{*}{20\% Reh.}
        & $\rho$ & $0.80$ {\tiny $(0.74, 0.85)$} $\pm$ $0.16$ {\tiny $(0.14, 0.19)$} & $0.77$ {\tiny $(0.74, 0.80)$} $\pm$ $0.10$ {\tiny $(0.08, 0.13)$} & $0.74$ {\tiny $(0.72, 0.76)$} $\pm$ $0.07$ {\tiny $(0.06, 0.09)$} & $0.77$ {\tiny $(0.75, 0.79)$} $\pm$ $0.12$ {\tiny $(0.10, 0.14)$}\\
        & $\rho_p$ & $0.34$ {\tiny $(0.18, 0.48)$} $\pm$ $0.40$ {\tiny $(0.35, 0.47)$} & $0.27$ {\tiny $(0.19, 0.35)$} $\pm$ $0.24$ {\tiny $(0.19, 0.31)$} & $0.38$ {\tiny $(0.33, 0.42)$} $\pm$ $0.14$ {\tiny $(0.11, 0.17)$} & $0.33$ {\tiny $(0.27, 0.39)$} $\pm$ $0.29$ {\tiny $(0.25, 0.33)$}\\
    \midrule
    \multirow{2}{*}{40\% Reh.}
        & $\rho$ & $0.79$ {\tiny $(0.73, 0.83)$} $\pm$ $0.14$ {\tiny $(0.12, 0.18)$} & $0.76$ {\tiny $(0.72, 0.79)$} $\pm$ $0.13$ {\tiny $(0.10, 0.17)$} & $0.72$ {\tiny $(0.70, 0.74)$} $\pm$ $0.07$ {\tiny $(0.06, 0.09)$} & $0.76$ {\tiny $(0.73, 0.78)$} $\pm$ $0.12$ {\tiny $(0.10, 0.14)$}\\
        & $\rho_p$ & $0.30$ {\tiny $(0.16, 0.43)$} $\pm$ $0.36$ {\tiny $(0.30, 0.43)$} & $0.35$ {\tiny $(0.28, 0.41)$} $\pm$ $0.20$ {\tiny $(0.17, 0.25)$} & $0.46$ {\tiny $(0.41, 0.50)$} $\pm$ $0.14$ {\tiny $(0.12, 0.18)$} & $0.37$ {\tiny $(0.32, 0.42)$} $\pm$ $0.26$ {\tiny $(0.23, 0.31)$}\\
    \midrule
     \addlinespace[7pt]
     \multirow{2}{*}{\textbf{Pooled}}
        & $\rho$ & $0.76$ {\tiny $(0.72, 0.79)$} $\pm$ $0.20$ {\tiny $(0.17, 0.24)$} & $0.74$ {\tiny $(0.72, 0.76)$} $\pm$ $0.13$ {\tiny $(0.11, 0.15)$} & $0.71$ {\tiny $(0.69, 0.72)$} $\pm$ $0.09$ {\tiny $(0.08, 0.11)$} & $-$\\
        & $\rho_p$ & $0.26$ {\tiny $(0.17, 0.34)$} $\pm$ $0.39$ {\tiny $(0.35, 0.43)$} & $0.27$ {\tiny $(0.23, 0.32)$} $\pm$ $0.24$ {\tiny $(0.21, 0.27)$} & $0.36$ {\tiny $(0.33, 0.39)$} $\pm$ $0.17$ {\tiny $(0.15, 0.20)$} & $-$\\
    \midrule
    \multicolumn{6}{c}{$\Omega_{\mathrm{C100}}^3$}\\
    \midrule
    \multirow{2}{*}{8\% Reh.}
        & $\rho$ & $0.63$ {\tiny $(0.58, 0.68)$} $\pm$ $0.16$ {\tiny $(0.12, 0.22)$} & $0.64$ {\tiny $(0.61, 0.68)$} $\pm$ $0.11$ {\tiny $(0.09, 0.13)$} & $-$ & $0.64$ {\tiny $(0.61, 0.67)$} $\pm$ $0.14$ {\tiny $(0.12, 0.18)$}\\
        & $\rho_p$ & $0.16$ {\tiny $(0.07, 0.24)$} $\pm$ $0.25$ {\tiny $(0.21, 0.31)$} & $0.11$ {\tiny $(0.05, 0.17)$} $\pm$ $0.18$ {\tiny $(0.15, 0.22)$} & $-$ & $0.14$ {\tiny $(0.08, 0.19)$} $\pm$ $0.22$ {\tiny $(0.19, 0.26)$}\\
    \midrule
    \multirow{2}{*}{20\% Reh.}
        & $\rho$ & $0.69$ {\tiny $(0.65, 0.74)$} $\pm$ $0.13$ {\tiny $(0.11, 0.15)$} & $0.68$ {\tiny $(0.64, 0.71)$} $\pm$ $0.11$ {\tiny $(0.10, 0.13)$} & $-$ & $0.69$ {\tiny $(0.66, 0.71)$} $\pm$ $0.12$ {\tiny $(0.11, 0.14)$}\\
        & $\rho_p$ & $0.23$ {\tiny $(0.14, 0.31)$} $\pm$ $0.25$ {\tiny $(0.21, 0.30)$} & $0.25$ {\tiny $(0.18, 0.31)$} $\pm$ $0.19$ {\tiny $(0.16, 0.23)$} & $-$ & $0.24$ {\tiny $(0.19, 0.29)$} $\pm$ $0.22$ {\tiny $(0.20, 0.26)$}\\
    \midrule
    \multirow{2}{*}{40\% Reh.}
        & $\rho$ & $0.62$ {\tiny $(0.57, 0.67)$} $\pm$ $0.17$ {\tiny $(0.14, 0.20)$} & $0.65$ {\tiny $(0.61, 0.68)$} $\pm$ $0.11$ {\tiny $(0.09, 0.13)$} & $-$ & $0.64$ {\tiny $(0.60, 0.67)$} $\pm$ $0.14$ {\tiny $(0.12, 0.17)$}\\
        & $\rho_p$ & $0.26$ {\tiny $(0.16, 0.35)$} $\pm$ $0.27$ {\tiny $(0.22, 0.32)$} & $0.31$ {\tiny $(0.24, 0.37)$} $\pm$ $0.19$ {\tiny $(0.16, 0.25)$} & $-$ & $0.28$ {\tiny $(0.23, 0.34)$} $\pm$ $0.23$ {\tiny $(0.20, 0.27)$}\\
    \midrule
    \addlinespace[7pt]
    \multirow{2}{*}{\textbf{Pooled}}
        & $\rho$ & $0.65$ {\tiny $(0.62, 0.68)$} $\pm$ $0.16$ {\tiny $(0.14, 0.18)$} & $0.66$ {\tiny $(0.64, 0.68)$} $\pm$ $0.11$ {\tiny $(0.10, 0.12)$} & $-$ & $-$\\
        & $\rho_p$ & $0.22$ {\tiny $(0.17, 0.26)$} $\pm$ $0.26$ {\tiny $(0.23, 0.29)$} & $0.22$ {\tiny $(0.19, 0.26)$} $\pm$ $0.20$ {\tiny $(0.18, 0.23)$} & $-$ & $-$\\
    \midrule
    \multicolumn{6}{c}{\textbf{Benchmark:} $\Omega_{\mathrm{TIN}}$}\\
    \midrule
    \multicolumn{6}{c}{ $\Omega_{\mathrm{TIN}}^2$}\\
    \midrule
    \multirow{2}{*}{8\% Reh.}
        & $\rho$ & $0.51$ {\tiny $(0.46, 0.56)$} $\pm$ $0.14$ {\tiny $(0.11, 0.18)$} & $0.46$ {\tiny $(0.42, 0.51)$} $\pm$ $0.13$ {\tiny $(0.11, 0.16)$} & $0.43$ {\tiny $(0.40, 0.46)$} $\pm$ $0.10$ {\tiny $(0.08, 0.12)$} & $0.47$ {\tiny $(0.44, 0.49)$} $\pm$ $0.13$ {\tiny $(0.11, 0.15)$}\\
        & $\rho_p$ & $0.25$ {\tiny $(0.17, 0.32)$} $\pm$ $0.22$ {\tiny $(0.19, 0.26)$} & $0.20$ {\tiny $(0.14, 0.27)$} $\pm$ $0.19$ {\tiny $(0.16, 0.22)$} & $0.24$ {\tiny $(0.20, 0.28)$} $\pm$ $0.11$ {\tiny $(0.09, 0.14)$} & $0.23$ {\tiny $(0.20, 0.26)$} $\pm$ $0.18$ {\tiny $(0.16, 0.20)$}\\
    \midrule
    \multirow{2}{*}{20\% Reh.}
        & $\rho$ & $0.55$ {\tiny $(0.50, 0.60)$} $\pm$ $0.17$ {\tiny $(0.14, 0.22)$} & $0.57$ {\tiny $(0.53, 0.61)$} $\pm$ $0.12$ {\tiny $(0.10, 0.17)$} & $0.58$ {\tiny $(0.55, 0.60)$} $\pm$ $0.07$ {\tiny $(0.06, 0.09)$} & $0.57$ {\tiny $(0.54, 0.59)$} $\pm$ $0.13$ {\tiny $(0.11, 0.16)$}\\
        & $\rho_p$ & $0.13$ {\tiny $(0.04, 0.22)$} $\pm$ $0.27$ {\tiny $(0.23, 0.34)$} & $0.16$ {\tiny $(0.10, 0.22)$} $\pm$ $0.17$ {\tiny $(0.14, 0.23)$} & $0.23$ {\tiny $(0.19, 0.27)$} $\pm$ $0.11$ {\tiny $(0.09, 0.15)$} & $0.17$ {\tiny $(0.14, 0.21)$} $\pm$ $0.20$ {\tiny $(0.17, 0.24)$}\\
    \midrule
    \multirow{2}{*}{40\% Reh.}
        & $\rho$ & $0.53$ {\tiny $(0.47, 0.58)$} $\pm$ $0.18$ {\tiny $(0.15, 0.21)$} & $0.53$ {\tiny $(0.49, 0.57)$} $\pm$ $0.13$ {\tiny $(0.10, 0.16)$} & $0.57$ {\tiny $(0.54, 0.59)$} $\pm$ $0.08$ {\tiny $(0.06, 0.09)$} & $0.54$ {\tiny $(0.52, 0.57)$} $\pm$ $0.14$ {\tiny $(0.12, 0.16)$}\\
        & $\rho_p$ & $0.09$ {\tiny $(0.01, 0.17)$} $\pm$ $0.24$ {\tiny $(0.20, 0.28)$} & $0.16$ {\tiny $(0.11, 0.21)$} $\pm$ $0.17$ {\tiny $(0.14, 0.20)$} & $0.30$ {\tiny $(0.26, 0.33)$} $\pm$ $0.10$ {\tiny $(0.08, 0.12)$} & $0.18$ {\tiny $(0.15, 0.22)$} $\pm$ $0.20$ {\tiny $(0.17, 0.23)$}\\
    \midrule
    \addlinespace[7pt]
    \multirow{2}{*}{\textbf{Pooled}}
        & $\rho$ & $0.53$ {\tiny $(0.50, 0.56)$} $\pm$ $0.16$ {\tiny $(0.15, 0.18)$} & $0.52$ {\tiny $(0.50, 0.55)$} $\pm$ $0.13$ {\tiny $(0.12, 0.15)$} & $0.53$ {\tiny $(0.51, 0.55)$} $\pm$ $0.11$ {\tiny $(0.09, 0.12)$} & $-$\\
        & $\rho_p$ & $0.16$ {\tiny $(0.11, 0.20)$} $\pm$ $0.25$ {\tiny $(0.23, 0.28)$} & $0.17$ {\tiny $(0.14, 0.21)$} $\pm$ $0.18$ {\tiny $(0.16, 0.20)$} & $0.25$ {\tiny $(0.23, 0.28)$} $\pm$ $0.11$ {\tiny $(0.10, 0.13)$} & $-$\\
    \midrule
    \multicolumn{6}{c}{ $\Omega_{\mathrm{TIN}}^3$}\\
    \midrule
    \multirow{2}{*}{8\% Reh.}
        & $\rho$ & $0.40$ {\tiny $(0.35, 0.44)$} $\pm$ $0.15$ {\tiny $(0.12, 0.19)$} & $0.38$ {\tiny $(0.33, 0.42)$} $\pm$ $0.14$ {\tiny $(0.11, 0.19)$} & $-$ & $0.39$ {\tiny $(0.36, 0.42)$} $\pm$ $0.14$ {\tiny $(0.12, 0.17)$}\\
        & $\rho_p$ & $0.10$ {\tiny $(0.05, 0.15)$} $\pm$ $0.16$ {\tiny $(0.14, 0.18)$} & $0.11$ {\tiny $(0.07, 0.15)$} $\pm$ $0.13$ {\tiny $(0.10, 0.16)$} & $-$ & $0.10$ {\tiny $(0.07, 0.14)$} $\pm$ $0.14$ {\tiny $(0.13, 0.16)$}\\
    \midrule
    \multirow{2}{*}{20\% Reh.}
        & $\rho$ & $0.53$ {\tiny $(0.48, 0.56)$} $\pm$ $0.13$ {\tiny $(0.10, 0.18)$} & $0.50$ {\tiny $(0.47, 0.53)$} $\pm$ $0.09$ {\tiny $(0.07, 0.11)$} & $-$ & $0.51$ {\tiny $(0.49, 0.54)$} $\pm$ $0.11$ {\tiny $(0.10, 0.14)$}\\
        & $\rho_p$ & $0.10$ {\tiny $(0.05, 0.16)$} $\pm$ $0.16$ {\tiny $(0.13, 0.22)$} & $0.10$ {\tiny $(0.06, 0.13)$} $\pm$ $0.11$ {\tiny $(0.09, 0.15)$} & $-$ & $0.10$ {\tiny $(0.07, 0.13)$} $\pm$ $0.14$ {\tiny $(0.12, 0.18)$}\\
    \midrule
    \multirow{2}{*}{40\% Reh.}
        & $\rho$ & $0.44$ {\tiny $(0.39, 0.48)$} $\pm$ $0.13$ {\tiny $(0.11, 0.17)$} & $0.50$ {\tiny $(0.47, 0.53)$} $\pm$ $0.08$ {\tiny $(0.07, 0.12)$} & $-$ & $0.47$ {\tiny $(0.44, 0.50)$} $\pm$ $0.12$ {\tiny $(0.10, 0.14)$}\\
        & $\rho_p$ & $0.11$ {\tiny $(0.06, 0.17)$} $\pm$ $0.17$ {\tiny $(0.14, 0.22)$} & $0.27$ {\tiny $(0.24, 0.31)$} $\pm$ $0.11$ {\tiny $(0.09, 0.13)$} & $-$ & $0.20$ {\tiny $(0.16, 0.23)$} $\pm$ $0.16$ {\tiny $(0.14, 0.19)$}\\
    \midrule
    \addlinespace[7pt]
    \multirow{2}{*}{\textbf{Pooled}}
        & $\rho$ & $0.45$ {\tiny $(0.43, 0.48)$} $\pm$ $0.15$ {\tiny $(0.13, 0.17)$} & $0.46$ {\tiny $(0.44, 0.48)$} $\pm$ $0.12$ {\tiny $(0.10, 0.15)$} & $-$ & $-$\\
        & $\rho_p$ & $0.11$ {\tiny $(0.08, 0.14)$} $\pm$ $0.16$ {\tiny $(0.15, 0.19)$} & $0.16$ {\tiny $(0.14, 0.19)$} $\pm$ $0.14$ {\tiny $(0.13, 0.16)$} & $-$ & $-$\\
    \bottomrule
  \end{tabular}
  }
\end{table*}

\begin{table}[htbp]
  \centering
  \caption{Type-III two-way ANOVA results on our benchmarks for (i) the marginal associations between SIC and class-wise forgetting, measured by Spearman’s correlation $\rho$, and (ii) the difference between marginal and conditional associations between SIC and class-wise forgetting, where conditional associations are measured by partial Spearman’s correlation $\rho_p$. Factors include rehearsal retention percentage and percentage of new classes, along with their interaction. Reported statistics include the $F$-statistic, $p$-value, and partial eta squared ($\eta^2_p$).}
  \label{tab:anova_SIC}
  \resizebox{\linewidth}{!}{%
  \begin{tabular}{llccc}
    \toprule
    \multicolumn{5}{c}{\textbf{\emph{Last-Layer Imbalanced Forgetting Coefficient:}} SIC}\\
    \midrule
    & & {$F$-statistic} & $p$-value & {$\eta^2_p$}  \\
    \midrule
    \multicolumn{5}{c}{\textbf{Benchmark:} $\Omega_{\mathrm{C100}}$}\\
    \midrule
    \multicolumn{5}{c}{$\Omega_{\mathrm{C100}}^2$}\\
    \midrule
    \multirow{3}{*}{$\rho$}
        & Rehearsal retention percentage & $10.85$ & $2.67 \times 10^{-5}$ & $0.06$ \\
        & Percentage new classes & $4.22$ & $1.55 \times 10^{-2}$ & $0.02$ \\
        & Interaction & $1.66$ & $1.60 \times 10^{-1}$ & $0.02$ \\
    \midrule
    \multirow{3}{*}{$\rho-\rho_p$}
        & Rehearsal retention percentage & $7.66$ & $5.54 \times 10^{-4}$ & $0.04$ \\
        & Percentage new classes & $0.26$ & $7.69 \times 10^{-1}$ & $0.00$ \\
        & Interaction & $0.42$ & $7.95 \times 10^{-1}$ & $0.00$\\
    \midrule
    \multicolumn{5}{c}{$\Omega_{\mathrm{C100}}^{2; \,\{10\%, 20\%\}}$}\\
    \midrule
    \multirow{3}{*}{$\rho$}
        & Rehearsal retention percentage & $10.85$ & $3.11 \times 10^{-5}$ & $0.08$ \\
        & Percentage new classes & $0.25$ & $6.19 \times 10^{-1}$ & $0.00$ \\
        & Interaction & $0.69$ & $5.03 \times 10^{-1}$ & $0.01$ \\
    \midrule
    \multirow{3}{*}{$\rho-\rho_p$}
        & Rehearsal retention percentage & $7.66$ & $5.99 \times 10^{-4}$ & $0.06$ \\
        & Percentage new classes & $0.44$ & $5.10 \times 10^{-1}$ & $0.00$ \\
        & Interaction & $0.03$ & $9.68 \times 10^{-1}$ & $0.00$ \\
    \midrule
    \multicolumn{5}{c}{$\Omega_{\mathrm{C100}}^3$}\\
    \midrule
    \multirow{3}{*}{$\rho$}
        & Rehearsal retention percentage & $28.78$ & $6.68 \times 10^{-12}$ & $0.20$ \\
        & Percentage new classes & $0.01$ & $9.18 \times 10^{-1}$ & $0.00$ \\
        & Interaction & $1.39$ & $2.50 \times 10^{-1}$ & $0.01$ \\
    \midrule
    \multirow{3}{*}{$\rho-\rho_p$}
        & Rehearsal retention percentage & $10.50$ & $4.29 \times 10^{-5}$ & $0.08$ \\
        & Percentage new classes & $0.18$ & $6.75 \times 10^{-1}$ & $0.00$ \\
        & Interaction & $0.11$ & $9.00 \times 10^{-1}$ & $0.00$ \\
    \midrule
    \multicolumn{5}{c}{\textbf{Benchmark:} $\Omega_{\mathrm{TIN}}$}\\
    \midrule
    \multicolumn{5}{c}{$\Omega_{\mathrm{TIN}}^2$}\\
    \midrule
    \multirow{3}{*}{$\rho$}
        & Rehearsal retention percentage & $8.79$ & $1.89 \times 10^{-4}$ & $0.05$ \\
        & Percentage new classes & $60.99$ & $1.84 \times 10^{-23}$ & $0.26$ \\
        & Interaction & $2.25$ & $6.38 \times 10^{-2}$ & $0.02$ \\
    \midrule
    \multirow{3}{*}{$\rho-\rho_p$}
        & Rehearsal retention percentage & $0.08$ & $9.23 \times 10^{-1}$ & $0.00$ \\
        & Percentage new classes & $3.67$ & $2.64 \times 10^{-2}$ & $0.02$ \\
        & Interaction &$7.03$ & $1.90 \times 10^{-5}$ & $0.07$\\
    \midrule
    \multicolumn{5}{c}{$\Omega_{\mathrm{TIN}}^{2; \,\{10\%, 20\%\}}$}\\
    \midrule
    \multirow{3}{*}{$\rho$}
        & Rehearsal retention percentage & $8.79$ & $2.09 \times 10^{-4}$ & $0.07$ \\
        & Percentage new classes & $19.16$ & $1.82 \times 10^{-5}$ & $0.08$ \\
        & Interaction & $1.36$ & $2.59 \times 10^{-1}$ & $0.01$ \\
    \midrule
    \multirow{3}{*}{$\rho-\rho_p$}
        & Rehearsal retention percentage & $0.08$ & $9.23 \times 10^{-1}$ & $0.00$ \\
        & Percentage new classes & $6.45$ & $1.18 \times 10^{-2}$ & $0.03$ \\
        & Interaction & $2.58$ & $7.80 \times 10^{-2}$ & $0.02$ \\
    \midrule
    \multicolumn{5}{c}{$\Omega_{\mathrm{TIN}}^3$}\\
    \midrule
    \multirow{3}{*}{$\rho$}
        & Rehearsal retention percentage & $21.12$ & $3.69 \times 10^{-9}$ & $0.15$ \\
        & Percentage new classes & $57.24$ & $8.78 \times 10^{-13}$ & $0.20$ \\
        & Interaction & $2.02$ & $1.35 \times 10^{-1}$ & $0.02$ \\
    \midrule
    \multirow{3}{*}{$\rho-\rho_p$}
        & Rehearsal retention percentage & $4.59$ & $1.10 \times 10^{-2}$ & $0.04$ \\
        & Percentage new classes & $10.16$ & $1.63 \times 10^{-3}$ & $0.04$\\
        & Interaction & $6.09$ & $2.64 \times 10^{-3}$ & $0.05$\\
    \bottomrule
    \end{tabular}
  }
\end{table}

\begin{table}[htbp]
  \centering
  \caption{Type-III two-way ANOVA results on our benchmarks for (i) the marginal associations between CIC and class-wise forgetting, measured by Spearman’s correlation $\rho$, and (ii) the difference between marginal and conditional associations between CIC and class-wise forgetting, where conditional associations are measured by partial Spearman’s correlation $\rho_p$. Factors include rehearsal retention percentage and percentage of new classes, along with their interaction. Reported statistics include the $F$-statistic, $p$-value, and partial eta squared ($\eta^2_p$).}
  \label{tab:anova_CIC}
  \resizebox{\linewidth}{!}{%
  \begin{tabular}{llccc}
    \toprule
    \multicolumn{5}{c}{\textbf{\emph{Last-Layer Imbalanced Forgetting Coefficient:}} CIC}\\
    \midrule
    & & {$F$-statistic} & $p$-value & {$\eta^2_p$}  \\
    \midrule
    \multicolumn{5}{c}{\textbf{Benchmark:} $\Omega_{\mathrm{C100}}$}\\
    \midrule
    \multicolumn{5}{c}{$\Omega_{\mathrm{C100}}^2$}\\
    \midrule
    \multirow{3}{*}{$\rho$}
        & Rehearsal retention percentage & $1.51$ & $2.22 \times 10^{-1}$ & $0.01$ \\
        & Percentage new classes & $2.11$ & $1.23 \times 10^{-1}$ & $0.01$  \\
        & Interaction & $1.05$ & $3.82 \times 10^{-1}$ & $0.01$  \\
    \midrule
    \multirow{3}{*}{$\rho-\rho_p$}
        & Rehearsal retention percentage & $0.25$ & $7.82 \times 10^{-1}$ & $0.00$\\
        & Percentage new classes & $0.75$ & $4.75 \times 10^{-1}$ & $0.00$\\
        & Interaction & $0.54$ & $7.09 \times 10^{-1}$ & $0.01$  \\
    \midrule
    \multicolumn{5}{c}{$\Omega_{\mathrm{C100}}^{2; \,\{10\%, 20\%\}}$}\\
    \midrule
    \multirow{3}{*}{$\rho$}
        & Rehearsal retention percentage & $1.51$ & $2.23 \times 10^{-1}$ & $0.01$  \\
        & Percentage new classes & $2.63$ & $1.06 \times 10^{-1}$ & $0.01$  \\
        & Interaction & $0.44$ & $6.44 \times 10^{-1}$ & $0.00$\\
    \midrule
    \multirow{3}{*}{$\rho-\rho_p$}
        & Rehearsal retention percentage & $0.25$ & $7.82 \times 10^{-1}$ & $0.00$  \\
        & Percentage new classes & $1.11$ & $2.94 \times 10^{-1}$ & $0.00$  \\
        & Interaction & $0.82$ & $4.41 \times 10^{-1}$ & $0.01$  \\
    \midrule
    \multicolumn{5}{c}{$\Omega_{\mathrm{C100}}^3$}\\
    \midrule
    \multirow{3}{*}{$\rho$}
        & Rehearsal retention percentage & $15.79$ & $3.68 \times 10^{-7}$ & $0.12$ \\
        & Percentage new classes & $5.77$ & $1.71 \times 10^{-2}$ & $0.02$  \\
        & Interaction & $1.38$ & $2.53 \times 10^{-1}$ & $0.01$  \\
    \midrule
    \multirow{3}{*}{$\rho-\rho_p$}
        & Rehearsal retention percentage & $10.79$ & $3.30 \times 10^{-5}$ & $0.08$  \\
        & Percentage new classes & $0.30$ & $5.83 \times 10^{-1}$ & $0.00$  \\
        & Interaction & $1.56$ & $2.13 \times 10^{-1}$ & $0.01$ \\
    \midrule
    \multicolumn{5}{c}{\textbf{Benchmark:} $\Omega_{\mathrm{TIN}}$}\\
    \midrule
    \multicolumn{5}{c}{$\Omega_{\mathrm{TIN}}^2$}\\
    \midrule
    \multirow{3}{*}{$\rho$}
        & Rehearsal retention percentage & $3.66$ & $2.69 \times 10^{-2}$ & $0.02$ \\
        & Percentage new classes & $1.47$ & $2.31 \times 10^{-1}$ & $0.01$  \\
        & Interaction & $3.20$ & $1.34 \times 10^{-2}$ & $0.04$  \\
    \midrule
    \multirow{3}{*}{$\rho-\rho_p$}
        & Rehearsal retention percentage & $10.04$ & $5.80 \times 10^{-5}$ & $0.05$  \\
        & Percentage new classes & $1.12$ & $3.29 \times 10^{-1}$ & $0.01$  \\
        & Interaction & $3.28$ & $1.17 \times 10^{-2}$ & $0.04$  \\
    \midrule
    \multicolumn{5}{c}{$\Omega_{\mathrm{TIN}}^{2; \,\{10\%, 20\%\}}$}\\
    \midrule
    \multirow{3}{*}{$\rho$}
        & Rehearsal retention percentage & $3.66$ & $2.73 \times 10^{-2}$ & $0.03$  \\
        & Percentage new classes & $0.96$ & $3.29 \times 10^{-1}$ & $0.00$ \\
        & Interaction & $2.02$ & $1.35 \times 10^{-1}$ & $0.02$  \\
    \midrule
    \multirow{3}{*}{$\rho-\rho_p$}
        & Rehearsal retention percentage & $10.04$ & $6.60 \times 10^{-5}$ & $0.08$  \\
        & Percentage new classes & $0.04$ & $8.39 \times 10^{-1}$ & $0.00$  \\
        & Interaction & $3.22$ & $4.16 \times 10^{-2}$ & $0.03$  \\
    \midrule
    \multicolumn{5}{c}{$\Omega_{\mathrm{TIN}}^3$}\\
    \midrule
    \multirow{3}{*}{$\rho$}
        & Rehearsal retention percentage & $67.08$ & $9.38 \times 10^{-24}$ & $0.36$  \\
        & Percentage new classes & $2.78$ & $9.70 \times 10^{-2}$ & $0.01$  \\
        & Interaction & $1.79$ & $1.69 \times 10^{-1}$ & $0.02$ \\
    \midrule
    \multirow{3}{*}{$\rho-\rho_p$}
        & Rehearsal retention percentage & $5.17$ & $6.36 \times 10^{-3}$ & $0.04$  \\
        & Percentage new classes & $0.14$ & $7.11 \times 10^{-1}$ & $0.00$  \\
        & Interaction & $3.75$ & $2.50 \times 10^{-2}$ & $0.03$  \\
    \bottomrule
\end{tabular}
  }
\end{table}

\begin{table}[htbp]
  \centering
  \caption{Type-III two-way ANOVA results on our benchmarks for (i) the marginal associations between NIC and class-wise forgetting, measured by Spearman’s correlation $\rho$, and (ii) the difference between marginal and conditional associations between NIC and class-wise forgetting, where conditional associations are measured by partial Spearman’s correlation $\rho_p$. Factors include rehearsal retention percentage and percentage of new classes, along with their interaction. Reported statistics include the $F$-statistic, $p$-value, and partial eta squared ($\eta^2_p$).}
  \label{tab:anova_NIC}
  \resizebox{\linewidth}{!}{%
  \begin{tabular}{llccc}
    \toprule
    \multicolumn{5}{c}{\textbf{\emph{Last-Layer Imbalanced Forgetting Coefficient:}} NIC}\\
    \midrule
    & & {$F$-statistic} & $p$-value & {$\eta^2_p$}  \\
    \midrule
    \multicolumn{5}{c}{\textbf{Benchmark:} $\Omega_{\mathrm{C100}}$}\\
    \midrule
    \multicolumn{5}{c}{$\Omega_{\mathrm{C100}}^2$}\\
    \midrule
    \multirow{3}{*}{$\rho$}
        & Rehearsal retention percentage &$5.37$ & $5.05 \times 10^{-3}$ & $0.03$  \\
        & Percentage new classes & $1.65$ & $1.94 \times 10^{-1}$ & $0.01$  \\
        & Interaction & $0.74$ & $5.68 \times 10^{-1}$ & $0.01$  \\
    \midrule
    \multirow{3}{*}{$\rho-\rho_p$}
        & Rehearsal retention percentage & $0.03$ & $9.70 \times 10^{-1}$ & $0.00$ \\
        & Percentage new classes & $2.53$ & $8.12 \times 10^{-2}$ & $0.01$ \\
        & Interaction & $1.00$ & $4.09 \times 10^{-1}$ & $0.01$
   \\
    \midrule
    \multicolumn{5}{c}{$\Omega_{\mathrm{C100}}^{2; \,\{10\%, 20\%\}}$}\\
    \midrule
    \multirow{3}{*}{$\rho$}
        & Rehearsal retention percentage & $5.37$ & $5.25 \times 10^{-3}$ & $0.04$  \\
        & Percentage new classes & $2.84$ & $9.31 \times 10^{-2}$ & $0.01$   \\
        & Interaction & $1.39$ & $2.52 \times 10^{-1}$ & $0.01$  \\
    \midrule
    \multirow{3}{*}{$\rho-\rho_p$}
        & Rehearsal retention percentage & $0.03$ & $9.70 \times 10^{-1}$ & $0.00$   \\
        & Percentage new classes & $0.01$ & $9.05 \times 10^{-1}$ & $0.00$   \\
        & Interaction & $0.78$ & $4.60 \times 10^{-1}$ & $0.01$  \\
    \midrule
    \multicolumn{5}{c}{$\Omega_{\mathrm{C100}}^3$}\\
    \midrule
    \multirow{3}{*}{$\rho$}
        & Rehearsal retention percentage & $3.00$ & $5.15 \times 10^{-2}$ & $0.03$ \\
        & Percentage new classes & $0.55$ & $4.60 \times 10^{-1}$ & $0.00$   \\
        & Interaction & $0.77$ & $4.66 \times 10^{-1}$ & $0.01$   \\
    \midrule
    \multirow{3}{*}{$\rho-\rho_p$}
        & Rehearsal retention percentage & $2.44$ & $8.91 \times 10^{-2}$ & $0.02$\\
        & Percentage new classes & $2.39$ & $1.23 \times 10^{-1}$ & $0.01$   \\
        & Interaction & $1.63$ & $1.98 \times 10^{-1}$ & $0.01$  \\
    \midrule
    \multicolumn{5}{c}{\textbf{Benchmark:} $\Omega_{\mathrm{TIN}}$}\\
    \midrule
    \multicolumn{5}{c}{$\Omega_{\mathrm{TIN}}^2$}\\
    \midrule
    \multirow{3}{*}{$\rho$}
        & Rehearsal retention percentage & $0.62$ & $5.39 \times 10^{-1}$ & $0.00$ \\
        & Percentage new classes & $3.16$ & $4.35 \times 10^{-2}$ & $0.02$   \\
        & Interaction & $3.15$ & $1.45 \times 10^{-2}$ & $0.03$  \\
    \midrule
    \multirow{3}{*}{$\rho-\rho_p$}
        & Rehearsal retention percentage & $7.19$ & $8.73 \times 10^{-4}$ & $0.04$   \\
        & Percentage new classes & $3.89$ & $2.13 \times 10^{-2}$ & $0.02$   \\
        & Interaction & $1.09$ & $3.62 \times 10^{-1}$ & $0.01$   \\
    \midrule
    \multicolumn{5}{c}{$\Omega_{\mathrm{TIN}}^{2; \,\{10\%, 20\%\}}$}\\
    \midrule
    \multirow{3}{*}{$\rho$}
        & Rehearsal retention percentage & $0.62$ & $5.39 \times 10^{-1}$ & $0.01$   \\
        & Percentage new classes & $1.61$ & $2.05 \times 10^{-1}$ & $0.01$  \\
        & Interaction & $1.24$ & $2.90 \times 10^{-1}$ & $0.01$  \\
    \midrule
    \multirow{3}{*}{$\rho-\rho_p$}
        & Rehearsal retention percentage & $7.19$ & $9.36 \times 10^{-4}$ & $0.06$   \\
        & Percentage new classes & $0.00$ & $9.77 \times 10^{-1}$ & $0.00$   \\
        & Interaction & $0.44$ & $6.43 \times 10^{-1}$ & $0.00$   \\
    \midrule
    \multicolumn{5}{c}{$\Omega_{\mathrm{TIN}}^3$}\\
    \midrule
    \multirow{3}{*}{$\rho$}
        & Rehearsal retention percentage & $8.89$ & $1.90 \times 10^{-4}$ & $0.07$   \\
        & Percentage new classes & $0.21$ & $6.45 \times 10^{-1}$ & $0.00$   \\
        & Interaction & $3.80$ & $2.38 \times 10^{-2}$ & $0.03$  \\
    \midrule
    \multirow{3}{*}{$\rho-\rho_p$}
        & Rehearsal retention percentage & $10.52$ & $4.22 \times 10^{-5}$ & $0.08$   \\
        & Percentage new classes & $0.52$ & $4.73 \times 10^{-1}$ & $0.00$   \\
        & Interaction & $3.19$ & $4.29 \times 10^{-2}$ & $0.03$   \\
    \bottomrule
\end{tabular}
  }
\end{table}

\section{Temporal Evolution of the Individual Predictive Strength of the Last-Layer Imbalanced Forgetting Coefficients}\label{sec:temporal_evolution_indiv_strength_imbalanced_forgetting}
\noindent In Section \ref{sec:individual_pred_strength}, we analyzed how effectively each \emph{Last-layer Imbalanced Forgetting Coefficient} predicts the ranking of past classes by their forgetting across our benchmarks. This was accomplished by computing the marginal and conditional associations between each coefficient and class-wise forgetting for the second and third incremental steps in both $\Omega_{\mathrm{C100}}$ and $\Omega_{\mathrm{TIN}}$. Here, we examine how these associations vary when, instead of being approximated over the full training trajectory of an incremental step (as described in Section \ref{sec:comput_imbalanced_forg_coeff_benchmarks}), the coefficients are computed using progressively more R-SGD checkpoints from the start of training on that step.

\begin{figure*}[!t]
  \centering
  \includegraphics[width=0.9\textwidth]{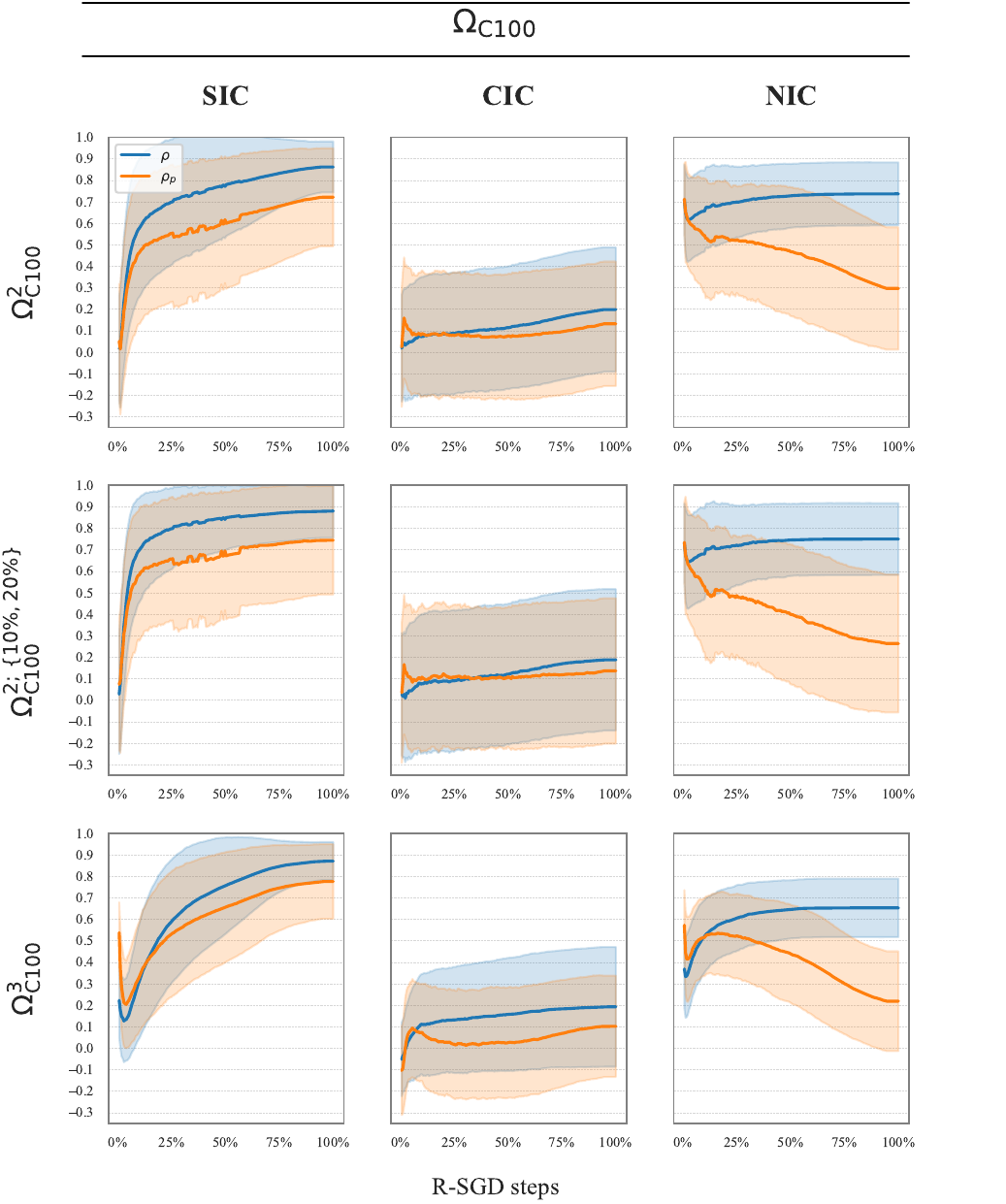}
  \caption{Line plots showing, for the second and third incremental steps in $\Omega_{\mathrm{C100}}$, how the estimated mean and standard deviation of the predictive strength achieved by each \emph{Last-Layer Imbalanced Forgetting Coefficient} evolve as progressively more R-SGD checkpoints from the start of training are used. The predictive strength is with respect to ranking past classes by their forgetting and is measured using Spearman’s correlation $\rho$ and partial Spearman’s correlation $\rho_p$.}
  \label{fig:SIC_CIC_NIC_across_steps_cifar}
\end{figure*}

\begin{figure*}[!t]
  \centering
  \includegraphics[width=0.9\textwidth]{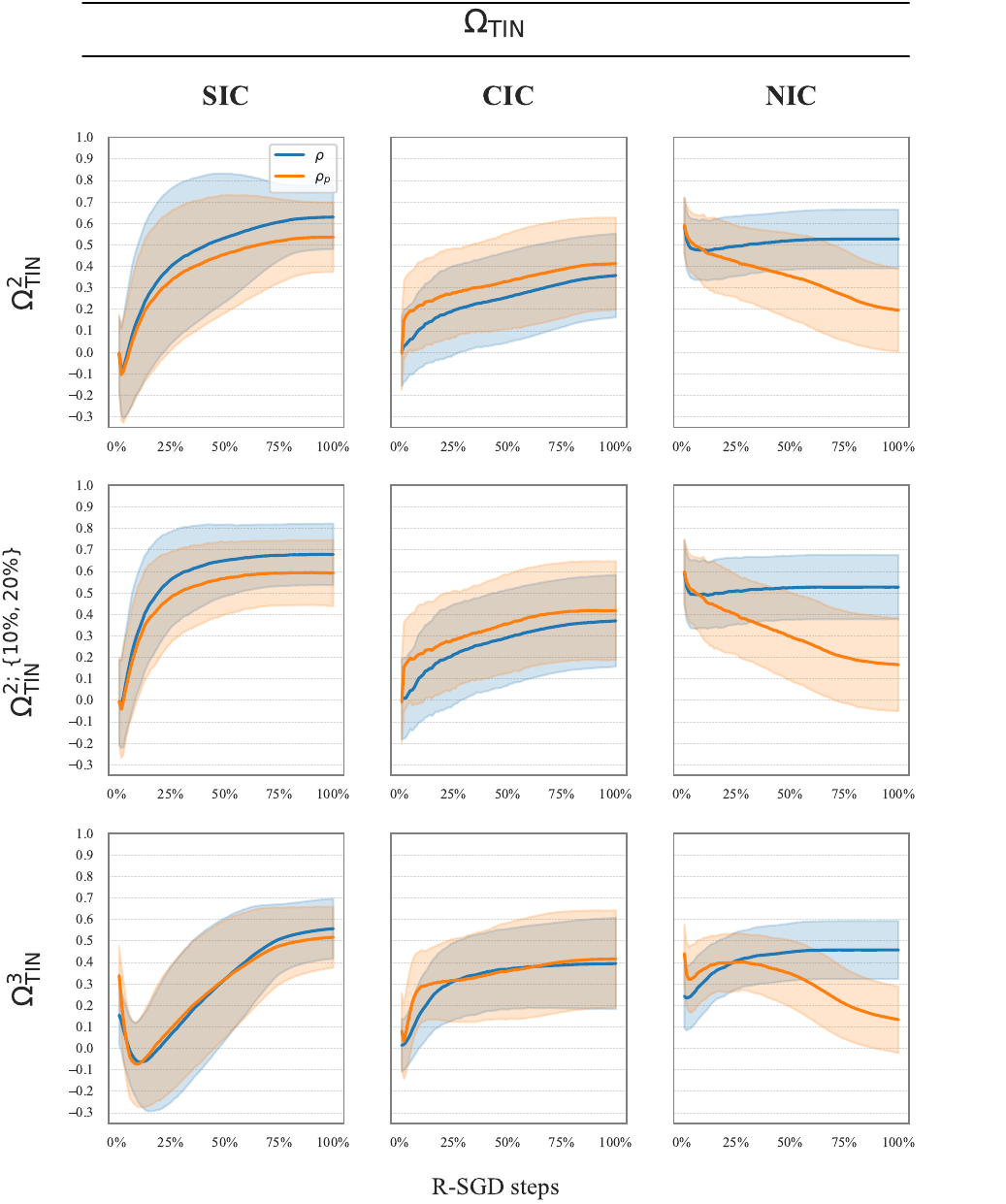}
  \caption{Line plots showing, for the second and third incremental steps in $\Omega_{\mathrm{TIN}}$, how the estimated mean and standard deviation of the predictive strength achieved by each \emph{Last-Layer Imbalanced Forgetting Coefficient} evolve as progressively more R-SGD checkpoints from the start of training are used. The predictive strength is with respect to ranking past classes by their forgetting and is measured using Spearman’s correlation $\rho$ and partial Spearman’s correlation $\rho_p$.}
\label{fig:SIC_CIC_NIC_across_steps_tinyin}
\end{figure*}

Figure \ref{fig:SIC_CIC_NIC_across_steps_cifar} and \ref{fig:SIC_CIC_NIC_across_steps_tinyin} show that, although including progressively more checkpoints increases the mean marginal associations achieved by our coefficients, using only early checkpoints (roughly the initial 25\%) yields comparable performance to that obtained with all checkpoints. This suggests that the last-layer
gradient-level interactions captured by our coefficients during the early training phase are almost as informative as those captured over the full training trajectory. The mean conditional associations of SIC and CIC follow a similar pattern to their marginal counterparts. In contrast, the mean conditional associations of NIC drop dramatically as more checkpoints are included, indicating increasing overlap between the predictive information captured by NIC and that captured by both SIC and CIC.

\section{Finer-Grained Analysis of the Joint Predictive Strength of the Last-Layer Imbalanced Forgetting Coefficients}\label{sec:finer_grained_anal_joint_pred_strength}

\noindent In Section \ref{sec:analysis_joint_pred_strength}, we analyzed the joint predictive strength of the \emph{Last-Layer Imbalanced Forgetting Coefficients} with respect to ranking past classes by their forgetting. This was performed for the second and third incremental steps in both $\hat{\Omega}_{\mathrm{C100}}$ and $\hat{\Omega}_{\mathrm{TIN}}$, aggregating across different percentages of rehearsal retention and newly introduced classes. Here, we provide a finer-grained analysis that explicitly characterizes how this joint predictive strength varies with these proportions.

Specifically, Table \ref{tab:spearman_All_coef_SIC_fine_grained} reports the mean and standard deviation of this joint predictive strength for each partition in $\hat{\Omega}_d^k$, with $d \in \{\mathrm{C100}, \mathrm{TIN}\}$ and $k \in \{2,3\}$. The results show that increasing both rehearsal percentage and proportion of new classes leads, on average, to lower joint predictive strength. Moreover, the standard deviations are relatively small compared to the means, indicating that the joint predictive strength is very similar across incremental steps of the same partition.

Furthermore, for each $\hat{\Omega}_d^k$, we also conduct a type-III two-way ANOVA to assess whether rehearsal retention percentage, percentage of newly introduced classes, and their interaction have statistically significant effects on the joint predictive strength obtained in $\hat{\Omega}_d^k$, and to quantify the corresponding effect sizes. To ensure robustness to potential heteroskedasticity, we perform the ANOVA using the HC3 heteroskedasticity-consistent covariance estimator. The results of this ANOVA are reported in Table \ref{tab:anova_all_coefficients}. Overall, the rehearsal retention percentage factor exhibits the strongest effect and consistently yields statistically significant contributions to the joint predictive strength.

\begin{table*}[htbp]
  \centering
  \caption{Mean $\pm$ standard deviation of the predictive strength achieved by a linear model using all \emph{Last-Layer Imbalanced Forgetting Coefficients}, evaluated under the SW-LOO protocol on each partition in our sampled benchmarks. The predictive strength is with respect to ranking past classes by their forgetting and is measured using Spearman’s correlation $\rho$. The predictive strength obtained on the same incremental steps using only raw SIC values is also reported. "Pooled" values are obtained from the union of partitions sharing the same rehearsal retention percentage or percentage of newly introduced classes.
}
  \label{tab:spearman_All_coef_SIC_fine_grained}
  \resizebox{0.75\textwidth}{!}{%
  \begin{tabular}{llccc@{\hskip 20pt}c}
    \toprule
    & & 10\% Classes & 20\% Classes & 50\% Classes  & \textbf{Pooled}\\
    \midrule
    \multicolumn{6}{c}{\textbf{Sampled benchmark:} $\hat{\Omega}_{\mathrm{C100}}$}\\
    \midrule
    \multicolumn{6}{c}{$\hat{\Omega}_{\mathrm{C100}}^2$}\\
    \midrule
    \multirow{2}{*}{8\% Reh.}
        & All coefficients & $0.92$ $\pm$ $0.07$
 & $0.93$ $\pm$ $0.04$
 & $0.90$ $\pm$ $0.03$
 & $0.92$ $\pm$ $0.05$
\\
        & SIC &$0.92$  $\pm$ $0.07$  & $0.92$  $\pm$ $0.05$  & $0.88$  $\pm$ $0.04$  & $0.91$  $\pm$ $0.06$ \\
    \midrule
    \multirow{2}{*}{20\% Reh.}
        & All coefficients & $0.96$ $\pm$ $0.12$
 & $0.91$ $\pm$ $0.04$
 & $0.87$ $\pm$ $0.04$ & $0.92$ $\pm$ $0.08$
\\
        & SIC & $0.91$  $\pm$ $0.16$  & $0.89$  $\pm$ $0.06$  & $0.83$  $\pm$ $0.06$  & $0.88$  $\pm$ $0.11$ \\
    \midrule
    \multirow{2}{*}{40\% Reh.}
        & All coefficients & $0.85$ $\pm$ $0.09$
 & $0.84$ $\pm$ $0.09$
 & $0.80$ $\pm$ $0.05$
 & $0.83$ $\pm$ $0.08$
\\
        & SIC & $0.82$  $\pm$ $0.14$  & $0.78$  $\pm$ $0.13$  & $0.70$  $\pm$ $0.08$ & $0.77$  $\pm$ $0.12$ \\
    \midrule
     \addlinespace[7pt]
     \multirow{2}{*}{\textbf{Pooled}}
        & All coefficients & $0.92$ $\pm$ $0.10$
 & $0.90$ $\pm$ $0.08$
 & $0.86$ $\pm$ $0.06$
 & $-$\\
        & SIC & $0.89$  $\pm$ $0.14$  & $0.87$  $\pm$ $0.11$  & $0.82$ $\pm$ $0.10$ & $-$\\
    \midrule
    \multicolumn{6}{c}{$\hat{\Omega}_{\mathrm{C100}}^3$}\\
    \midrule
    \multirow{2}{*}{8\% Reh.}
        & All coefficients & $0.93$ $\pm$ $0.04$
 & $0.92$ $\pm$ $0.02$
 & $-$ & $0.93$ $\pm$ $0.03$
\\
        & SIC & $0.92$  $\pm$ $0.04$  & $0.92$  $\pm$ $0.03$  & $-$ & $0.92$  $\pm$ $0.04$ \\
    \midrule
    \multirow{2}{*}{20\% Reh.}
        & All coefficients & $0.91$ $\pm$ $0.08$
 & $0.89$ $\pm$ $0.05$
 & $-$ & $0.90$ $\pm$ $0.07$
\\
        & SIC & $0.90$  $\pm$ $0.08$  & $0.87$  $\pm$ $0.05$  & $-$ & $0.88$  $\pm$ $0.07$ \\
    \midrule
    \multirow{2}{*}{40\% Reh.}
        & All coefficients & $0.81$ $\pm$ $0.09$
 & $0.80$ $\pm$ $0.06$
 & $-$ & $0.81$ $\pm$ $0.08$
\\
        & SIC & $0.80$  $\pm$ $0.09$  & $0.76$  $\pm$ $0.07$  & $-$ & $0.78$  $\pm$ $0.08$\\
    \midrule
    \addlinespace[7pt]
    \multirow{2}{*}{\textbf{Pooled}}
        & All coefficients & $0.89$ $\pm$ $0.09$
 & $0.88$ $\pm$ $0.07$
 & $-$ & $-$\\
        & SIC & $0.88$  $\pm$ $0.09$  & $0.86$  $\pm$ $0.09$  & $-$ & $-$\\
    \midrule
    \multicolumn{6}{c}{\textbf{Sampled benchmark:} $\hat{\Omega}_{\mathrm{TIN}}$}\\
    \midrule
    \multicolumn{6}{c}{ $\hat{\Omega}_{\mathrm{TIN}}^2$}\\
    \midrule
    \multirow{2}{*}{8\% Reh.}
        & All coefficients & $0.78$ $\pm$ $0.10$
 & $0.76$ $\pm$ $0.09$
 & $0.70$ $\pm$ $0.07$
 & $0.75$ $\pm$ $0.10$
\\
        & SIC & $0.67$  $\pm$ $0.11$  & $0.56$  $\pm$ $0.13$  & $0.42$  $\pm$ $0.09$  & $0.56$  $\pm$ $0.15$ \\
    \midrule
    \multirow{2}{*}{20\% Reh.}
        & All coefficients & $0.82$ $\pm$ $0.09$
 & $0.79$ $\pm$ $0.08$
 & $0.73$ $\pm$ $0.05$
 & $0.79$ $\pm$ $0.08$
\\
        & SIC & $0.78$  $\pm$ $0.12$  & $0.71$  $\pm$ $0.10$  & $0.59$  $\pm$ $0.07$  & $0.70$ $\pm$ $0.12$ \\
    \midrule
    \multirow{2}{*}{40\% Reh.}
        & All coefficients & $0.73$ $\pm$ $0.14$
 & $0.68$ $\pm$ $0.09$
 & $0.64$ $\pm$ $0.07$
 & $0.68$ $\pm$ $0.11$
\\
        & SIC & $0.70$  $\pm$ $0.17$  & $0.61$  $\pm$ $0.12$  & $0.52$  $\pm$ $0.07$  & $0.62$ $\pm$ $0.14$ \\
    \midrule
    \addlinespace[7pt]
    \multirow{2}{*}{\textbf{Pooled}}
        & All coefficients & $0.78$ $\pm$ $0.12$
 & $0.75$ $\pm$ $0.10$
 & $0.69$ $\pm$ $0.08$
 & $-$\\
        & SIC & $0.72$  $\pm$ $0.14$   & $0.63$  $\pm$ $0.13$  & $0.51$  $\pm$ $0.11$ & $-$\\
    \midrule
    \multicolumn{6}{c}{ $\hat{\Omega}_{\mathrm{TIN}}^3$}\\
    \midrule
    \multirow{2}{*}{8\% Reh.}
        & All coefficients & $0.77$ $\pm$ $0.07$
 & $0.74$ $\pm$ $0.06$
 & $-$ & $0.76$ $\pm$ $0.07$
\\
        & SIC & $0.56$  $\pm$ $0.11$  & $0.38$  $\pm$ $0.10$  & $-$ & $0.48$  $\pm$ $0.14$ \\
    \midrule
    \multirow{2}{*}{20\% Reh.}
        & All coefficients & $0.79$ $\pm$ $0.07$
 & $0.70$ $\pm$ $0.06$
 & $-$ & $0.75$ $\pm$ $0.08$
\\
        & SIC & $0.72$  $\pm$ $0.10$  & $0.55$  $\pm$ $0.08$  & $-$ & $0.64$  $\pm$ $0.12$ \\
    \midrule
    \multirow{2}{*}{40\% Reh.}
        & All coefficients & $0.65$ $\pm$ $0.10$
 & $0.60$ $\pm$ $0.07$
 & $-$ & $0.63$ $\pm$ $0.09$
\\
        & SIC & $0.60$  $\pm$ $0.12$  & $0.48$  $\pm$ $0.07$  & $-$ & $0.55$  $\pm$ $0.11$ \\
    \midrule
    \addlinespace[7pt]
    \multirow{2}{*}{\textbf{Pooled}}
        & All coefficients & $0.74$ $\pm$ $0.10$
 & $0.68$ $\pm$ $0.08$
 & $-$ & $-$\\
        & SIC & $0.63$  $\pm$ $0.13$  & $0.47$  $\pm$ $0.11$  & $-$ & $-$\\
    \bottomrule
\end{tabular}
  }
\end{table*}

\begin{table}[htbp]
  \centering
  \caption{Type-III two-way ANOVA results for the predictive strength achieved by a linear model using all \emph{Last-Layer Imbalanced Forgetting Coefficients}, evaluated under the SW-LOO protocol on our sampled benchmarks, using rehearsal retention percentage and percentage of new classes as factors, together with their interaction term. The predictive strength is with respect to ranking past classes by their forgetting and is measured using Spearman’s correlation $\rho$. Reported statistics include the $F$-statistic, $p$-value, and partial eta squared ($\eta^2_p$).}
  \label{tab:anova_all_coefficients}
  \resizebox{\linewidth}{!}{%
  \begin{tabular}{llccc}
    \toprule
    & & {$F$-statistic} & $p$-value & {$\eta^2_p$}  \\
    \midrule
    \multicolumn{5}{c}{\textbf{Sampled benchmark:} $\hat{\Omega}_{\mathrm{C100}}$}\\
    \midrule
    \multirow{3}{*}{$\hat{\Omega}_{\mathrm{C100}}^2$}
        & Rehearsal retention percentage & $6.22$ & $2.21 \times 10^{-3}$ & $0.03$  \\
        & Percentage new classes & $2.71$ & $6.79 \times 10^{-2}$ & $0.02$   \\
        & Interaction & $1.03$ & $3.91 \times 10^{-1}$ & $0.01$   \\
    \midrule
    \multirow{3}{*}{$\hat{\Omega}_{\mathrm{C100}}^{2; \,\{10\%, 20\%\}}$}
        & Rehearsal retention percentage & $6.22$ & $2.32 \times 10^{-3}$ & $0.05$   \\
        & Percentage new classes & $1.87$ & $1.73 \times 10^{-1}$ & $0.01$  \\
        & Interaction & $0.86$ & $4.24 \times 10^{-1}$ & $0.01$  \\
    \midrule
    \multirow{3}{*}{$\hat{\Omega}_{\mathrm{C100}}^3$}
        & Rehearsal retention percentage & $30.03$ & $2.46 \times 10^{-12}$ & $0.20$  \\
        & Percentage new classes & $0.42$ & $5.18 \times 10^{-1}$ & $0.00$  \\
        & Interaction & $0.41$ & $6.62 \times 10^{-1}$ & $0.00$  \\
    \midrule
    \multicolumn{5}{c}{\textbf{Sampled benchmark:} $\hat{\Omega}_{\mathrm{TIN}}$}\\
    \midrule
    \multirow{3}{*}{$\hat{\Omega}_{\mathrm{TIN}}^2$}
        & Rehearsal retention percentage & $8.36$ & $2.85 \times 10^{-4}$ & $0.05$ \\
        & Percentage new classes & $7.11$ & $9.41 \times 10^{-4}$ & $0.04$   \\
        & Interaction & $0.14$ & $9.67 \times 10^{-1}$ & $0.00$  \\
    \midrule
    \multirow{3}{*}{$\hat{\Omega}_{\mathrm{TIN}}^{2; \,\{10\%, 20\%\}}$}
        & Rehearsal retention percentage & $8.36$ & $3.12 \times 10^{-4}$ & $0.07$  \\
        & Percentage new classes & $0.65$ & $4.20 \times 10^{-1}$ & $0.00$ \\
        & Interaction & $0.04$ & $9.64 \times 10^{-1}$ & $0.00$   \\
    \midrule
    \multirow{3}{*}{$\hat{\Omega}_{\mathrm{TIN}}^3$}
        & Rehearsal retention percentage & $26.77$ & $3.40 \times 10^{-11}$ & $0.19$   \\
        & Percentage new classes & $3.58$ & $5.96 \times 10^{-2}$ & $0.02$   \\
        & Interaction & $3.95$ & $2.05 \times 10^{-2}$ & $0.03$  \\
    \bottomrule
\end{tabular}
  }
\end{table}

\section{Complementary Analysis of the Joint Predictive Strength of the Last-Layer Imbalanced Forgetting Coefficients}\label{sec:complementary_joint_pred_strength}

\noindent In Section \ref{sec:analysis_joint_pred_strength}, we showed, through the SW-LOO protocol, that a linear model incorporating all \emph{Last-Layer Imbalanced Forgetting Coefficients} predicts the forgetting-based ranking of past classes more accurately than SIC alone. Here, we complement that analysis by assessing how closely the model’s predictions match the ground-truth forgetting values. To this end, we again employ the SW-LOO protocol, using the mean absolute error (MAE) as the evaluation metric. For comparison, we perform the same evaluation with a linear model that includes only SIC as a predictor.

The results in Table \ref{tab:SW-LOO_protocol_MAE} indicate that the model incorporating SIC, CIC, and NIC produces predictions that are relatively close to the ground truth, and consistently closer than those obtained using SIC alone. However, the difference in MAE between the two models is less pronounced than the difference observed in their ability to predict the forgetting-based ranking.

For completeness, Table \ref{tab:coefficients} also reports the weights (the fixed-effects ones) obtained from fitting the linear model incorporating all \emph{Last-Layer Imbalanced Forgetting Coefficients} across different partitions in our benchmarks. As for the SW-LOO protocol, a mixed-effects formulation is used. We observe that, although the weights of this model vary across benchmark partitions, they are consistently positive, indicating that each \emph{Last-Layer Imbalanced Forgetting Coefficient} has a positive partial effect on class-wise forgetting.

\begin{table}[htbp]
  \centering
  \caption{Mean $\pm$ standard deviation of the predictive strength and numerical closeness to the ground truth values achieved by both a linear model using all \emph{Last-Layer Imbalanced Forgetting Coefficients} and a linear model using only SIC, evaluated under the SW-LOO protocol on the second and third incremental steps in $\hat{\Omega}_{\mathrm{C100}}$ and $\hat{\Omega}_{\mathrm{TIN}}$. The predictive strength is with respect to ranking past classes by their forgetting
  and is measured using Spearman’s correlation $\rho$. The numerical closeness to the ground truth values is measured using MAE. The predictive strength of the model using only SIC matches that obtained using the raw SIC values, as previously reported in Table \ref{tab:SW-LOO_protocol}.}
  \label{tab:SW-LOO_protocol_MAE}
  \renewcommand{\arraystretch}{0.5}
  \resizebox{0.85\linewidth}{!}{%
  \begin{tabular}{llcc}
    \toprule
    & & $\rho$ & MAE\\
    \midrule
    \multicolumn{2}{l}{} & \multicolumn{2}{c}{$\hat{\Omega}_{\mathrm{C100}}$}\\
    \midrule
    \multirow{2}{*}{$\hat{\Omega}^2_{\mathrm{C100}}$}
        & All coefficients & 0.90 $\pm$ 0.08  & 0.06 $\pm$ 0.01 \\
        & SIC & 0.86  $\pm$ 0.12  & 0.11  $\pm$ 0.11 \\ 
    \midrule
    \multirow{2}{*}{$\hat{\Omega}^{2;\,\{10\%, 20\%\}}_{\mathrm{C100}}$}
        & All coefficients & 0.91 $\pm$ 0.09 & 0.05 $\pm$ 0.01 \\
        & SIC & 0.88  $\pm$ 0.13  & 0.06  $\pm$ 0.02 \\
    \midrule
    \multirow{2}{*}{$\hat{\Omega}^3_{\mathrm{C100}}$}
        & All coefficients & 0.88 $\pm$ 0.08 & 0.06 $\pm$ 0.01 \\
        & SIC & 0.87  $\pm$ 0.09  & 0.06  $\pm$ 0.01 \\
    \midrule
    \multicolumn{2}{l}{} & \multicolumn{2}{c}{$\hat{\Omega}_{\mathrm{TIN}}$}\\
    \midrule
    \multirow{2}{*}{$\hat{\Omega}^2_{\mathrm{TIN}}$}
        & All coefficients & 0.74 $\pm$ 0.11  & 0.09 $\pm$ 0.02 \\
        & SIC & 0.63  $\pm$ 0.15  & 0.17  $\pm$ 0.19 \\
    \midrule
    \multirow{2}{*}{$\hat{\Omega}^{2; \, \{10\%, 20\%\}}_{\mathrm{TIN}}$}
        & All coefficients & 0.77 $\pm$ 0.11 & 0.09 $\pm$ 0.02 \\
        & SIC & 0.68 $\pm$ 0.14 & 0.10  $\pm$ 0.03 \\
    \midrule
    \multirow{2}{*}{$\hat{\Omega}^3_{\mathrm{TIN}}$}
        & All coefficients & 0.72  $\pm$ 0.10  & 0.10 $\pm$ 0.02 \\
        & SIC & 0.56  $\pm$ 0.14  & 0.11  $\pm$ 0.02 \\
    \bottomrule
  \end{tabular}
  }
\end{table}

\begin{table}[htbp]
  \centering
  \caption{Weights (the fixed-effects ones) obtained from fitting a linear model incorporating all \emph{Last-Layer Imbalanced Forgetting Coefficients} separately on each partition in our sampled benchmarks. A mixed-effects formulation with random intercepts and slopes is used, where class-wise forgetting serves as the dependent variable. All variables (SIC, CIC, NIC, and class-wise forgetting) are standardized within each partition.}
  \label{tab:coefficients}
  \resizebox{\linewidth}{!}{%
  \begin{tabular}{llccc}
    \toprule
    & & 10\% Classes & 20\% Classes & 50\% Classes \\
    \midrule
    \multicolumn{5}{c}{\textbf{Sampled benchmark:} $\hat{\Omega}_{\mathrm{C100}}$}\\
    \midrule
    \multicolumn{5}{c}{$\hat{\Omega}_{\mathrm{C100}}^2$}\\
    \midrule
    \multirow{3}{*}{8\% Reh.}
        & SIC & $0.87$ & $0.83$ & $0.78$ \\
        & CIC & $0.10$ & $0.11$ & $0.13$ \\
        & NIC & $0.08$ & $0.08$ & $0.13$ \\
    \midrule
    \multirow{3}{*}{20\% Reh.}
        & SIC & $0.75$ & $0.75$ & $0.62$ \\
        & CIC & $0.15$ & $0.12$ & $0.08$ \\
        & NIC & $0.21$ & $0.17$ & $0.30$ \\
    \midrule
    \multirow{3}{*}{40\% Reh.}
        & SIC & $0.64$ & $0.60$ & $0.39$ \\
        & CIC & $0.13$ & $0.17$ & $0.04$ \\
        & NIC & $0.28$ & $0.29$ & $0.49$ \\
    \midrule
    \multicolumn{5}{c}{$\hat{\Omega}_{\mathrm{C100}}^3$}\\
    \midrule
    \multirow{3}{*}{8\% Reh.}
        & SIC & $0.85$ & $0.84$ & $-$ \\
        & CIC & $0.08$ & $0.16$ & $-$ \\
        & NIC & $0.07$ & $0.04$ & $-$ \\
    \midrule
    \multirow{3}{*}{20\% Reh.}
        & SIC & $0.79$ & $0.75$ & $-$ \\
        & CIC & $0.05$ & $0.09$ & $-$ \\
        & NIC & $0.15$ & $0.18$ & $-$ \\
    \midrule
    \multirow{3}{*}{40\% Reh.}
        & SIC & $0.68$ & $0.54$ & $-$ \\
        & CIC & $0.05$ & $0.00$ & $-$ \\
        & NIC & $0.22$ & $0.37$ & $-$ \\
    \midrule
    \multicolumn{5}{c}{\textbf{Sampled benchmark:} $\hat{\Omega}_{\mathrm{TIN}}$}\\
    \midrule
    \multicolumn{5}{c}{$\hat{\Omega}_{\mathrm{TIN}}^2$}\\
    \midrule
    \multirow{3}{*}{8\% Reh.}
        & SIC & $0.59$ & $0.52$ & $0.42$ \\
        & CIC & $0.45$ & $0.54$ & $0.53$ \\
        & NIC & $0.17$ & $0.14$ & $0.20$ \\
    \midrule
    \multirow{3}{*}{20\% Reh.}
        & SIC & $0.70$ & $0.61$ & $0.54$ \\
        & CIC & $0.30$ & $0.37$ & $0.37$ \\
        & NIC & $0.05$ & $0.11$ & $0.14$ \\
    \midrule
    \multirow{3}{*}{40\% Reh.}
        & SIC & $0.62$ & $0.58$ & $0.41$ \\
        & CIC & $0.24$ & $0.29$ & $0.22$ \\
        & NIC & $0.08$ & $0.11$ & $0.24$ \\
    \midrule
    \multicolumn{5}{c}{$\hat{\Omega}_{\mathrm{TIN}}^3$}\\
    \midrule
    \multirow{3}{*}{8\% Reh.}
        & SIC & $0.55$ & $0.48$ & $-$ \\
        & CIC & $0.59$ & $0.64$ & $-$ \\
        & NIC & $0.03$ & $0.03$ & $-$ \\
    \midrule
    \multirow{3}{*}{20\% Reh.}
        & SIC & $0.64$ & $0.54$ & $-$ \\
        & CIC & $0.35$ & $0.42$ & $-$ \\
        & NIC & $0.04$ & $0.03$ & $-$ \\
    \midrule
    \multirow{3}{*}{40\% Reh.}
        & SIC & $0.58$ & $0.39$ & $-$ \\
        & CIC & $0.23$ & $0.18$ & $-$ \\
        & NIC & $0.05$ & $0.23$ & $-$ \\
    \bottomrule
\end{tabular}
}
\end{table}

\section{Empirical Comparison between LOG-SIM and the Last-Layer Imbalanced Forgetting Coefficients}\label{sec:comparison_logits_with_our_coefficients}
\noindent A recent study \cite{RN152} hypothesizes that imbalanced forgetting stems from unequal initial similarity between newly introduced classes and previously learned ones, i.e., past classes that are more similar to the new ones at the beginning of an incremental step will suffer greater forgetting at the end of that step. For each past class, the authors measure the degree of initial similarity as the average logit assigned to that class over all samples from the new classes prior to starting training on the corresponding incremental step. Hereafter, we refer to this metric as LOG-SIM.

As shown in Table \ref{tab:SW-LOO_protocol_LOG_similarity}, LOG-SIM exhibits, on average, a modest ability to predict the ranking of past classes in terms of forgetting across the second incremental steps in our sampled benchmarks; however, its average performance deteriorates substantially across the third steps. This suggests that the predictive power of LOG-SIM rapidly loses effectiveness as the sequence of incremental steps progresses, indicating that it provides limited explanatory power for imbalanced forgetting. 

Notably, as further reported in Table \ref{tab:SW-LOO_protocol_LOG_similarity}, the \emph{Last-Layer Imbalanced Forgetting Coefficients} substantially outperform LOG-SIM, on average, in predicting the forgetting-based ranking of past classes. Additionally, the mean predictive performance of NIC alone---the coefficient most closely related to LOG-SIM\footnote{NIC is the only coefficient among the \emph{Last-Layer Imbalanced Forgetting Coefficients} that depends on the new classes}---is also substantially higher than that of LOG-SIM.

\begin{table}[htbp]
  \centering
  \caption{Mean $\pm$ standard deviation of the predictive strength achieved by LOG-SIM for the second and third incremental steps in $\hat{\Omega}_{\mathrm{C100}}$ and $\hat{\Omega}_{\mathrm{TIN}}$. The predictive strength is with respect to ranking past classes by their forgetting and is measured using Spearman's correlation $\rho$. For comparison, the predictive strength obtained on the same incremental steps using (i) a linear model incorporating all \emph{Last-Layer Imbalanced Forgetting Coefficients}, evaluated under the SW-LOO protocol, and (ii) NIC are also reported.}
  \label{tab:SW-LOO_protocol_LOG_similarity}
  \renewcommand{\arraystretch}{0.5}
  \resizebox{0.80\linewidth}{!}{%
  \begin{tabular}{lccc}
    \toprule
    & All coefficients & NIC & LOG-SIM \\
    \midrule
    & \multicolumn{3}{c}{ $\hat{\Omega}_{\mathrm{C100}}$}\\
    \midrule
    $\hat{\Omega}^2_{\mathrm{C100}}$
        & 0.90 $\pm$ 0.08 & 0.74 $\pm$ 0.15 & 0.63 $\pm$ 0.18\\ 
    \midrule
    $\hat{\Omega}^{2;\,\{10\%, 20\%\}}_{\mathrm{C100}}$
        & 0.91 $\pm$ 0.09 & 0.75 $\pm$ 0.17 & 0.67 $\pm$ 0.19\\
    \midrule
    $\hat{\Omega}^3_{\mathrm{C100}}$
        & 0.88 $\pm$ 0.08 & 0.65 $\pm$ 0.14 & 0.39 $\pm$ 0.18\\
    \midrule
    & \multicolumn{3}{c}{ $\hat{\Omega}_{\mathrm{TIN}}$}\\
    \midrule
    $\hat{\Omega}^2_{\mathrm{TIN}}$
        & 0.74 $\pm$ 0.11 & 0.53 $\pm$ 0.14 & 0.45 $\pm$ 0.15\\
    \midrule
    $\hat{\Omega}^{2; \, \{10\%, 20\%\}}_{\mathrm{TIN}}$
        & 0.77 $\pm$ 0.11 & 0.53 $\pm$ 0.15 & 0.49 $\pm$ 0.17\\
    \midrule
    $\hat{\Omega}^3_{\mathrm{TIN}}$
        & 0.72 $\pm$ 0.10 & 0.46 $\pm$ 0.14 & 0.26 $\pm$ 0.15\\
    \bottomrule
\end{tabular}
  }
\end{table}

\section{Finer-Grained Analysis of the Associations between NIC and SIC}\label{sec:finer_grained_anal_assoc_NIC_SIC}

\noindent In Section \ref{sec:ana_relationship_SIC_NIC}, we analyzed the associations between NIC and SIC for the second and third incremental steps in both $\Omega_{\mathrm{C100}}$ and $\Omega_{\mathrm{TIN}}$, aggregating across different percentages of rehearsal retention and newly introduced classes. Here, we provide a finer-grained analysis that explicitly characterizes how these associations vary with these proportions.

Specifically, Table \ref{tab:spearman_NIC-SIC} reports the estimated mean and standard deviation of these associations for each partition in $\Omega_d^k$, with $d \in \{\mathrm{C100}, \mathrm{TIN}\}$ and $k \in \{2,3\}$. The results show that increasing the proportion of new classes leads to lower associations on average. Moreover, these associations increase, on average, from $8\%$ to $20\%$ rehearsal retention, and drop from $20\%$ to $40\%$. Importantly, the standard deviations are small relative to the means, indicating that these associations are consistent within partitions.

Furthermore, for each $\Omega_d^k$, we also conduct a type-III two-way ANOVA to assess whether rehearsal retention percentage, percentage of newly introduced classes, and their interaction have statistically significant effects on the associations in $\hat{\Omega}_d^k$, and to quantify the corresponding effect sizes. To ensure robustness to potential heteroskedasticity, we perform the ANOVA using the HC3 heteroskedasticity-consistent covariance estimator. The results of this ANOVA are reported in Table \ref{tab:anova_SIC_NIC}. Overall, none of the considered factors has a statistically significant effect on the associations between NIC and SIC in $\Omega_{\mathrm{C100}}$ and for the second incremental steps in $\Omega_{\mathrm{TIN}}$. For the third incremental steps in $\Omega_{\mathrm{TIN}}$, only rehearsal retention percentage shows a statistically significant effect, though with a small effect size.

\begin{table*}[htbp]
  \centering
  \caption{Estimated mean $\pm$ standard deviation (with 95\% CIs) of the association between NIC and SIC, measured by Spearman's correlation $\rho$,  for each partition in our benchmarks. "Pooled" estimates are obtained from the union of partitions sharing the same rehearsal retention percentage or percentage of newly introduced classes.}
  \label{tab:spearman_NIC-SIC}
  \resizebox{\textwidth}{!}{%
  \begin{tabular}{lccc@{\hskip 20pt}c}
    \toprule
    & 10\% Classes & 20\% Classes & 50\% Classes  & \textbf{Pooled}\\
    \midrule
    \multicolumn{5}{c}{\textbf{Benchmark:} $\Omega_{\mathrm{C100}}$}\\
    \midrule
    \multicolumn{5}{c}{$\Omega_{\mathrm{C100}}^2$}\\
    \midrule
    8\% Reh.  & $0.68$ {\tiny $(0.60, 0.75)$} $\pm$ $0.23$ {\tiny $(0.19, 0.29)$} & $0.69$ {\tiny $(0.65, 0.72)$} $\pm$ $0.11$ {\tiny $(0.09, 0.13)$} & $0.64$ {\tiny $(0.60, 0.67)$} $\pm$ $0.12$ {\tiny $(0.09, 0.17)$} & $0.67$ {\tiny $(0.64, 0.70)$} $\pm$ $0.17$ {\tiny $(0.14, 0.20)$}\\
    \midrule
    20\% Reh. & $0.79$ {\tiny $(0.72, 0.85)$} $\pm$ $0.23$ {\tiny $(0.17, 0.32)$} & $0.75$ {\tiny $(0.72, 0.79)$} $\pm$ $0.12$ {\tiny $(0.10, 0.16)$} & $0.67$ {\tiny $(0.64, 0.70)$} $\pm$ $0.10$ {\tiny $(0.08, 0.14)$} & $0.74$ {\tiny $(0.71, 0.77)$} $\pm$ $0.16$ {\tiny $(0.14, 0.21)$}\\
    \midrule
    40\% Reh. & $0.79$ {\tiny $(0.71, 0.84)$} $\pm$ $0.22$ {\tiny $(0.17, 0.28)$} & $0.72$ {\tiny $(0.67, 0.77)$} $\pm$ $0.20$ {\tiny $(0.15, 0.26)$} & $0.60$ {\tiny $(0.58, 0.63)$} $\pm$ $0.08$ {\tiny $(0.07, 0.09)$} & $0.71$ {\tiny $(0.67, 0.75)$} $\pm$ $0.18$ {\tiny $(0.16, 0.22)$}\\
    \midrule
    \addlinespace[7pt]
    \textbf{Pooled} & $0.76$ {\tiny $(0.72, 0.80)$} $\pm$ $0.23$ {\tiny $(0.20, 0.27)$} & $0.72$ {\tiny $(0.70, 0.75)$} $\pm$ $0.15$ {\tiny $(0.13, 0.19)$} & $0.64$ {\tiny $(0.62, 0.66)$} $\pm$ $0.11$ {\tiny $(0.09, 0.13)$} & $-$\\
    \midrule
    \multicolumn{5}{c}{$\Omega_{\mathrm{C100}}^3$}\\
    \midrule
    8\% Reh.  & $0.61$ {\tiny $(0.56, 0.66)$} $\pm$ $0.16$ {\tiny $(0.12, 0.23)$} & $0.63$ {\tiny $(0.60, 0.67)$} $\pm$ $0.11$ {\tiny $(0.09, 0.13)$} & $-$ & $0.62$ {\tiny $(0.60, 0.65)$} $\pm$ $0.14$ {\tiny $(0.11, 0.18)$}\\
    \midrule
    20\% Reh. & $0.66$ {\tiny $(0.61, 0.71)$} $\pm$ $0.16$ {\tiny $(0.11, 0.24)$} & $0.63$ {\tiny $(0.58, 0.67)$} $\pm$ $0.14$ {\tiny $(0.12, 0.17)$} & $-$ & $0.64$ {\tiny $(0.61, 0.68)$} $\pm$ $0.15$ {\tiny $(0.13, 0.20)$}\\
    \midrule
    40\% Reh. & $0.61$ {\tiny $(0.54, 0.66)$} $\pm$ $0.18$ {\tiny $(0.15, 0.22)$} & $0.59$ {\tiny $(0.56, 0.63)$} $\pm$ $0.11$ {\tiny $(0.09, 0.13)$} & $-$ & $0.60$ {\tiny $(0.56, 0.63)$} $\pm$ $0.15$ {\tiny $(0.13, 0.18)$}\\
    \midrule
    \addlinespace[7pt]
    \textbf{Pooled} & $0.63$ {\tiny $(0.60, 0.66)$} $\pm$ $0.17$ {\tiny $(0.15, 0.20)$} & $0.62$ {\tiny $(0.60, 0.64)$} $\pm$ $0.12$ {\tiny $(0.11, 0.14)$} & $-$ & $-$\\
    \midrule
    \multicolumn{5}{c}{\textbf{Benchmark:} $\Omega_{\mathrm{TIN}}$}\\
    \midrule
    \multicolumn{5}{c}{$\Omega_{\mathrm{TIN}}^2$}\\
    \midrule
    8\% Reh.  & $0.56$ {\tiny $(0.51, 0.62)$} $\pm$ $0.17$ {\tiny $(0.14, 0.21)$} & $0.52$ {\tiny $(0.48, 0.56)$} $\pm$ $0.13$ {\tiny $(0.11, 0.15)$} & $0.54$ {\tiny $(0.51, 0.57)$} $\pm$ $0.09$ {\tiny $(0.07, 0.10)$} & $0.54$ {\tiny $(0.52, 0.57)$} $\pm$ $0.13$ {\tiny $(0.12, 0.15)$}\\
    \midrule
    20\% Reh. & $0.58$ {\tiny $(0.53, 0.63)$} $\pm$ $0.16$ {\tiny $(0.12, 0.21)$} & $0.64$ {\tiny $(0.60, 0.68)$} $\pm$ $0.12$ {\tiny $(0.10, 0.15)$} & $0.60$ {\tiny $(0.57, 0.62)$} $\pm$ $0.08$ {\tiny $(0.07, 0.10)$} & $0.61$ {\tiny $(0.58, 0.63)$} $\pm$ $0.13$ {\tiny $(0.11, 0.16)$}\\
    \midrule
    40\% Reh. & $0.59$ {\tiny $(0.53, 0.65)$} $\pm$ $0.19$ {\tiny $(0.16, 0.24)$} & $0.55$ {\tiny $(0.51, 0.60)$} $\pm$ $0.15$ {\tiny $(0.12, 0.19)$} & $0.57$ {\tiny $(0.54, 0.59)$} $\pm$ $0.08$ {\tiny $(0.07, 0.10)$} & $0.57$ {\tiny $(0.54, 0.60)$} $\pm$ $0.15$ {\tiny $(0.13, 0.17)$}\\
    \midrule
    \addlinespace[7pt]
    \textbf{Pooled} & $0.58$ {\tiny $(0.55, 0.61)$} $\pm$ $0.17$ {\tiny $(0.15, 0.20)$} & $0.57$ {\tiny $(0.55, 0.60)$} $\pm$ $0.14$ {\tiny $(0.13, 0.16)$} & $0.57$ {\tiny $(0.55, 0.58)$} $\pm$ $0.09$ {\tiny $(0.08, 0.09)$} & $-$\\
    \midrule
    \multicolumn{5}{c}{$\Omega_{\mathrm{TIN}}^3$}\\
    \midrule
    8\% Reh.  & $0.43$ {\tiny $(0.37, 0.49)$} $\pm$ $0.18$ {\tiny $(0.15, 0.22)$} & $0.42$ {\tiny $(0.38, 0.46)$} $\pm$ $0.11$ {\tiny $(0.09, 0.14)$} & $-$ & $0.43$ {\tiny $(0.39, 0.46)$} $\pm$ $0.15$ {\tiny $(0.13, 0.18)$}\\
    \midrule
    20\% Reh. & $0.52$ {\tiny $(0.47, 0.56)$} $\pm$ $0.13$ {\tiny $(0.11, 0.16)$} & $0.50$ {\tiny $(0.46, 0.53)$} $\pm$ $0.11$ {\tiny $(0.09, 0.13)$} & $-$ & $0.51$ {\tiny $(0.48, 0.53)$} $\pm$ $0.12$ {\tiny $(0.11, 0.14)$}\\
    \midrule
    40\% Reh. & $0.49$ {\tiny $(0.45, 0.54)$} $\pm$ $0.14$ {\tiny $(0.11, 0.19)$} & $0.42$ {\tiny $(0.38, 0.46)$} $\pm$ $0.12$ {\tiny $(0.10, 0.14)$} & $-$ & $0.46$ {\tiny $(0.42, 0.49)$} $\pm$ $0.14$ {\tiny $(0.12, 0.16)$}\\
    \midrule
    \addlinespace[7pt]
    \textbf{Pooled} & $0.48$ {\tiny $(0.45, 0.51)$} $\pm$ $0.16$ {\tiny $(0.14, 0.18)$} & $0.45$ {\tiny $(0.42, 0.47)$} $\pm$ $0.12$ {\tiny $(0.11, 0.13)$} & $-$ & $-$\\
    \bottomrule
  \end{tabular}
  }
\end{table*}

\begin{table}[htbp]
  \centering
  \caption{Type-III two-way ANOVA results for the associations between NIC and SIC on our benchmarks, using rehearsal retention percentage and percentage of new classes as factors, together with their interaction term. Reported statistics include the $F$-statistic, $p$-value, and partial eta squared ($\eta^2_p$). }
  \label{tab:anova_SIC_NIC}
  \resizebox{\linewidth}{!}{%
  \begin{tabular}{llccc}
    \toprule
    & & {$F$-statistic} & $p$-value & {$\eta^2_p$}  \\
    \midrule
    \multicolumn{5}{c}{\textbf{Benchmark:} $\Omega_{\mathrm{C100}}$}\\
    \midrule
    \multirow{3}{*}{$\Omega_{\mathrm{C100}}^2$}
        & Rehearsal retention percentage & $2.01$ & $1.36 \times 10^{-1}$ & $0.01$  \\
        & Percentage new classes & $2.06$ & $1.29 \times 10^{-1}$ & $0.01$    \\
        & Interaction & $1.06$ & $3.75 \times 10^{-1}$ & $0.01$    \\
    \midrule
    \multirow{3}{*}{$\Omega_{\mathrm{C100}}^{2; \,\{10\%, 20\%\}}$}
        & Rehearsal retention percentage & $2.01$ & $1.36 \times 10^{-1}$ & $0.02$    \\
        & Percentage new classes & $1.38$ & $2.41 \times 10^{-1}$ & $0.01$   \\
        & Interaction & $0.77$ & $4.62 \times 10^{-1}$ & $0.01$   \\
    \midrule
    \multirow{3}{*}{$\Omega_{\mathrm{C100}}^3$}
        & Rehearsal retention percentage & $1.59$ & $2.05 \times 10^{-1}$ & $0.01$   \\
        & Percentage new classes & $0.85$ & $3.56 \times 10^{-1}$ & $0.00$   \\
        & Interaction & $0.91$ & $4.03 \times 10^{-1}$ & $0.01$   \\
    \midrule
    \multicolumn{5}{c}{\textbf{Benchmark:} $\Omega_{\mathrm{TIN}}$}\\
    \midrule
    \multirow{3}{*}{$\Omega_{\mathrm{TIN}}^2$}
        & Rehearsal retention percentage & $0.24$ & $7.86 \times 10^{-1}$ & $0.00$  \\
        & Percentage new classes & $0.69$ & $5.00 \times 10^{-1}$ & $0.00$   \\
        & Interaction & $1.33$ & $2.59 \times 10^{-1}$ & $0.01$   \\
    \midrule
    \multirow{3}{*}{$\Omega_{\mathrm{TIN}}^{2; \,\{10\%, 20\%\}}$}
        & Rehearsal retention percentage & $0.24$ & $7.86 \times 10^{-1}$ & $0.00$   \\
        & Percentage new classes & $0.88$ & $3.50 \times 10^{-1}$ & $0.00$  \\
        & Interaction & $2.16$ & $1.18 \times 10^{-1}$ & $0.02$   \\
    \midrule
    \multirow{3}{*}{$\Omega_{\mathrm{TIN}}^3$}
        & Rehearsal retention percentage & $3.20$ & $4.25 \times 10^{-2}$ & $0.03$    \\
        & Percentage new classes & $0.00$ & $9.94 \times 10^{-1}$ & $0.00$    \\
        & Interaction &$1.56$ & $2.11 \times 10^{-1}$ & $0.01$   \\
    \bottomrule
\end{tabular}
  }
\end{table}

\section{Disentangling NIC’s Predictive Overlap with SIC and CIC}\label{sec:disentangling_NIC_pred_overlap_SIC_CIC}

\noindent In Section \ref{sec:individual_pred_strength}, we showed that the mean predictive strength of NIC decreases considerably across our benchmarks after controlling for both SIC and CIC, indicating that a substantial portion of the predictive information it captures overlaps with that of these two coefficients. Here, we examine whether this overlap is primarily with SIC or CIC. Results in Table \ref{tab:all_partial_spearmans_NIC} show that the overlap is mainly attributable to SIC.

\begin{table*}[htbp]
    \centering
    \caption{Estimated mean $\pm$ standard deviation (with 95\% CIs) of multiple types of associations between NIC and class-wise forgetting: (i) Spearman's correlations ($\rho$); (ii) partial Spearman's correlations controlling for both SIC and CIC ($\rho_p$); (iii) partial Spearman's correlations controlling only for SIC ($\rho_p^{\text{SIC}}$); (iv) partial Spearman's correlations controlling only for CIC ($\rho_p^{\text{CIC}}$). Results are reported for the second and third incremental steps in $\Omega_{\mathrm{C100}}$ and $\Omega_{\mathrm{TIN}}$. }
  \label{tab:all_partial_spearmans_NIC}
    \resizebox{\textwidth}{!}{%
    \begin{tabular}{lcccc}
    \toprule
    & $\rho$ & $\rho_p$ &  $\rho_p^{\text{SIC}}$ & $\rho_p^{\text{CIC}}$ \\
    \midrule
    \multicolumn{5}{c}{\textbf{Benchmark:} $\Omega_{\mathrm{C100}}$}\\
    \midrule
    $\Omega_{\mathrm{C100}}^2$ & 0.74 {\tiny(0.72, 0.75)} $\pm$ 0.15 {\tiny(0.13, 0.17)} & 0.30 {\tiny(0.26, 0.33)} $\pm$ 0.28 {\tiny(0.26, 0.31)} & 0.35 {\tiny (0.32, 0.39)} $\pm$ 0.29 {\tiny (0.27, 0.31)} & 0.72 {\tiny (0.71, 0.74)} $\pm$ 0.16 {\tiny (0.15, 0.19)}\\
    \midrule
    $\Omega_{\mathrm{C100}}^{2; \, \{10\%, 20\%\}}$ & 0.75 {\tiny(0.73, 0.77)} $\pm$ 0.17 {\tiny(0.15, 0.20)} & 0.26 {\tiny(0.22, 0.31)} $\pm$ 0.32 {\tiny(0.30, 0.35)} & 0.32 {\tiny (0.28, 0.37)} $\pm$ 0.32 {\tiny (0.30, 0.35)} & 0.74 {\tiny (0.71, 0.76)} $\pm$ 0.19 {\tiny (0.17, 0.22)}\\
    \midrule
    $\Omega_{\mathrm{C100}}^3$ & 0.65 {\tiny(0.64, 0.67)} $\pm$ 0.14 {\tiny(0.12, 0.15)} & 0.22 {\tiny(0.19, 0.25)} $\pm$ 0.23 {\tiny(0.22, 0.25)} & 0.29 {\tiny (0.26, 0.32)} $\pm$ 0.23 {\tiny (0.21, 0.25)} & 0.64 {\tiny (0.62, 0.66)} $\pm$ 0.15 {\tiny (0.13, 0.16)}\\
    \midrule
    \multicolumn{5}{c}{\textbf{Benchmark:} $\Omega_{\mathrm{TIN}}$}\\
    \midrule
    $\Omega_{\mathrm{TIN}}^2$ & 0.53 {\tiny(0.51, 0.54)} $\pm$ 0.14 {\tiny(0.13, 0.15)} & 0.20 {\tiny(0.17, 0.22)} $\pm$ 0.19 {\tiny(0.18, 0.21)} & 0.26 {\tiny (0.24, 0.28)} $\pm$ 0.20 {\tiny (0.18, 0.21)} & 0.52 {\tiny (0.50, 0.53)} $\pm$ 0.14 {\tiny (0.13, 0.16)}\\
    \midrule
    $\Omega_{\mathrm{TIN}}^{2; \, \{10\%, 20\%\}}$ & 0.53 {\tiny(0.51, 0.55)} $\pm$ 0.15 {\tiny(0.14, 0.16)} & 0.17 {\tiny(0.14, 0.19)} $\pm$ 0.22 {\tiny(0.20, 0.24)} & 0.22 {\tiny (0.19, 0.25)} $\pm$ 0.21 {\tiny (0.20, 0.23)} & 0.51 {\tiny (0.49, 0.53)} $\pm$ 0.16 {\tiny (0.15, 0.18)}\\
    \midrule
    $\Omega_{\mathrm{TIN}}^3$ & 0.46 {\tiny(0.44, 0.47)} $\pm$ 0.14 {\tiny(0.12, 0.15)} & 0.13 {\tiny(0.11, 0.15)} $\pm$ 0.16 {\tiny(0.14, 0.17)} & 0.27 {\tiny (0.25, 0.29)} $\pm$ 0.16 {\tiny (0.15, 0.18)} & 0.40 {\tiny (0.38, 0.42)} $\pm$ 0.14 {\tiny (0.13, 0.15)}\\
    \bottomrule
  \end{tabular}
  }
\end{table*}

\section{Pairwise Associations among the Last-Layer Imbalanced Forgetting Coefficients}\label{sec:pairwise_associations_imbalanced_forgetting}

\noindent In Table \ref{tab:spearman_pairwise_Imbalanced_Forgetting_Coefficients}, we present the estimated means and standard deviations of the pairwise associations among SIC, CIC, and NIC, measured by Spearman’s correlation $\rho$, across our benchmarks.

The results indicate that SIC and CIC exhibit highly unstable associations across our benchmarks, with mean associations close to zero, suggesting no consistent relationship. By contrast, CIC and NIC show modest average associations, which are higher across the third incremental steps than across the second in both $\Omega_{\mathrm{C100}}$ and $\Omega_{\mathrm{TIN}}$. The associations between NIC and SIC were analyzed in Section \ref{sec:ana_relationship_SIC_NIC}.

\begin{table}[htbp]
  \centering
  \caption{Estimated mean $\pm$ standard deviation (with $95\%$ CIs) of the pairwise associations among SIC, CIC, and NIC, measured by Spearman's correlation $\rho$, for the second and third incremental steps in $\Omega_{\mathrm{C100}}$ and $\Omega_{\mathrm{TIN}}$.}
  \label{tab:spearman_pairwise_Imbalanced_Forgetting_Coefficients}
  \resizebox{0.8\linewidth}{!}{%
  \begin{tabular}{lc@{\hskip 30pt}c}
    \toprule
    &  & $\rho$\\
    \midrule
    \multicolumn{3}{c}{\textbf{Benchmark:} $\Omega_{\mathrm{C100}}$}\\
    \midrule
    \multicolumn{3}{c}{$\Omega_{\mathrm{C100}}^2$}\\
    \midrule
    SIC-CIC &  & \phantom{$-$}$0.05$ {\tiny $(0.02, 0.09)$} $\pm$ $0.32$ {\tiny $(0.30, 0.34)$}\\
    \midrule
    SIC-NIC &  &  \phantom{$-$}$0.71$ {\tiny $(0.69, 0.73)$} $\pm$ $0.17$ {\tiny $(0.16, 0.19)$}\\
    \midrule
    CIC-NIC &  & \phantom{$-$}$0.26$ {\tiny $(0.23, 0.29)$} $\pm$ $0.28$ {\tiny $(0.26, 0.31)$}\\
    \midrule
    \multicolumn{3}{c}{$\Omega_{\mathrm{C100}}^{2;\, \{10\%,20\%\}}$}\\
    \midrule
    SIC-CIC &  & \phantom{$-$}$0.05$ {\tiny $(0.00, 0.10)$} $\pm$ $0.35$ {\tiny $(0.33, 0.38)$}\\
    \midrule
    SIC-NIC &  &  \phantom{$-$}$0.74$ {\tiny $(0.72, 0.76)$} $\pm$ $0.19$ {\tiny $(0.17, 0.22)$}\\
    \midrule
    CIC-NIC &  &  \phantom{$-$}$0.24$ {\tiny $(0.20, 0.29)$} $\pm$ $0.32$ {\tiny $(0.30, 0.36)$}\\
    \midrule
    \multicolumn{3}{c}{$\Omega_{\mathrm{C100}}^3$}\\
    \midrule
    SIC-CIC &  & \phantom{$-$}$0.07$ {\tiny $(0.03, 0.12)$} $\pm$ $0.31$ {\tiny $(0.29, 0.33)$}\\
    \midrule
    SIC-NIC &  &  \phantom{$-$}$0.62$ {\tiny $(0.60, 0.64)$} $\pm$ $0.15$ {\tiny $(0.13, 0.17)$}\\
    \midrule
    CIC-NIC &  & \phantom{$-$}$0.40$ {\tiny $(0.37, 0.43)$} $\pm$ $0.21$ {\tiny $(0.20, 0.23)$}\\
    \midrule
    \multicolumn{3}{c}{\textbf{Benchmark:} $\Omega_{\mathrm{TIN}}$}\\
    \midrule
    \multicolumn{3}{c}{$\Omega_{\mathrm{TIN}}^2$}\\
    \midrule
    SIC-CIC & &  \phantom{$-$}$-0.01$ {\tiny $(-0.04, 0.01)$} $\pm$ $0.22$ {\tiny $(0.20, 0.23)$}\\
    \midrule
    SIC-NIC & &  \phantom{$-$}$0.57$ {\tiny $(0.56, 0.59)$} $\pm$ $0.14$ {\tiny $(0.13, 0.15)$}\\
    \midrule
    CIC-NIC &  &  \phantom{$-$}$0.18$ {\tiny $(0.16, 0.20)$} $\pm$ $0.21$ {\tiny $(0.19, 0.23)$}\\
    \midrule
    \multicolumn{3}{c}{$\Omega_{\mathrm{TIN}}^{2;\, \{10\%,20\%\}}$}\\
    \midrule
    SIC-CIC &  & \phantom{$-$}$0.05$ {\tiny $(0.02, 0.08)$} $\pm$ $0.23$ {\tiny $(0.22, 0.25)$}\\
    \midrule
    SIC-NIC &  &  \phantom{$-$}$0.58$ {\tiny $(0.56, 0.60)$} $\pm$ $0.16$ {\tiny $(0.14, 0.17)$}\\
    \midrule
    CIC-NIC &  &  \phantom{$-$}$0.19$ {\tiny $(0.16, 0.22)$} $\pm$ $0.24$ {\tiny $(0.22, 0.26)$}\\
    \midrule
    \multicolumn{3}{c}{$\Omega_{\mathrm{TIN}}^3$}\\
    \midrule
    SIC-CIC &  &  \phantom{$-$}$-0.04$ {\tiny $(-0.06, -0.01)$} $\pm$ $0.20$ {\tiny $(0.18, 0.22)$}\\
    \midrule
    SIC-NIC &  &  \phantom{$-$}$0.46$ {\tiny $(0.45, 0.48)$} $\pm$ $0.14$ {\tiny $(0.13, 0.15)$}\\
    \midrule
    CIC-NIC &  & \phantom{$-$}$0.33$ {\tiny $(0.30, 0.35)$} $\pm$ $0.18$ {\tiny $(0.16, 0.19)$}\\
    \bottomrule
\end{tabular}
  }
\end{table}

\section{NIC-SIC Controlled Experiments}\label{sec:controlled_exps_NIC_SIC}
\noindent In Section \ref{sec:ana_relationship_SIC_NIC}, we hypothesized that higher NIC values for a past class during an incremental step lead to higher SIC by steering the training trajectory toward regions of the weight space where rehearsal samples from that class induce greater interfering bias in the last layer. Here, we provide empirical support for this through a set of controlled experiments (CON-EXPS). 

Specifically,  for each partition in $\hat{\Omega}_d^k$, with $d \in \{\mathrm{C100}, \mathrm{TIN}\}$ and $k \in \{2,3\}$, we randomly select a single incremental step and re-run it 19 times using different sets of new classes, while keeping all other factors unchanged (i.e., the rehearsal set, the random seed, and the initial network parameters).\footnote{This procedure is not applied to partitions containing incremental steps in which 50\% of new classes are introduced, as it would not be possible to sample alternative class sets for the re-runs.} After these runs, we analyze, for each past class, whether higher NIC values are associated with higher SIC values across runs. This association is quantified using Spearman's rank correlation. 

\begin{figure*}[!t]
  \centering
  \includegraphics[width=0.9\textwidth]{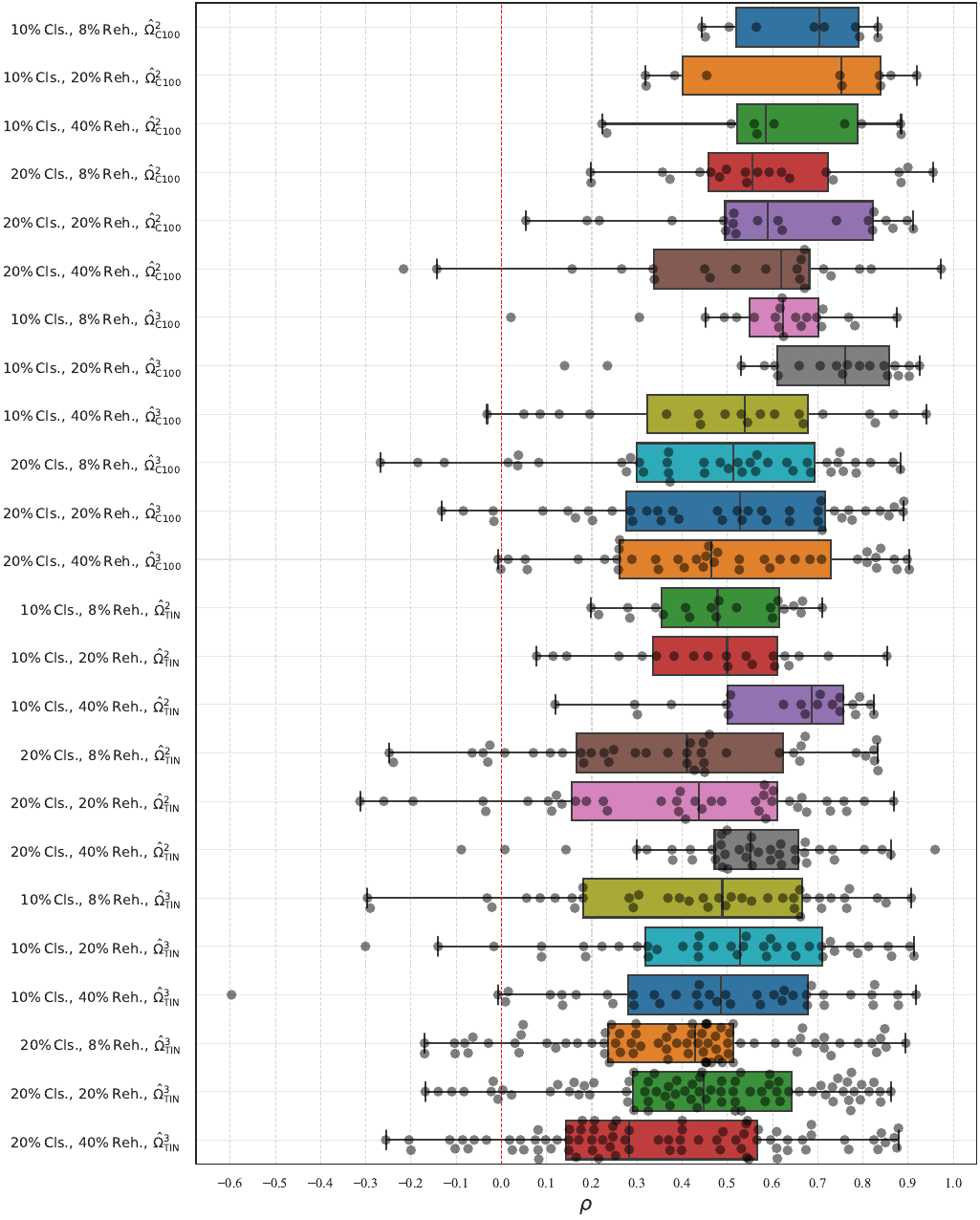}
  \caption{Box-and-whisker plot illustrating the distribution of class-wise associations between NIC and SIC, measured across re-runs via Spearman's correlation $\rho$, obtained for each partition from the CON-EXPS set. Black points represent individual data points.}
\label{fig:controlled_exps_distributions}
\end{figure*}

The mean and standard deviation of the class-wise associations for each partition are reported in Table \ref{tab:controlled_experiments}, while Figure \ref{fig:controlled_exps_distributions} shows their distributions. Notably, the mean correlations range from modest to strong, with standard deviations that are non-negligible relative to the means. This suggests that rehearsal sets corresponding to different past classes respond differently to increases in NIC. Nevertheless, as illustrated in Figure \ref{fig:controlled_exps_distributions}, most classes exhibit at least a modest positive association across runs, indicating that their rehearsal sets tends to produce higher SIC values in runs where the new classes interfere more strongly with them, as measured by NIC. Since all factors other than the new classes are identical
across re-runs, these results provide empirical support for our hypothesis stated in Section \ref{sec:ana_relationship_SIC_NIC}.

\begin{table}[htbp]
  \centering
  \caption{Mean $\pm$ standard deviation of the class-wise associations between NIC and SIC, measured across re-runs via Spearman's correlation $\rho$, obtained for each partition from the CON-EXPS set.}
  \label{tab:controlled_experiments}
  \begin{tabular}{lcc}
    \toprule
    & 10\% Classes & 20\% Classes \\
    \midrule
    \multicolumn{3}{c}{\textbf{Sampled benchmark:} $\hat{\Omega}_{\mathrm{C100}}$}\\
    \midrule
    \multicolumn{3}{c}{$\hat{\Omega}_{\mathrm{C100}}^2$}\\
    \midrule
    8\% Reh.  & $0.69 \pm 0.15$ & $0.64 \pm 0.21$  \\
    \midrule
    20\% Reh. & $0.71 \pm 0.23$ & $0.66 \pm 0.24$
 \\
    \midrule
    40\% Reh. & $0.66 \pm 0.23$ & $0.58 \pm 0.30$ \\
    \midrule
    \multicolumn{3}{c}{$\hat{\Omega}_{\mathrm{C100}}^3$}\\
    \midrule
    8\% Reh.  & $0.62 \pm 0.18$ & $0.51 \pm 0.29$ \\
    \midrule
    20\% Reh. & $0.76 \pm 0.21$ & $0.54 \pm 0.29$ \\
    \midrule
    40\% Reh. & $0.56 \pm 0.28$ &  $0.55 \pm 0.27$\\
    \midrule
    \multicolumn{3}{c}{\textbf{Sampled benchmark:} $\hat{\Omega}_{\mathrm{TIN}}$}\\
    \midrule
    \multicolumn{3}{c}{$\hat{\Omega}_{\mathrm{TIN}}^2$}\\
    \midrule
    8\% Reh.  & $0.49 \pm 0.16$ & $0.41 \pm 0.30$ \\
    \midrule
    20\% Reh. & $0.49 \pm 0.20$ & $0.43 \pm 0.30$ \\
    \midrule
    40\% Reh. & $0.64 \pm 0.20$ & $0.58 \pm 0.21$ \\
    \midrule
    \multicolumn{3}{c}{$\hat{\Omega}_{\mathrm{TIN}}^3$}\\
    \midrule
    8\% Reh.  & $0.49 \pm 0.30$ & $0.43 \pm 0.26$ \\
    \midrule
    20\% Reh. & $0.54 \pm 0.29$ & $0.48 \pm 0.26$ \\
    \midrule
    40\% Reh. & $0.52 \pm 0.31$ & $0.40 \pm 0.29$ \\
    \bottomrule
    \end{tabular}
\end{table}

\end{document}